\documentclass{IEEEtran}

\usepackage{multirow}
\usepackage{floatrow}
\usepackage[figuresright]{rotating}
\usepackage{booktabs}
\usepackage{subfigure}
\usepackage{epsfig}
\usepackage[ruled,vlined,lined,linesnumbered]{algorithm2e}
\usepackage{color, soul}
\usepackage{multirow}
\usepackage{arydshln}
\usepackage{amsmath}
\usepackage{amsfonts}
\usepackage{float}
\usepackage{graphicx}
\usepackage{enumitem}
\usepackage{colortbl}
\usepackage{color}
\usepackage[table*]{xcolor}
\xdefinecolor{gray95}{gray}{0.55}
\xdefinecolor{gray25}{gray}{0.8}
\newlist{abbrv}{itemize}{1}
\setlist[abbrv,1]{label=,labelwidth=0.4in,align=parleft,itemsep=-0.2\baselineskip,leftmargin=!}
\usepackage{amssymb}
\usepackage{cases}
\usepackage{float}
\floatstyle{plaintop}
\restylefloat{table}
\usepackage{xpatch}
\makeatletter
\patchcmd\@makecaption{\\}{.~}{}{\fail}
\makeatletter
\usepackage{lettrine}
\usepackage{lipsum}
\usepackage[numbers,sort&compress]{natbib}
\usepackage{amsfonts,amssymb}
\usepackage[font={footnotesize}]{caption}
{ \bgroup
  \addtolength\abovedisplayshortskip{#1}
  \addtolength\abovedisplayskip{#1}
  \addtolength\belowdisplayshortskip{#1}
  \addtolength\belowdisplayskip{#1}}
{\egroup\ignorespacesafterend}
\newcommand\blfootnote[1]{%
\begingroup 
\renewcommand\thefootnote{}\footnote{#1}%
\addtocounter{footnote}{-1}%
\endgroup 
}

\begin{document}

\title{Variable Division and Optimization for Constrained Multiobjective Portfolio Problems}
\author{Yi Chen, Aimin Zhou, \emph{Senior Member, IEEE}}

\maketitle

\blfootnote{Y. Chen and A. Zhou are with Shanghai Key Laboratory of Multidimensional Information Processing, and the Department of Computer Science and Technology, East China Normal University, Shanghai 200241, China (e-mail: 52164500006@stu.ecnu.edu.cn; amzhou@cs.ecnu.edu.cn). Corresponding Author: A. Zhou}

\begin{abstract}
Variable division and optimization (D\&O) is a frequently utilized algorithm design paradigm in Evolutionary Algorithms (EAs). A D\&O EA divides a variable 
into partial variables and then optimize them respectively. A complicated problem is thus divided into simple subtasks. For example, a variable of portfolio problem can be divided into two partial variables, i.e. the selection of assets and the allocation of capital. Thereby, we optimize these two partial variables respectively. There is no formal discussion about how are the partial variables iteratively optimized and why can it work for both single- and multi-objective problems in D\&O. In this paper, this gap is filled. According to the discussion, an elitist selection method for partial variables in multiobjective problems is developed. Then this method is incorporated into the Decomposition-Based Multiobjective Evolutionary Algorithm (D\&O-MOEA/D). With the help of a mathematical programming optimizer, it is achieved on the constrained multiobjective portfolio problems. In the empirical study, D\&O-MOEA/D is implemented for 20 instances and recent Chinese stock markets. The results show the superiority and versatility of D\&O-MOEA/D on large-scale instances while the performance of it on small-scale problems is also not bad. The former targets convergence towards the Pareto front and the latter helps promote diversity among the non-dominated solutions during the search process.
\end{abstract}


\IEEEpeerreviewmaketitle
\section{Introduction}
\lettrine[lines=2]{M}{aking} the complicated simple is one effective idea frequently used in the realm of computer science~\cite{boyd2004convex, cormen2009introduction}. Variable division and optimization (D\&O), dividing the variable into partial variables and then optimizing them, is naturally inspired by this idea. It is not a new thing for Evolutionary Algorithms (EAs) since it has been frequently utilized in Cooperative Coevolution~\cite{antonio2017coevolutionary,ma2018survey}, Neuroevolution~\cite{yao1999evolving} and constrained portfolio optimization~\cite{moral2006selection}. The original problems are simplified by dividing variables in these works. On the other hand, D\&O is reminiscent of Dived-and-Conquer, however, they are different because the latter one emphasizes solving subproblems recursively~\cite{cormen2009introduction}. 

\begin{figure}[htbp]\footnotesize
\graphicspath{{figs/}}
\centerline{\includegraphics[width=0.5\columnwidth]{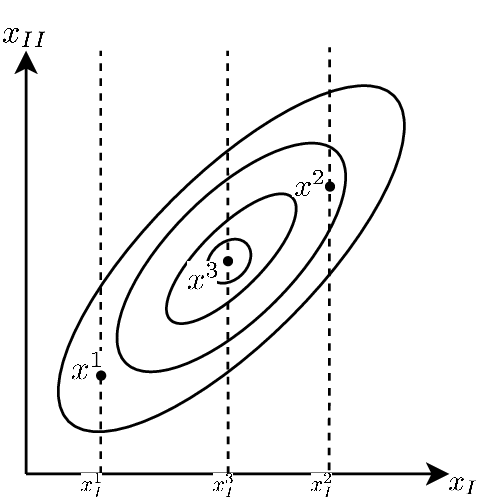}}
 \caption{Simplified local optimal solution search for a single objective problem.} 
 \label{local_optimum_search_1}
\end{figure}

\begin{figure}[htbp]\footnotesize
\graphicspath{{figs/}}
\centerline{\includegraphics[width=0.5\columnwidth]{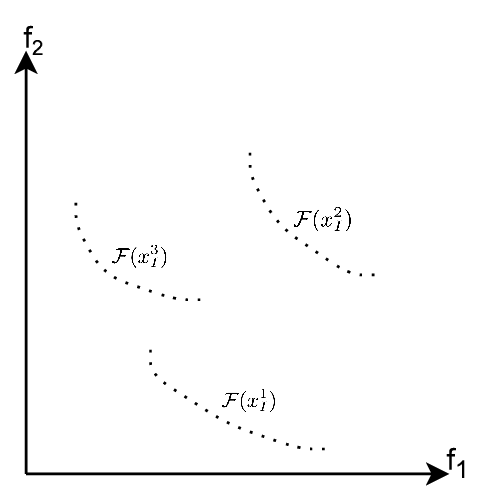}}
 \caption{Simplified local optimal solution search for a multiobjective problem.} 
 \label{local_optimum_search_2}
\end{figure}

Formally, a minimization problem is given as
\begin{equation}
\label{minimization}
\begin{gathered}
\mathop{\min}_x \ F(x) \  \equiv \mathop{\min}_{x_{\uppercase\expandafter{\romannumeral1}},x_{\uppercase\expandafter{\romannumeral2}}}\ F(x_{\uppercase\expandafter{\romannumeral1}},x_{\uppercase\expandafter{\romannumeral2}}) \\
s.t \quad x \in \Omega, \  x_{\uppercase\expandafter{\romannumeral1}} \in \Omega_{\uppercase\expandafter{\romannumeral1}} , \  x_{\uppercase\expandafter{\romannumeral2}} \in \Omega_{\uppercase\expandafter{\romannumeral2}}  \ ,
\end{gathered}
\end{equation}where $x$ is the decision variable, $x_{\uppercase\expandafter{\romannumeral1}}$ and $x_{\uppercase\expandafter{\romannumeral2}}$ are two partial variables, and $x=(x_{\uppercase\expandafter{\romannumeral1}},x_{\uppercase\expandafter{\romannumeral2}})$. $\Omega_{(\cdot)}$ is the search space of each variable, and $\Omega_{\uppercase\expandafter{\romannumeral1}}\times \Omega_{\uppercase\expandafter{\romannumeral2}} = \Omega$. With D\&O, one variable is divided into partial variables and thereby they can be optimized separately, i.e. the right problem in Eq. (\ref{minimization}). Thereafter, the divided problem is generally presented as two sequential subtasks
\begin{equation}
\label{minimization_sub1}
\begin{gathered}
\mathop{\arg\min}_{x_{\uppercase\expandafter{\romannumeral1}}}\ F(x_{\uppercase\expandafter{\romannumeral1}},x_{\uppercase\expandafter{\romannumeral2}}) \\
s.t \quad x_{\uppercase\expandafter{\romannumeral1}} \in \Omega_{\uppercase\expandafter{\romannumeral1}},\ x_{\uppercase\expandafter{\romannumeral2}} \in \Omega_{\uppercase\expandafter{\romannumeral2}}.
\end{gathered}
\end{equation}
\begin{equation}
\label{minimization_sub2}
\begin{gathered}
\mathop{\arg\min}_{x_{\uppercase\expandafter{\romannumeral2}}}\ F(x_{\uppercase\expandafter{\romannumeral2}};x_{\uppercase\expandafter{\romannumeral1}}) \\
s.t  \quad x_{\uppercase\expandafter{\romannumeral2}} \in \Omega_{\uppercase\expandafter{\romannumeral2}}.
\end{gathered}
\end{equation}
After the division has been made, a major challenge induced by the problem (\ref{minimization_sub1}) should be tackled when $x_{\uppercase\expandafter{\romannumeral1}}$ is optimized in EAs. This is how to perform an elitist selection
among partial variables $x_{\uppercase\expandafter{\romannumeral1}}$ to optimize them iteratively? Because there is a dimensionality mismatch between a partial variable $x_{\uppercase\expandafter{\romannumeral1}}$ and the variable $x$ of the original problem, making the evaluation of the original objective function not possible~\cite{yang2017turning}. The fitness of a partial variable $x_{\uppercase\expandafter{\romannumeral1}}$ should be assigned as the fitness of its local optimal function values. A set of these values are stated as
\begin{equation}
\label{minimization_set}
\begin{gathered}
\mathcal{F}(x_{\uppercase\expandafter{\romannumeral1}})=\mathop{\min}_{x_{\uppercase\expandafter{\romannumeral2}} \in \Omega_{\uppercase\expandafter{\romannumeral2}}}F(x_{\uppercase\expandafter{\romannumeral2}};x_{\uppercase\expandafter{\romannumeral1}}).
\end{gathered}
\end{equation}In this manner, we will never miss the global optimal variable if every local optimum are checked. Concisely, for D\&O based EAs, we can get the a global optimal variable if each partial variable is evaluated as the performance of its corresponding local optima. Figs.~\ref{local_optimum_search_1} and~\ref{local_optimum_search_2} illustrate this for single and multiobjective problems respectively. Fig.~\ref{local_optimum_search_1} shows the variable space of $x_{\uppercase\expandafter{\romannumeral1}}$ and $x_{\uppercase\expandafter{\romannumeral2}}$. The local optimum of each $x_{\uppercase\expandafter{\romannumeral1}}$ is just one point, i.e. $|\mathcal{F}(x_{\uppercase\expandafter{\romannumeral1}})|=1$. It is easy to determine $x^3$ is the optimal solution. Notwithstanding, it is not natural for multiobjective problems. Fig.~\ref{local_optimum_search_2} shows the objective space of a biobjective problem. The local optima of each $x_{\uppercase\expandafter{\romannumeral1}}$ consists of a local Pareto front (PF) where $|\mathcal{F}(x_{\uppercase\expandafter{\romannumeral1}})|\ge1$. We are hindered from extending the advantages of D\&O to multiobjective problems since no method exists for performing an elitist selection among local PFs.


In the discussion above, methods, involving D\&O, are mainly concentrated in Cooperative Coevolution, Neuroevolution, and constrained portfolio optimization. Concerning the ways of getting the $\mathcal{F}(x_{\uppercase\expandafter{\romannumeral1}})$, they can also be roughly classified as straightforward and approximate methods as the taxonomy in~\cite{jin2005comprehensive,jin2011surrogate}. The related literature will be briefly presented by fields and ways. Note that although D\&O looks like bi-level optimization~\cite{sinha2017review}, they are two different models. Because D\&O only involves one objective function contrary to bi-level optimization. Furthermore, most of the problems introduced below merely include one level optimization.

First, in Cooperative Coevolution, algorithms are special cases that frequently utilize D\&O~\cite{antonio2017coevolutionary,ma2018survey}. For instance, the variables are divided into partial variables (some groups), and each partial variable is taken as $x_{\uppercase\expandafter{\romannumeral1}}$ recursively. Then, the $\mathcal{F}(x_{\uppercase\expandafter{\romannumeral1}})$ is approximated via incorporating $x_{\uppercase\expandafter{\romannumeral1}}$ and the partial variables from other groups~\cite{omidvar2013cooperative,yang2008large}. Second, in Neurovelution, some methods utilize EAs for the task of neural architecture search ($x_{\uppercase\expandafter{\romannumeral1}}$). Then backpropagation algorithms are implemented for the training. These hybrid methods are straightforward in a broad definition, because they find the $\mathcal{F}(x_{\uppercase\expandafter{\romannumeral1}})$ directly~\cite{sun2018evolving,lu2019nsga}. The approximate $\mathcal{F}(x_{\uppercase\expandafter{\romannumeral1}})$ is also employed in this field, EAs are employed to find approximate optimal connection weights of small networks for the non-deep neural networks~\cite{yao1997new,liu1996population}. Nowadays, a new technique, supernet, is implemented to approximate optimal connection weights efficiently for deep neural networks~\cite{lu2020nsganetv2}. Third, in constrained portfolio optimization, a portfolio problem can be divided into a combinatorial ($x_{\uppercase\expandafter{\romannumeral1}}$) and continuous subtask when a cardinality constraint is involved. Some methods use EAs to address the combinational tasks and thereby call exact mathematical approaches for the continuous problems. The straightforward $\mathcal{F}(x_{\uppercase\expandafter{\romannumeral1}})$ is adopted in these hybrid methods~\cite{moral2006selection,branke2009portfolio}. All of them can be considered as D\&O-based EAs since they do divide the variable into partial variables and then optimize them separately.


In this work, regarding D\&O, we propose an algorithm to perform the elitist selection for local PFs of partial variables with a Decomposition-Based Multiobjective Evolutionary Algorithm (D\&O-MOEA/D). This algorithm is achieved on constrained portfolio problems, which are mixed-integer problems, via calling a mathematical optimizer.

In summary, the contributions of this paper are as follows:
\begin{itemize}
\item For the first time, the concept of D\&O is summarized formally. We point out that how should the elitist selection for partial variables perform. 
\item According to the discussion about D\&O, D\&O-MOEA/D is achieved. Conducted simulation experiments on 20 instances demonstrate the superiority of D\&O-MOEA/D. First, it keeps small gaps with the exact multiobjective method on small-scale problems while it is more universal than the exact method on large-scale problems. Second, it dominates the implemented EAs when all 20 problems are considered.
\item Further, it is applied to recent Chinese stock markets, Shanghai and Shenzhen stock markets. The results confirm the feasibility and the value in the application of the proposed method.
\end{itemize}

The remainder of this paper is presented as follows. Section \uppercase\expandafter{\romannumeral2} presents the formulation of the constrained multiobjective portfolio problem and emphasizes the emerging challenge for performing elitist selection among local PFs. In Section \uppercase\expandafter{\romannumeral3}, the basic MOEA/D is introduced. Meanwhile, we discuss how is the division of the variable $x$ made, and how is the optimization of $x_{\uppercase\expandafter{\romannumeral1}}$ and $x_{\uppercase\expandafter{\romannumeral2}}$ performed. Then, empirical studies are presented in Section \uppercase\expandafter{\romannumeral4}. Finally, a conclusion of this work and promising topics constitute Section \uppercase\expandafter{\romannumeral5}.


\section{Elitist Selection for Local PFs}
In this section, firstly, the formulation of the constrained multiobjective problem is presented. Then we introduce how to perform an elitist selection of the local PFs.
\subsection{Basic Problem Formulation}
\label{basic_model}
Portfolio optimization has been instrumental in the development of financial markets~\cite{kolm201460}. As portfolio problems develop in practical applications, EAs outperform conventional optimization algorithms since they can tackle complex constraints~\cite{ponsich2013survey,ertenlice2018survey}.
In this subsection, multiobjective portfolio problems with cardinality constraints are introduced, because D\&O is very suitable for them~\cite{moral2006selection,branke2009portfolio}.

This constrained multiobjective portfolio problem, which is an extension Mean-Variance~\cite{1952MarkowitzPS} model, is a biobjective problem, finding a trade-off between return and risk. Moreover, it involves four constraints: (i) cardinality constraint, (ii) floor and ceiling constraint, (iii) pre-assignment constraint, and (iv) round lot constraint~\cite{DBLP:journals/corr/abs-1909-08748}. Let $\Gamma$ be the set of available assets, $\Pi$ be the set of pre-assigned assets. Define $|\Gamma|=n$, $|\Pi|=L$. The parameter $\mu_i$ and $\sigma_{i}$ are the expectation and standard deviation of return for asset $i$, $i\in \Gamma$. The parameter $\rho_{ij}$ is the correlation coefficient of the assets $i$ and $j$, $i,j \in \Gamma$, and $\sigma_{ij}=\rho_{ij}\sigma_{i}\sigma_{j}$, where all $\sigma_{ij}$ constitute a covariance matrix (risk). Let $\epsilon_i$ and $\upsilon_i$ be the lower and upper bound, and $\tau$ be the round lot size of the assets; $z_i$ be the binary parameter representing pre-assignment of the asset $i$. The problem formulation follows as
{
\begin{equation}
\label{risk}
{\rm min}\quad f_1=\sum_{i\in \Gamma}\sum_{j\in \Gamma}w_{i}w_{j}\sigma _{ij},
\end{equation}\vspace{-2ex}
\begin{equation}
\label{return}
{\rm max}\quad f_2=\sum_{i \in \Gamma}w_{i}\mu _{i},
\end{equation}\vspace{-2ex}
\begin{equation}
\label{sum_to_one}
s.t. \sum_{i \in \Gamma}w_{i}= 1,
\end{equation}\vspace{-2ex}
\begin{equation}
\label{Cardinality Constraint}
\sum_{i \in \Gamma}s_{i}=K,
\end{equation}\vspace{-2ex}
\begin{equation}
\label{floor_ceiling}
\epsilon _{i}s_{i}\leq w_{i}\leq \upsilon _{i}s_{i}, \qquad i \in \Gamma,
\end{equation}\vspace{-2ex}
\begin{equation}
\label{Pre-assignment}
s_{i}\leq z_{i}, \qquad i \in \Gamma,
\end{equation}\vspace{-2ex}
\begin{equation}
\label{Round Lot}
w_{i}=\tau \delta_{i}, \qquad i \in \Gamma,\quad \delta_{i}\in \mathbb{Z}_{+},
\end{equation}\vspace{-2ex}
\begin{equation}
\label{discrete_binary}
s_{i}\in \{0,1\}, \qquad i \in \Gamma,
\end{equation}
}where $w=\{w_1,\dots,w_{n}\}^T$ is a portfolio vector, Eqs.~(\ref{risk}) and (\ref{return}) are two respective objectives, minimizing the risk and maximizing the return, in the portfolio optimization that conflict with each other. Eq.~(\ref{sum_to_one}) requires that all the capital should be invested in a valid portfolio. Eq.~(\ref{Cardinality Constraint}) is the cardinality constraint (i.e., just $K$ assets are selected) and $s=\{s_1,\dots,s_n\}^T$ is an indicator vector; $s_i=1$ if the asset $i$ is selected, and $s_i=0$ otherwise. Eq.~(\ref{floor_ceiling}) is the floor and ceiling constraint. Besides, Eq.~(\ref{Pre-assignment}) represents that the asset $i$ must be included in a portfolio if $z_i=1$. It is a pre-assignment constraint. Thereafter, Eq.~(\ref{Round Lot}) defines the round lot constraint and $\delta_i$ is a multiplier. The round lot sizes $\tau$ is set to be a same constant, with which 1 is divisible, since it will be really hard to handle a flexible one. In that case, it will be beyond the scope of the main discussion in this work. Finally, Eq.~(\ref{discrete_binary}), which is the discrete constraint, implies that the $s_i$ must be binary.

According to D\&O, a portfolio $w$ can be divided into $x_{\uppercase\expandafter{\romannumeral1}}$ and $x_{\uppercase\expandafter{\romannumeral2}}$, where $x_{\uppercase\expandafter{\romannumeral1}}$ is the combination of selected assets, i.e. a selection vector $s$, and $x_{\uppercase\expandafter{\romannumeral2}}$ represents the corresponding weights. Thereby, $x_{\uppercase\expandafter{\romannumeral1}}$ and $x_{\uppercase\expandafter{\romannumeral2}}$ are optimized in the EA and an exact mathematical optimizer, $\textbf{Opt}$, respectively.
\subsection{Elitist Selection of The Local PFs}
As for the local PF of each $x_{\uppercase\expandafter{\romannumeral1}}$, we find that only in $two$ extreme cases that an elitist selection can be performed with existing methods, and the discussion is presented in the supplement. Moreover, it has been proved that the shape of a local PF in the Mean-Variance model is a convex parabola and it will be discontinuous when the round lot constraint is included~\cite{steinmean}. Therefore, no method exists for performing an elitist selection for the local PFs in this problem.

\begin{figure*}[htbp]\footnotesize
\graphicspath{{figs/}}
\begin{minipage}{0.47\linewidth}
\centerline{\includegraphics[width=0.5\columnwidth]{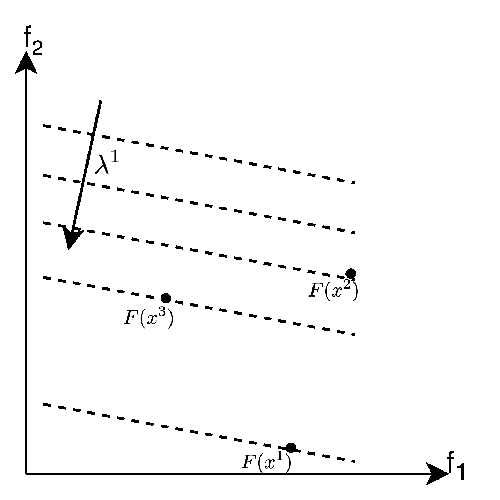}}
\centerline{(a)}
\end{minipage}
\begin{minipage}{0.47\linewidth}
\centerline{\includegraphics[width=0.5\columnwidth]{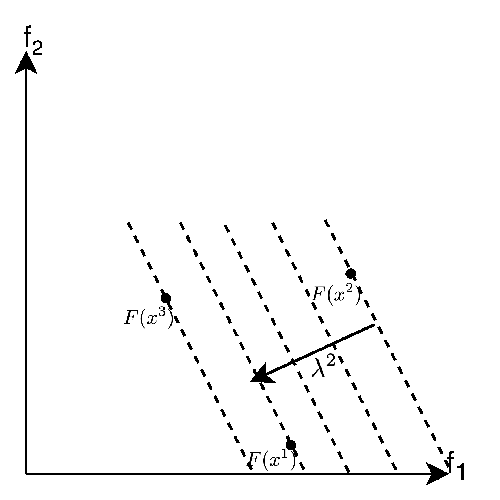}}
\centerline{(b)}
\end{minipage}
 \caption{The weighting method for variables on a biobjective problem with two different weight vectors.} 
 \label{W12}
\end{figure*}

To our best knowledge, the only work, including a PF-based (envelope-based) algorithm, proposed by J. Branke et al~\cite{branke2009portfolio} is one Dominance-based algorithm. Although this PF-based algorithm has some limitations, it encourages us to develop a more versatile one. Since it is hard to determine the dominance among local PFs, we try to conduct a decomposition approach for them through exploiting MOEA/D~\cite{2007QingfuMOEAD}. MOEA/D is a decomposition-based algorithm, who uses aggregation functions to decompose the PF into a number of scalar objective subproblems~\cite{zhou2011multiobjective}. After solving these subproblems, optimal solutions will construct approximate PFs~\cite{,li2009multiobjective,wang2015adaptive,qi2014moea,ma2017tchebycheff}. Several methods for constructing aggregation functions can be found in the literature~\cite{miettinen2012nonlinear}. Without loss of generality, the weighting method is employed. It is defined as
\begin{equation}
\label{WS_minimization}
\begin{gathered}
{\rm min}\ g^{ws}(x;\lambda)=\sum_i^m \lambda_if_i(x) \quad \\
s.t \quad x \in \Omega,
\end{gathered}
\end{equation}where $\lambda=(\lambda_1,\lambda_2,\dots,\lambda_m)$ is a weight vector. Fig.~\ref{W12} illustrates how is the elitist selection for variables performed on a biobjective problem through the weighting method. Every solution are assigned with different $g^{ws}$ according to different weight vectors. For instance, regarding the weight vector $\lambda^1$, $x^1$ is the best, on the contrary, $x^3$ is the best according to $\lambda^2$. 

\begin{figure*}[htbp]\footnotesize
\graphicspath{{figs/}}
\begin{minipage}{0.47\linewidth}
\centerline{\includegraphics[width=0.5\columnwidth]{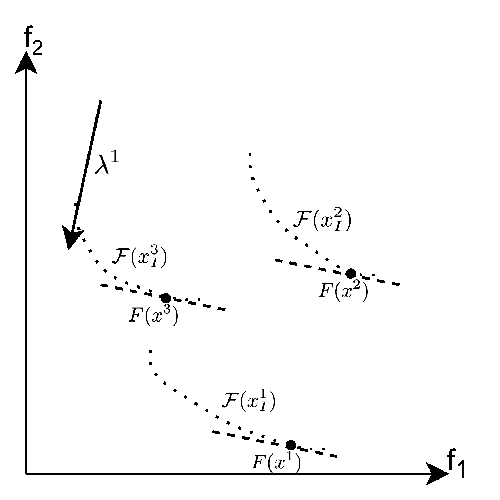}}
\centerline{(a)}
\end{minipage}
\begin{minipage}{0.47\linewidth}
\centerline{\includegraphics[width=0.5\columnwidth]{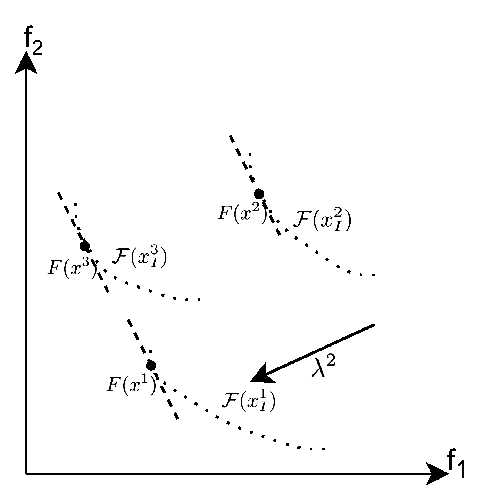}}
\centerline{(b)}
\end{minipage}
 \caption{The weighting method for partial variables on a biobjective problem with two different weight vectors.} 
 \label{local_optimum_search_WS12}
\end{figure*}

This aggregation function can also be used to evaluate the fitness of each $x_{\uppercase\expandafter{\romannumeral1}}$, which reduces a local PF $\mathcal{F}(x_{\uppercase\expandafter{\romannumeral1}})$ to a variable $x$. In these cases, a $x$ can be obtained through slightly altering the problem~(\ref{WS_minimization}) and then solving it, the alteration is given as
\begin{equation}
\label{IWS_minimization}
\begin{gathered}
\mathop{\min}_{x_{\uppercase\expandafter{\romannumeral2}} }\ g^{lws}(x_{\uppercase\expandafter{\romannumeral2}} ;x_{\uppercase\expandafter{\romannumeral1}} ,\lambda)= \sum_i^m \lambda_if_i(x_{\uppercase\expandafter{\romannumeral1}},x_{\uppercase\expandafter{\romannumeral2}})\quad \\
s.t \quad x_{\uppercase\expandafter{\romannumeral2}}  \in \Omega_{\uppercase\expandafter{\romannumeral2}},
\end{gathered}
\end{equation}where the solution is $x_{\uppercase\expandafter{\romannumeral2}}$. Consequently, we can evaluate $x_{\uppercase\expandafter{\romannumeral1}}$ by $x=(x_{\uppercase\expandafter{\romannumeral1}},x_{\uppercase\expandafter{\romannumeral2}})$. For example, in Fig.~\ref{local_optimum_search_WS12}, $x$ can be obtained via settling the problem~(\ref{IWS_minimization}) with the $\textbf{Opt}$ for a partial variable $x_{\uppercase\expandafter{\romannumeral1}}$ and a weight vector $\lambda^1$. Then $\mathcal{F}(x_{\uppercase\expandafter{\romannumeral1}})$ is reduced to $x$ and thereby $x^1$ is the best concerning the weight vector $\lambda^1$, denoting $x_{\uppercase\expandafter{\romannumeral1}}^1$ is the best. According to this elitist selection method for local PFs, D\&O-MOEA/D is proposed and detailed in the next section.
\section{D\&O-MOEA/D} 
In this section, the framework of MOEA/D is introduced firstly and then we develop it for the constrained portfolio problems. 


MOEA/D is an iterative algorithm, where a population of $N$ solutions $x^1,\dots,x^N$ is optimized iteratively. Each solution $x^i$ corresponds to the subproblem $i$. Meanwhile, each subproblem $i$ has a $T$-neighborhood structure, involving closest $T$ subproblems according to the weight vector. Assume each subproblem is a minimization problem and the objective function of subproblem $i$ is $g^i(x)$. The main loop of MOEA/D works as follows.

For each subproblem $i$:
\subsubsection{Mating} Select some solutions from the $T$-neighborhood of subproblem $i$ and generate new solution $x_{new}^i$ with the selected ones via reproduction operators.
\subsubsection{Evaluating} For each subproblem $j$ in the $T$-neighborhood of subproblem $i$, evaluate $g^j(x^i_{new})$.

\subsubsection{Replacing} For each solution $x^j$ in the $T$-neighborhood of subproblem $i$, replace $x^j$ with $x^i_{new}$ if $g^j(x^i_{new})<g^j(x^j)$. 

We take MOEA/D-DE in~\cite{li2009multiobjective} as the basic of D\&O-MOEA/D. D\&O-MOEA/D is achieved via searching the $x_{\uppercase\expandafter{\romannumeral1}}$ (combination of assets) with evolutionary method and optimizing $x_{\uppercase\expandafter{\romannumeral2}}$ (weights) with an exact mathematical optimizer, $\textbf{Opt}$. Specifically, $x_{\uppercase\expandafter{\romannumeral1}}$ is searched in the $Matting$ and $x_{\uppercase\expandafter{\romannumeral2}}$ is optimized in the $Evaluating$. 

Regarding $x_{\uppercase\expandafter{\romannumeral1}}$, it is the combination of assets. Several representations and reproduction operators can be chosen for $x_{\uppercase\expandafter{\romannumeral1}}$, our real-valued representation, CCS~\cite{DBLP:journals/corr/abs-1909-08748}, is followed in this work. Specifically, a real-valued $n$-dimensional vector, in $[0,1]^{n}$, is utilized. Assets which are pre-selected or correspond to the $K-L$ highest values are selected, therefore, real-valued vectors can represent combinations of selected assets, the cardinality and pre-assignment constraints are handled~\cite{1994BeanGenetic}. As for the reproduction, it involves a combination of Differential Evolution~\cite{1996RainerDE,das2010differential} and Polynomial Mutation~\cite{2002DebNSGA2} as in MOEA/D-DE and a swap operator as in~\cite{DBLP:journals/corr/abs-1909-08748}. The combination operator and the swap operator will occur with a same probability, i.e. 50\% for each one. 

Concerning $x_{\uppercase\expandafter{\romannumeral2}}$, it is the capital weight. After the $x_{\uppercase\expandafter{\romannumeral1}}$ is determined, we are going to settle the subtask, the problem~(\ref{IWS_minimization}) for $x_{\uppercase\expandafter{\romannumeral2}}$. Specifically, when $x_{\uppercase\expandafter{\romannumeral1}}$, i.e. variable $s$, is found by the EA, this subtask is specified as  
{\setlength{\abovedisplayskip}{3pt}
\setlength{\belowdisplayskip}{3pt}
\begin{equation}
\label{ICDVs_subtask}
\begin{gathered}
{\rm min}\quad g^{lws} = \lambda_1\sum_{i\in \Gamma'}\sum_{j\in \Gamma'}w_{i}w_{j}\sigma _{ij}-\lambda_2\sum_{i \in \Gamma'}w_{i}\mu _{i}\\
{\rm s.t} \sum_{i \in \Gamma'}w_{i}= 1,\\
\epsilon _{i}\leq w_{i}\leq \upsilon _{i}, \qquad i \in \Gamma',\\
w_{i}=\tau\delta_{i}, \qquad i \in \Gamma',\quad \delta_{i} \in \mathbb{Z}_{+},
\end{gathered}
\end{equation}
}where $\lambda=(\lambda_1,\lambda_2)$ is the weight vector and the $\Gamma'$ is a set of assets, where $\Gamma' \subset \Gamma$ and $|\Gamma'|=K$. Thus it implies Eqs.~(\ref{Cardinality Constraint}),~(\ref{Pre-assignment}) and~(\ref{discrete_binary}) are satisfied. The problem~(\ref{ICDVs_subtask}) is a quadratic problem with an integer constraint (round lot constraint). Therefore, a quadratic optimizer of the commercial solver CPLEX is taken as the $\textbf{Opt}$ and a proved optimal solution of $g^i(x)$ can be obtained in a very short time, about 30 milliseconds, with an ordinary personal computer.

\section{Empirical Studies}
%

In the empirical studies, D\&O-MOEA/D is compared with three methods, two state-of-the-art MOEAs, and an exact multiobjective method AUGMECON2\&CPLEX, on 20 constrained portfolio instances. In the field of optimization, both exact methods and heuristics take important positions. We would like to analyze the different performances among the exact mathematical programming method, EAs, and the hybrid one since they are supposed to have their advantages. This section is hence composed of two parts. First, the description about implementation details, including instances, a constraint set, employed algorithms, and parameter settings. Besides, the performance metrics. Second, the performance analysis.

\subsection{Implementation Details}
All of the problems consist of instances from OR-Library and NGINX. The first one, OR-Library, contains 5 classic instances, and the second one, NGINX, includes 15 instances established by the historical stock data from Yahoo Finance website. The instances are sorted with the number of available assets and are shown in Table~\ref{20instances}\footnote{All the data is available at \textcolor{blue}{https://github.com/CYLOL2019/Portfolio-Instances}}. Concerning the constraint set, it follows one set from~\cite{DBLP:journals/corr/abs-1909-08748} as cardinality ${K=10}$, floor $\epsilon=0.01$, ceiling $\upsilon=1.0$, pre-assignment ${z_{30}=1}$ and round lot $\tau=0.008$.
\begin{table}[htbp]\small
\centering 
\caption{Twenty Used Instances.}
\label{20instances}
\begin{tabular}{|llll|}\hline
Instance&\multicolumn{1}{c}{Origin}&\multicolumn{1}{c}{Name}&\multicolumn{1}{c|}{\#Assets}\\\hline
D1    &Hong Kong    &Hang Seng    &31\\
D2    &Germany &DAX 100 &                      85\\
D3    &UK   &FTSE 100 &                        89\\
D4    &USA  &NASDQ  Financial00   &            91\\
D5    &USA  &NYSE   US100            &         94\\
D6    &USA  &S\&P  100    &                     98\\
D7    &UK   &FTSE ACT250     &                 128\\
D8    &USA  &NASDAQ Biotech     &              130\\
D9    &USA  &NASDQ  Telecom     &              139\\
D10    &Italy  &MIBTEL     &                    167\\
D11    &USA  &NYSE   World        &             170\\
D12    &Japan    &Nikkei    &$225$\\
D13    &Australia &All ordinaries     &         264\\
D14    &USA  &NASDAQ Bank    &                  380\\
D15    &USA & NASDQ  Computer         &         417\\
D16    &USA  &S\&P    500        &               469\\
D17    &Korea    &KOSPI Composite    &562\\
D18    &USA  &NASDQ  Industrial      &          808\\
D19    &USA    &AMEX Composite    & 1893\\
D20    &USA    &NASDAQ     &2235\\
\hline
\end{tabular}
\end{table}

Three algorithms are implemented for comparison, they are listed below\footnote{MODEwAwL and CCS/MOEA/D are available at \textcolor{blue}{https://github.com/CYLOL2019/SEC-CCS}, and AUGMECON2\&CPLEX, D\&O-MOEA/D, and related files have been uploaded to \textcolor{blue}{https://github.com/CYLOL2019/D\&O-MOEAD}.}.

\subsubsection{MODEwAwL~\cite{2014KhinMODEwawl}} It is a learning guided MOEA. It represents $w$ with binary and real-valued vectors. The search of the binary vector, representing the combination, is guided by a learning mechanism, and the search of the real-valued vector, representing the weights, is mainly based on DE.
\subsubsection{CCS/MOEA/D~\cite{DBLP:journals/corr/abs-1909-08748,chen2018evolutionary}} It incorporates a compressed coding scheme into MOEA/D. It utilizes only one real-valued vector to represent both the combination and weights, making use of the dependence among combinations and weights.
\subsubsection{AUGMECON2\&CPLEX~\cite{mavrotas2009effective,mavrotas2013improved}} An exact mathematical multiobjective algorithm based on $\varepsilon$-constraint method (AUGMECON2) is employed. It is incorporated with a quadratic optimizer of CPLEX for the constrained multiobjective portfolio problems and the number of grids is set as the size of the population in the EAs.

All of the implemented algorithms involve some parameters. There are common parameters, such as population size and parameters for common generating operators. In the meantime, there are also parameters that are independent of the algorithms, namely, parameters for the neighborhood in MOEA/D and the number of grids in AUGMECON2. They are all shown in Table~\ref{parameter_settings}. Stop criteria are non-trivial for these methods, however, they are always different~\cite{liu2018termination}. Therefore a maximum fitness evaluation and a time budget are both used for the termination detection. 20 times independent executions for every method with two different stop criteria are carried out on a server computer. This server computer is constructed with 2 64 cores AMD EPYC CPUs and 256GB memories.
\begin{table}[htbp] \small
\centering
\caption{The parameter settings for all algorithms.}
\label{parameter_settings}
\begin{tabular}{|ll|}\hline
Common parameters&\\\hline
Population size $N$    &100    \\
Scaling factor $F$    &0.5     \\
Crossover probability $CR$    &0.9 \\
Index parameter $\eta_m$     &20    \\
Polynomial mutation probability $p_m$ &$\frac{1}{N_p}$\\\hline
Parameters for MOEA/D&\\\hline
Neighborhood size $T_m$     &10\\
Maximum replacement size $n_r$ & 2\\
The probability that parents from neighborhood $p_{\delta}$ &0.9\\\hline
Parameter for AUGMECON2&\\\hline
Number of grids& $100$\\
\hline
\end{tabular}
\end{table}

Furthermore, two performance metrics are applied, Hypervolume (HV)~\cite{zitzler1999multiobjective} and Generational Distance (GD)~\cite{van1999multiobjective}, both of them are easy to use in PlatEMO~\cite{tian2017platemo}. HV prefers heuristics because it takes the diversity of solutions into account. Heuristics, like EAs, always pay much attention to the diversity of solutions while exact methods hardly do it~\cite{miettinen2012nonlinear,ishibuchi2018specify}. It is given as
\begin{equation}
HV = {\rm volume}(\cup_{i=1}^{|Q|}hc_i),
\end{equation}where $Q$ is the set of obtained solutions of algorithms, and $hc_i$ is the hypercube bounded by solution $i$ and the reference point $r$. Better solutions lead to higher HV. On the other hand, GD evaluates the convergence ability of algorithms. In this problem, it tends to prefer exact mathematical programming approaches since exact methods are going to find the optimum if possible yet heuristics will search for approximations. It is defined as 
\begin{equation}
GD=\frac{\sqrt{\sum_i^Q{d_i}}}{|Q|},
\end{equation}where $Q$ is the set of obtained solutions of algorithms and $d_i$ is the shortest Euclidean distance among solution $i$ and the representatives of PF. Better solutions lead to lower GD.

Note that the implementations of HV and GD require the true PFs or reference points. The PFs are obtained by implementing AUGMECON2\&CPLEX on the original model with 2000 grids. However, they are unavailable while running the exact method after 7 days for some large-scale problems in this work. For these cases, the best known unconstrained Pareto fronts (UCPFs) are considered with the following modification. They are truncated by the highest return solutions (since the function of return is linear, it is easy to be solved). This truncated version of the UCPF is referred to as TUCPF. TUCPFs are used for instances including more than 225 assets. Concerning the reference point, it is recommended to set $r$ slightly dominated by the nadir point of the true PF or TUCPF, specifically, $r$ is set as $(1.1,1.1)$.

\subsection{Simulation Experiments}
This subsection comprises three sets of simulation experiments. The first two are about two different stop criteria and the third one is about applying the proposed method to the real stock markets in China.

We put the conclusion in advance since some readers are probably concerned about the conclusion yet do not have the patience to go through the analysis. Concisely, with maximum fitness evaluations as 1000, D\&O-MOEA/D shows prominent superiority. With a 2000 seconds time budget, AUGMECON2\&CPLEX is the best method for small-scale problems while D\&O-MOEA/D is the best method if all 20 instances are involved. Furthermore, the application for Chinese stock markets also demonstrates the feasibility and potential of D\&O-MOEA/D.

\subsubsection{Maximum Number of Fitness Evaluations}
\begin{figure*}[htbp]\footnotesize
\graphicspath{{figs/}}
\centering
\begin{minipage}{0.24\linewidth}
\centerline{\includegraphics[width=1\columnwidth]{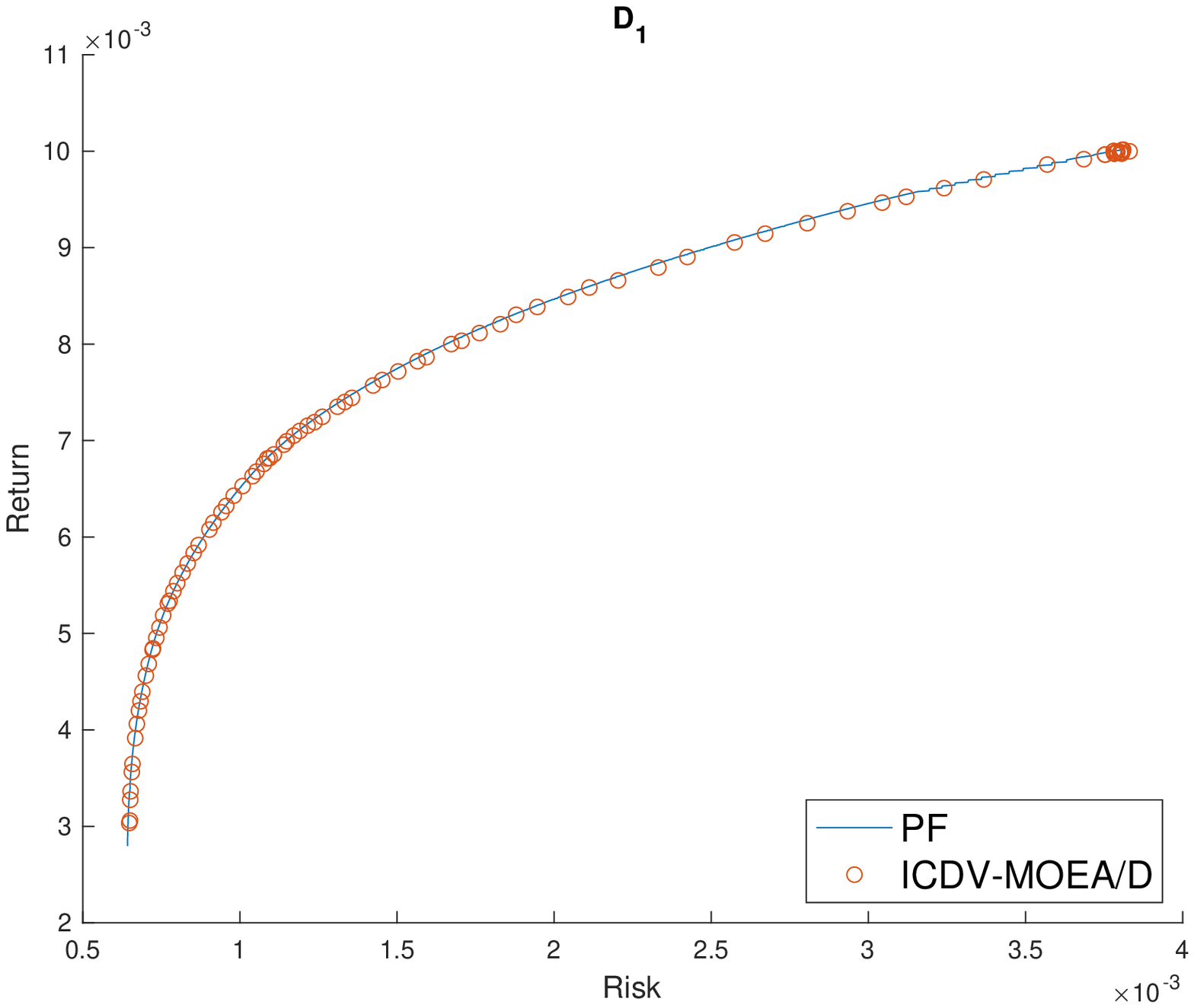}}
\centerline{(a)}
\end{minipage}
\begin{minipage}{0.24\linewidth}
\centerline{\includegraphics[width=1\columnwidth]{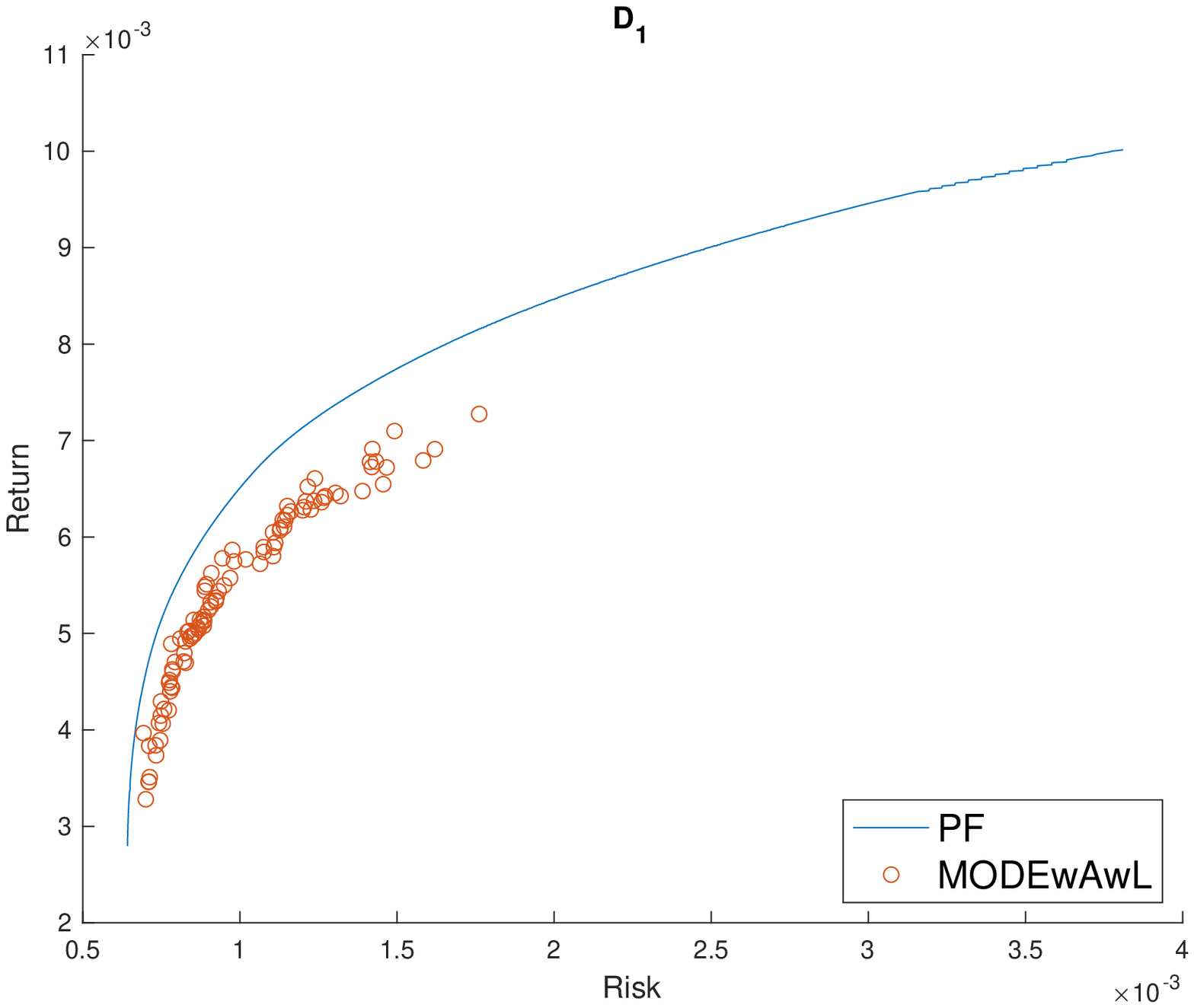}}
\centerline{(b)}
\end{minipage}
\begin{minipage}{0.24\linewidth}
\centerline{\includegraphics[width=1\columnwidth]{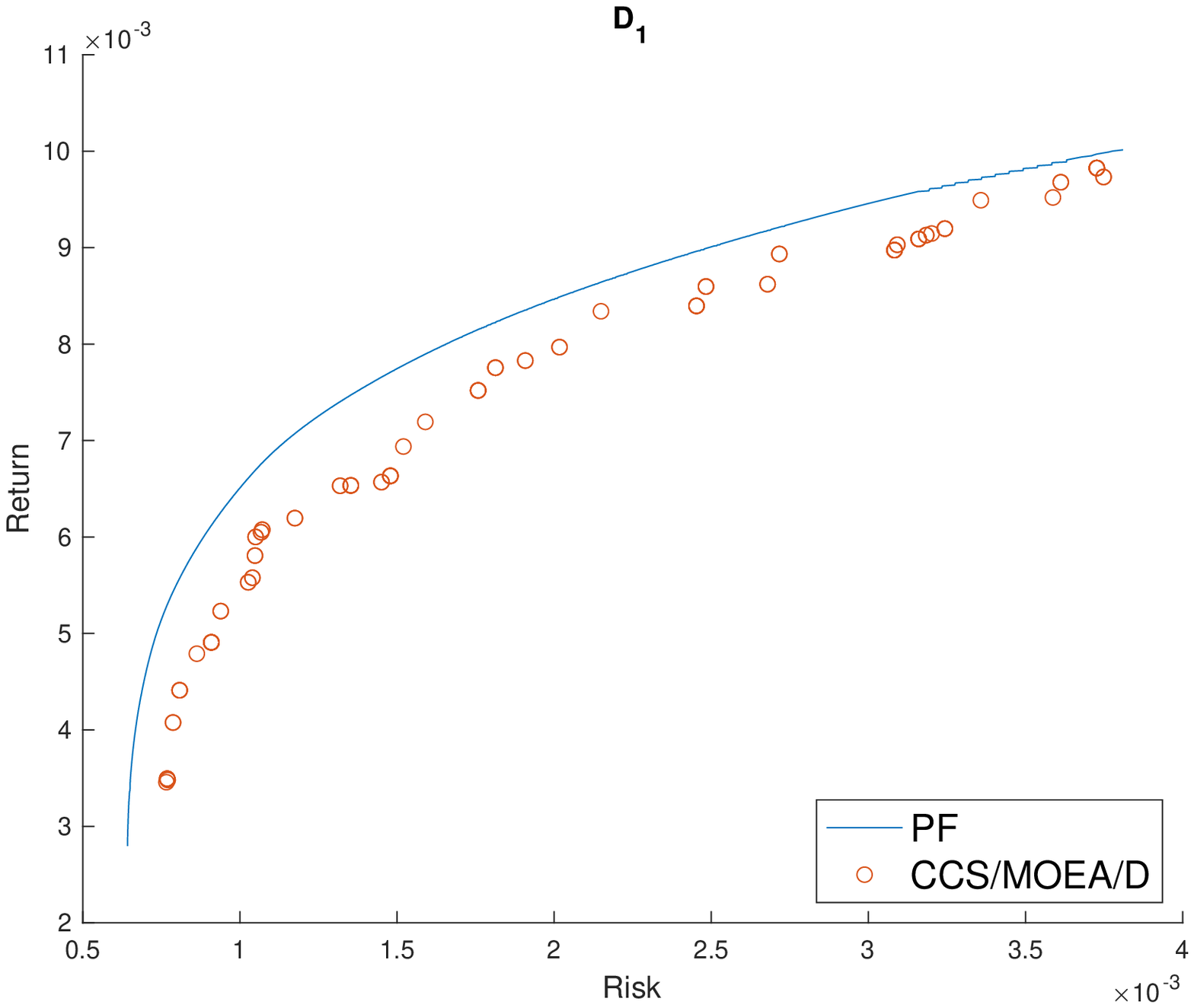}}
\centerline{(c)}
\end{minipage}
 \caption{The efficient fronts with the best HVs obtained by D\&O-MOEA/D, MODEwAwL and CCS/MOEA/D with 1000 times of fitness evaluations.} 
 \label{FES_each_fronts}
\end{figure*}

First, D\&O-MOEA/D, MODEwAwL, and CCS-MOEA/D are implemented and 1000 times fitness evaluations are set as the stop criterion. For these two EAs, the function evaluation is the fitness evaluation. Notwithstanding, for hybrid methods about EAs, sometimes function evaluations are called many times, but one time of fitness evaluation is counted when a final solution is obtained~\cite{lu2019nsga,lu2020nsganetv2,elsken2018neural}. In this case, the results obtained are shown in Table~\ref{FES_HV}. The rank of each algorithm on all instances is listed in parenthesis, and the total and final ranks of every algorithm are aggregated at the bottom. Furthermore, the symbol ``$+,-,\thickapprox$" indicates the corresponding algorithm is significantly better than, worse than and similar to D\&O-MOEA/D in terms of Wilcoxon rank-sum test at 5\% significant level. For brevity, `A', `B' and `C' represent the methods D\&O-MOEA/D, MODEwAwL and CCS-MOEA/D respectively. From $D_1$ to $D_{12}$, the PFs are obtained by the exact method. As for $D_{13}$ to $D_{20}$, the same method does not get results after 7 days. The demanded computational resources will exponentially grow when the number of assets increases since these problems are NP-hard~\cite{moral2006selection}. TUCPFs are hence applied. Table~\ref{FES_HV} shows D\&O-MOEA/D outperforms two EAs dramatically. It wins all the first places on 20 instances. Furthermore, the superiority of D\&O-MOEA/D is still significant regarding the obtained solutions in the objective spaces. Fig.~\ref{FES_each_fronts} only illustrates the obtained fronts of every method on $D_{1}$ due to the space limit. Solutions from D\&O-MOEA/D almost lie on the PF while the solutions of the other two are unsatisfactory. In this case, one is reasonable to comment that although these methods take the same times of fitness evaluations, D\&O-MOEA/D takes much more computational resources in fact. More simulation experiments are hence carried out.
\begin{table}\scriptsize
\caption{Results on $D_1-D_{20}$ concerning HV with 1000 times of fitness evaluations.}
\label{FES_HV}
\begin{tabular}{|ccccc|}\hline
\multicolumn{2}{|c}{Algo.}&\multicolumn{1}{c}{A}&\multicolumn{1}{c}{B}&\multicolumn{1}{c|}{C}\\\hline
\multirow{2}{*}{$D_{1}$}
&Mean	&\cellcolor{gray25}8.03e-01[1]	&5.67e-01[3]	&7.00e-01[2]	\\
&Std	&9.42e-04	&2.79e-02	&5.60e-02	\\\hline
\multirow{2}{*}{$D_{2}$}
&Mean	&\cellcolor{gray25}8.93e-01[1]	&6.90e-01[2]	&6.24e-01[3]	\\
&Std	&3.23e-03	&3.60e-02	&1.32e-01	\\\hline
\multirow{2}{*}{$D_{3}$}
&Mean	&\cellcolor{gray25}8.05e-01[1]	&6.71e-01[2]	&5.23e-01[3]	\\
&Std	&2.50e-03	&1.63e-02	&1.15e-01	\\\hline
\multirow{2}{*}{$D_{4}$}
&Mean	&\cellcolor{gray25}9.20e-01[1]	&8.24e-01[2]	&6.63e-01[3]	\\
&Std	&5.33e-03	&1.74e-02	&1.09e-01	\\\hline
\multirow{2}{*}{$D_{5}$}
&Mean	&\cellcolor{gray25}8.49e-01[1]	&6.88e-01[2]	&5.68e-01[3]	\\
&Std	&4.05e-03	&2.18e-02	&1.06e-01	\\\hline
\multirow{2}{*}{$D_{6}$}
&Mean	&\cellcolor{gray25}8.58e-01[1]	&7.74e-01[2]	&5.79e-01[3]	\\
&Std	&2.40e-03	&2.23e-02	&8.89e-02	\\\hline
\multirow{2}{*}{$D_{7}$}
&Mean	&\cellcolor{gray25}7.60e-01[1]	&4.70e-01[3]	&6.09e-01[2]	\\
&Std	&4.80e-04	&5.43e-02	&1.31e-01	\\\hline
\multirow{2}{*}{$D_{8}$}
&Mean	&\cellcolor{gray25}8.23e-01[1]	&6.45e-01[2]	&5.87e-01[3]	\\
&Std	&3.86e-03	&1.96e-02	&1.11e-01	\\\hline
\multirow{2}{*}{$D_{9}$}
&Mean	&\cellcolor{gray25}8.24e-01[1]	&5.91e-01[2]	&3.94e-01[3]	\\
&Std	&2.27e-03	&2.67e-02	&9.57e-02	\\\hline
\multirow{2}{*}{$D_{10}$}
&Mean	&\cellcolor{gray25}8.90e-01[1]	&8.12e-01[2]	&4.06e-01[3]	\\
&Std	&8.52e-03	&1.65e-02	&8.80e-02	\\\hline
\multirow{2}{*}{$D_{11}$}
&Mean	&\cellcolor{gray25}8.03e-01[1]	&6.26e-01[2]	&4.19e-01[3]	\\
&Std	&2.86e-03	&2.00e-02	&8.64e-02	\\\hline
\multirow{2}{*}{$D_{12}$}
&Mean	&\cellcolor{gray25}8.36e-01[1]	&8.07e-01[2]	&3.40e-01[3]	\\
&Std	&9.14e-03	&9.55e-03	&1.09e-01	\\\hline
\multirow{2}{*}{$D_{13}$}
&Mean	&\cellcolor{gray25}8.65e-01[1]	&7.11e-01[2]	&5.60e-01[3]	\\
&Std	&4.27e-03	&3.13e-02	&1.24e-01	\\\hline
\multirow{2}{*}{$D_{14}$}
&Mean	&\cellcolor{gray25}8.28e-01[1]	&6.49e-01[2]	&5.09e-01[3]	\\
&Std	&1.47e-03	&3.33e-02	&1.30e-01	\\\hline
\multirow{2}{*}{$D_{15}$}
&Mean	&\cellcolor{gray25}8.07e-01[1]	&6.41e-01[2]	&4.25e-01[3]	\\
&Std	&2.54e-03	&2.18e-02	&8.48e-02	\\\hline
\multirow{2}{*}{$D_{16}$}
&Mean	&\cellcolor{gray25}9.04e-01[1]	&8.63e-01[2]	&5.76e-01[3]	\\
&Std	&4.83e-03	&4.65e-03	&7.51e-02	\\\hline
\multirow{2}{*}{$D_{17}$}
&Mean	&\cellcolor{gray25}7.65e-01[1]	&5.36e-01[2]	&4.35e-01[3]	\\
&Std	&1.56e-03	&4.16e-02	&1.33e-01	\\\hline
\multirow{2}{*}{$D_{18}$}
&Mean	&\cellcolor{gray25}8.63e-01[1]	&7.13e-01[2]	&5.44e-01[3]	\\
&Std	&2.37e-03	&2.34e-02	&7.59e-02	\\\hline
\multirow{2}{*}{$D_{19}$}
&Mean	&\cellcolor{gray25}7.29e-01[1]	&4.48e-01[2]	&3.63e-01[3]	\\
&Std	&4.03e-04	&4.84e-02	&1.69e-01	\\\hline
\multirow{2}{*}{$D_{20}$}
&Mean	&\cellcolor{gray25}8.57e-01[1]	&6.89e-01[2]	&4.86e-01[3]	\\
&Std	&3.42e-03	&2.68e-02	&4.48e-02	\\\hline
\multicolumn{2}{|c}{Total$^*$}
&20	&42	&58	\\\hline
\multicolumn{2}{|c}{Final Rank$^*$}
&1	&2	&3	\\\hline
\multicolumn{2}{|c}{$+,-,\thickapprox$}
&-	&0/20/0	&0/20/0	\\\hline
\end{tabular}
\end{table}

\subsubsection{Maximum Running Time}
\begin{figure}[htbp]\footnotesize
\graphicspath{{figs/}}
\centering
\centerline{\includegraphics[width=0.75\columnwidth]{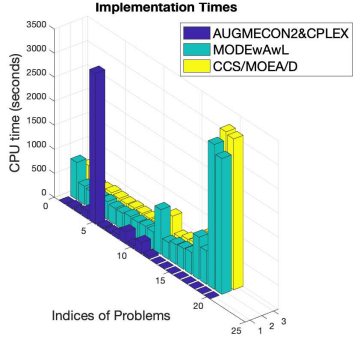}}
 \caption{The implementation times for every methods on all problems.} 
 \label{est_time}
\end{figure}

Second, a time budget should be an alternative fair stop criterion for these methods, although it is hardly used in the realm of EAs. As in~\cite{2014KhinMODEwawl,DBLP:journals/corr/abs-1909-08748}, maximum fitness evaluations for these two EAs are set as 10e5. Therefore, we reimplement the two algorithms independently 20 times on all instances to estimate the required times. Further, AUGMECON2\&CPLEX with 100 grids is also carried out while the time budget is 1 day. Fig.~\ref{est_time} shows the average times of three methods on every problem. The running times of the two EAs are similar and most of them vary from 500 to 1000 seconds. For the largest two instances, they are about 3000 seconds. Regarding, the exact method, we can only get the solutions for $D_{1}$ to $D_{12}$ in 1 day. 65.27 seconds is the median of the running times among 12 problems. The running time on $D_5$ is 3136.59 seconds, it should be a special case that the data of $D_5$ slows down the convergence of the used quadratic optimizer. In summary, a 2000 seconds execution time is set as the stop criterion for the EAs and D\&O-MOEA/D, with which most of them can converge. Then AUGMECON2\&CPLEX is completely executed on the first 12 instances.

The statistics of HV and GD obtained by these four algorithms are shown in Tables~\ref{HV_table} and~\ref{GD_table}. There are some differences from Tables~\ref{HV_table} and~\ref{GD_table} to Table~\ref{FES_HV}. Firstly, AUGMECON2\&CPLEX is added and is represented by `D'. Two sets of statistical indicators such as `Total$^*$' and `Total' are used, because the counts are made twice since AUGMECON2\&CPLEX with 100 grids can only be completed on problems including no more 225 assets with the time budget of 1 day. The statistical indicators with a star (*) indicate the results based on 4 algorithms for 12 instances, and the latter ones indicate the results based on 3 algorithms for 20 instances. All the comparisons consist of two parts: (i) the performance of all methods on $D_1$ to $D_{12}$ and (ii) the results of two EAs and D\&O-MOEA/D for all problems. 

\begin{figure*}[htbp]\footnotesize
\graphicspath{{figs/}}
\centering
\begin{minipage}{0.24\linewidth}
\centerline{\includegraphics[width=1\columnwidth]{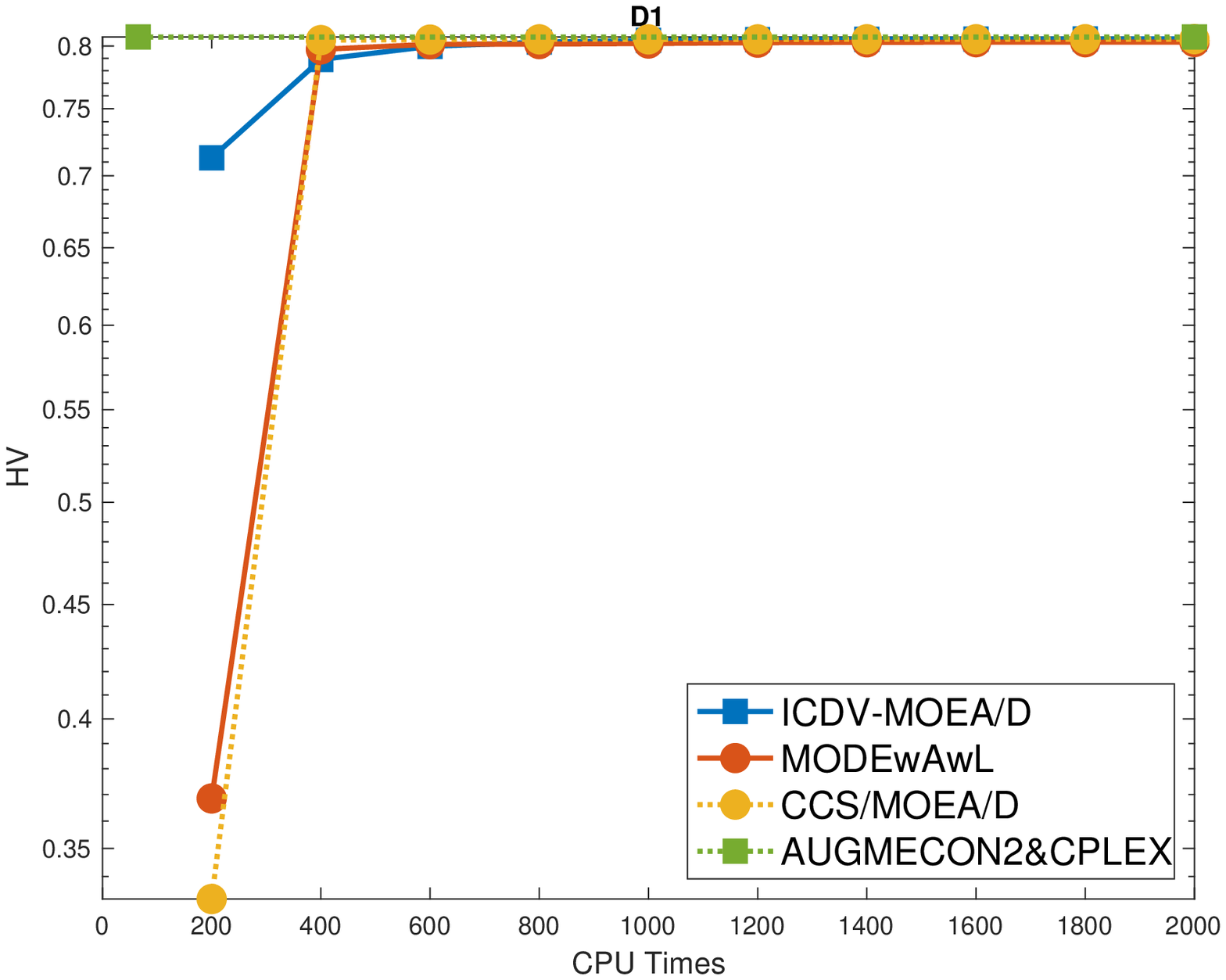}}
\centerline{(a)}
\end{minipage}
\begin{minipage}{0.24\linewidth}
\centerline{\includegraphics[width=1\columnwidth]{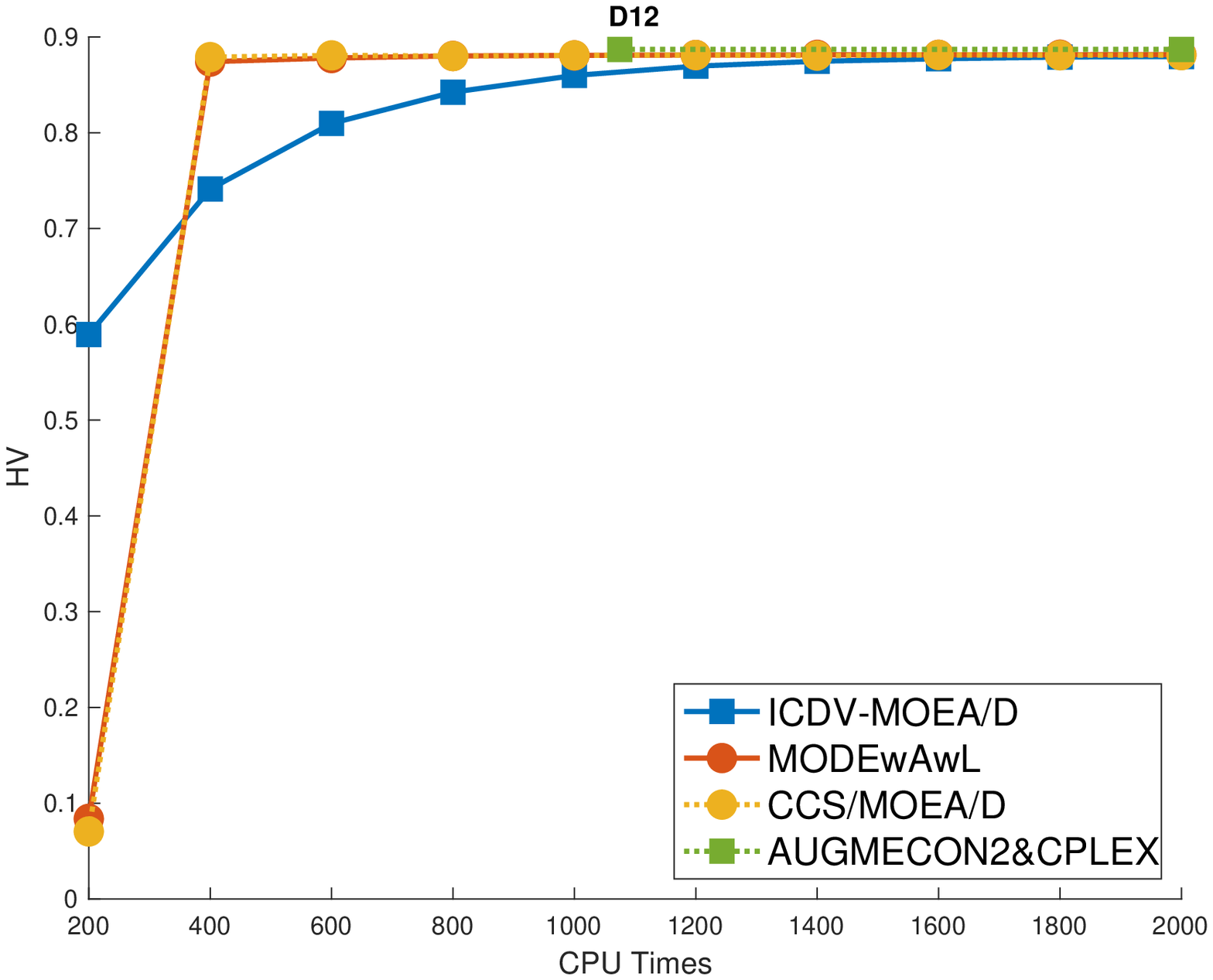}}
\centerline{(b)}
\end{minipage}
\begin{minipage}{0.24\linewidth}
\centerline{\includegraphics[width=1\columnwidth]{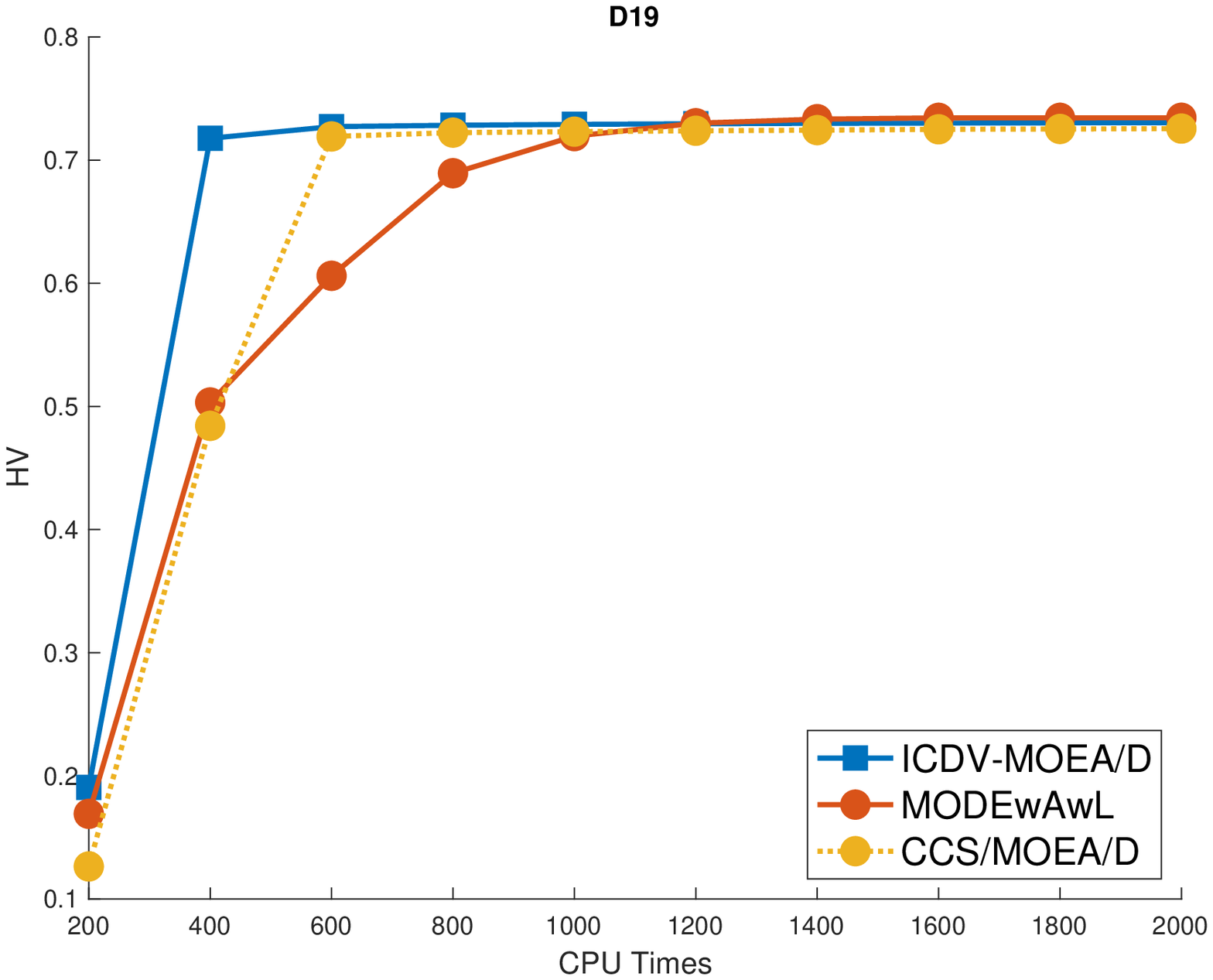}}
\centerline{(c)}
\end{minipage}
\begin{minipage}{0.24\linewidth}
\centerline{\includegraphics[width=1\columnwidth]{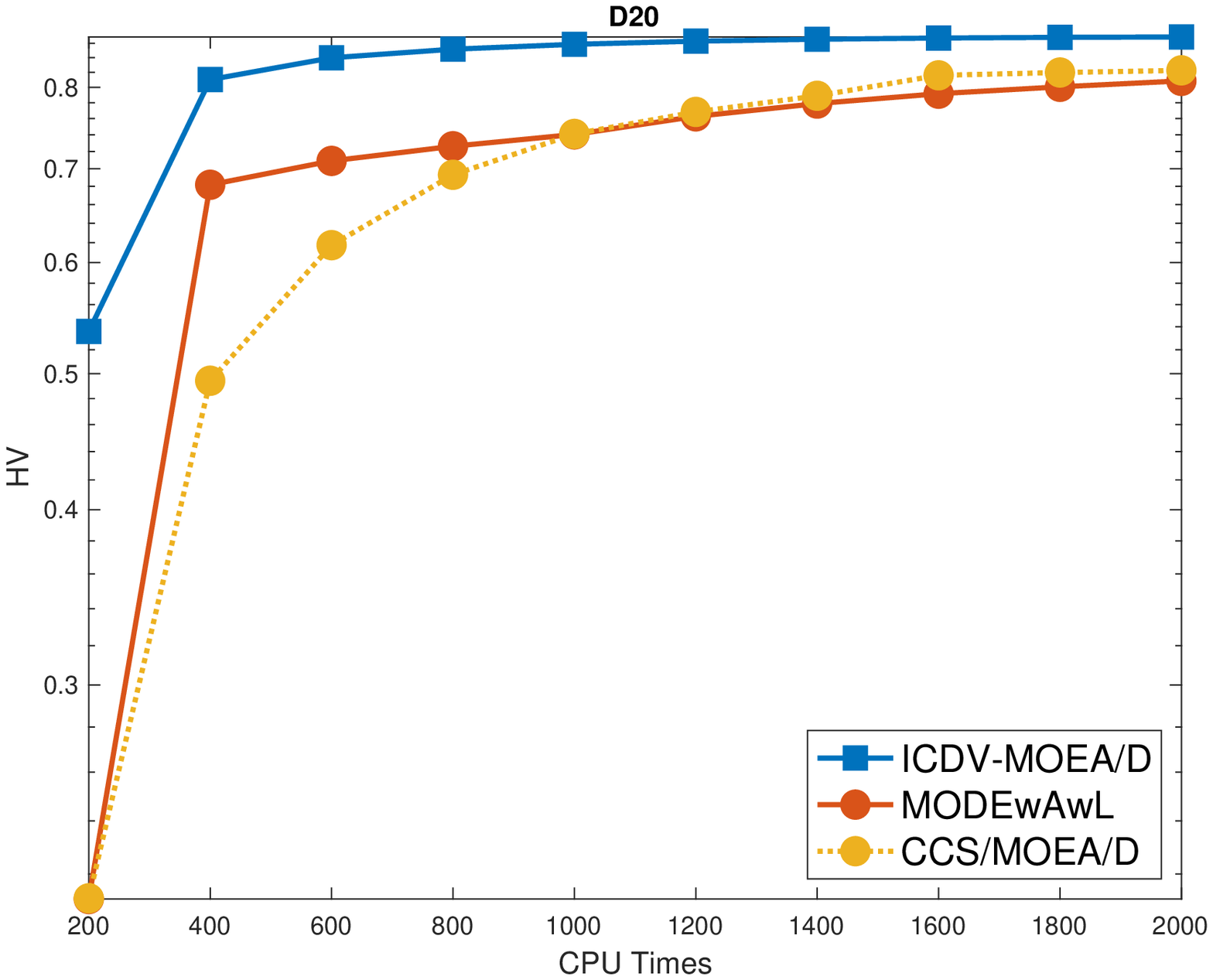}}
\centerline{(d)}
\end{minipage}
\begin{minipage}{0.24\linewidth}
\centerline{\includegraphics[width=1\columnwidth]{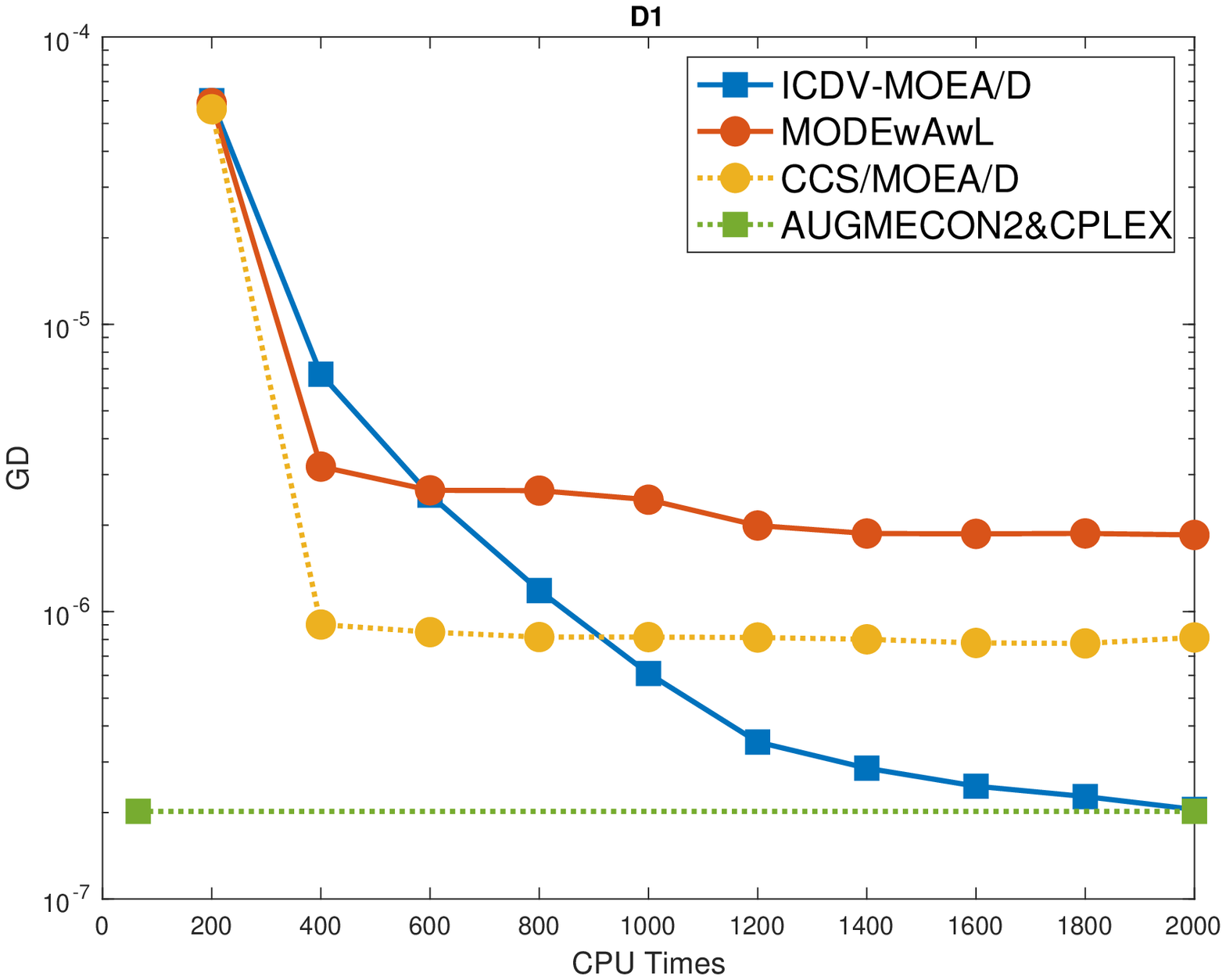}}
\centerline{(e)}
\end{minipage}
\begin{minipage}{0.24\linewidth}
\centerline{\includegraphics[width=1\columnwidth]{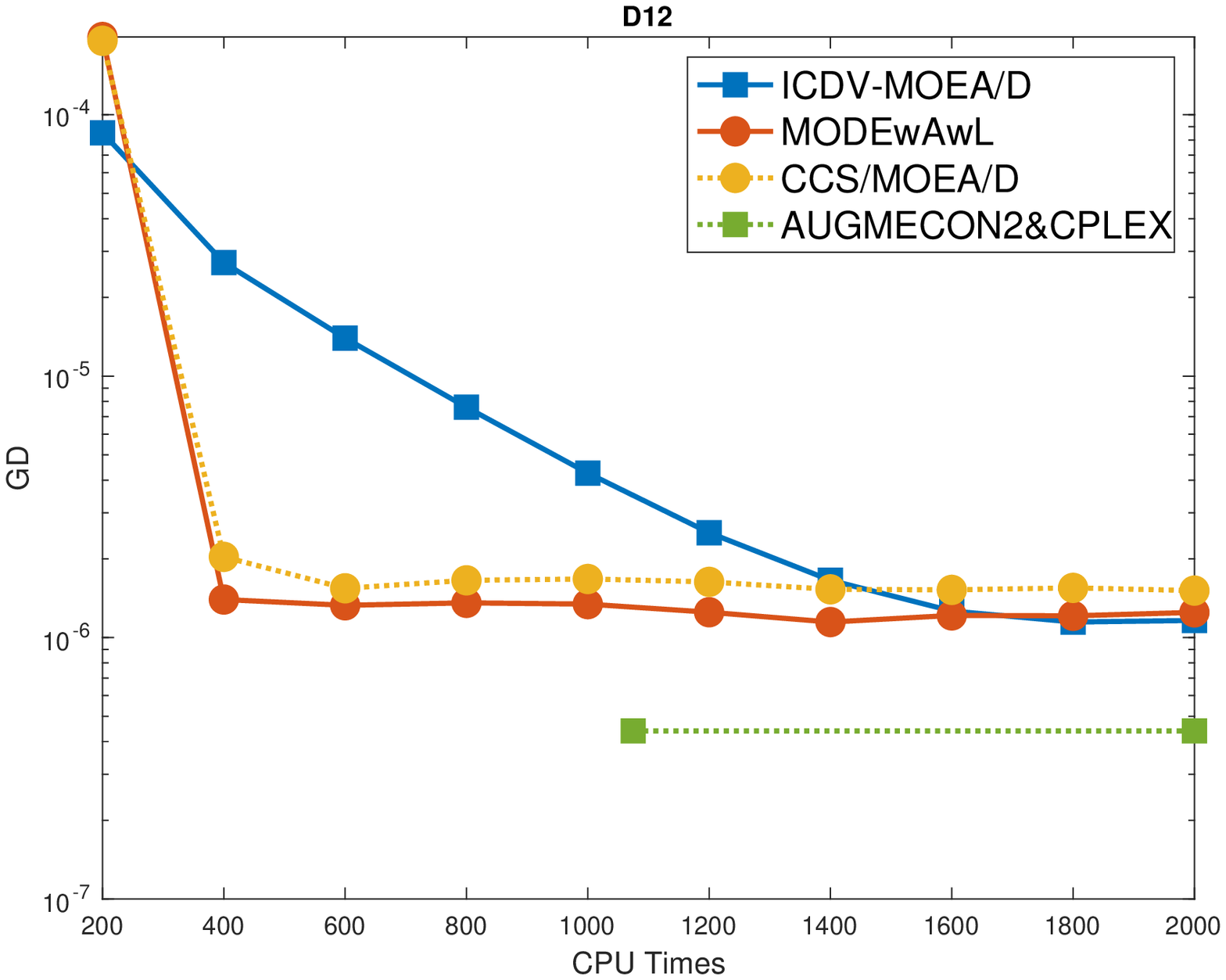}}
\centerline{(f)}
\end{minipage}
\begin{minipage}{0.24\linewidth}
\centerline{\includegraphics[width=1\columnwidth]{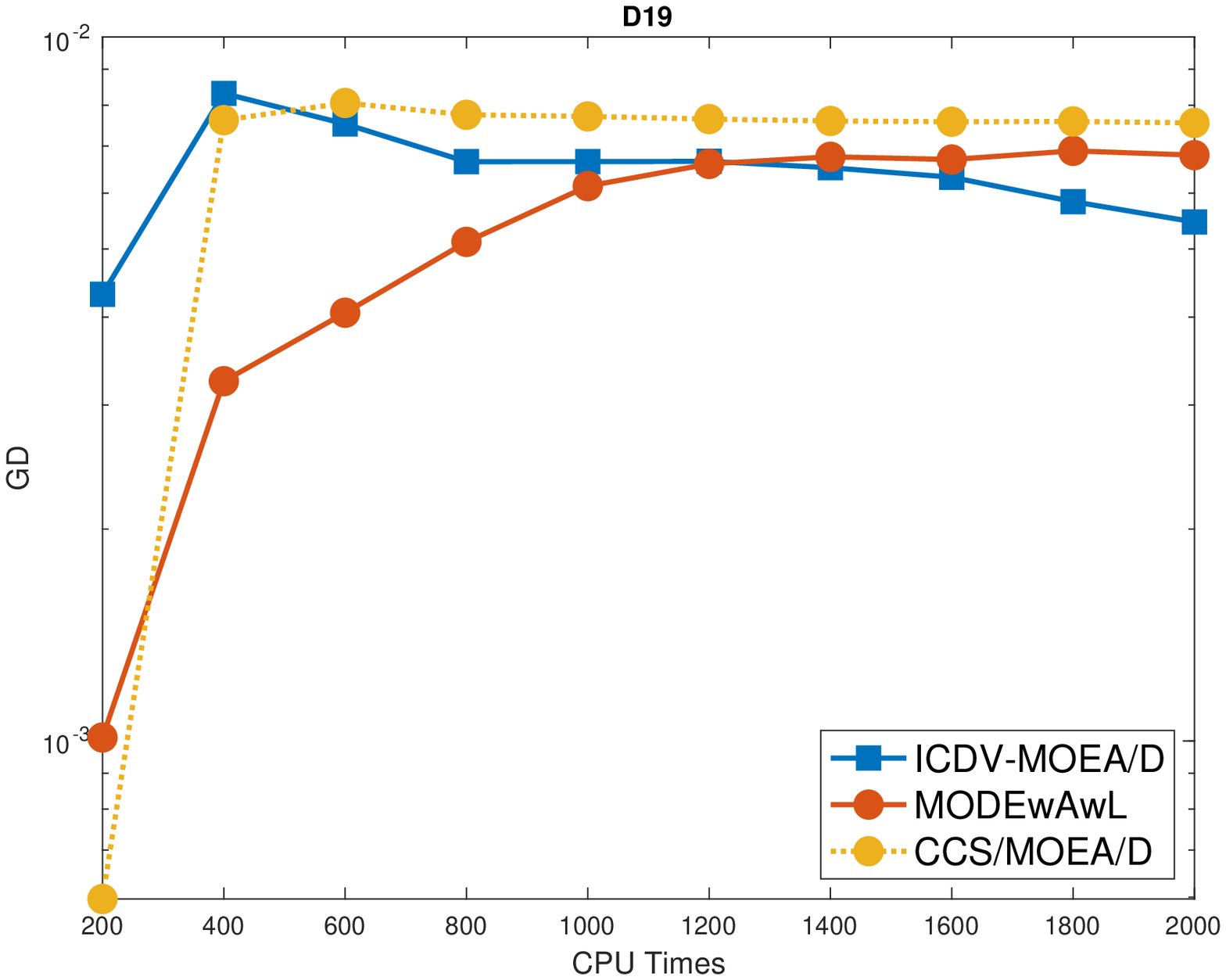}}
\centerline{(g)}
\end{minipage}
\begin{minipage}{0.24\linewidth}
\centerline{\includegraphics[width=1\columnwidth]{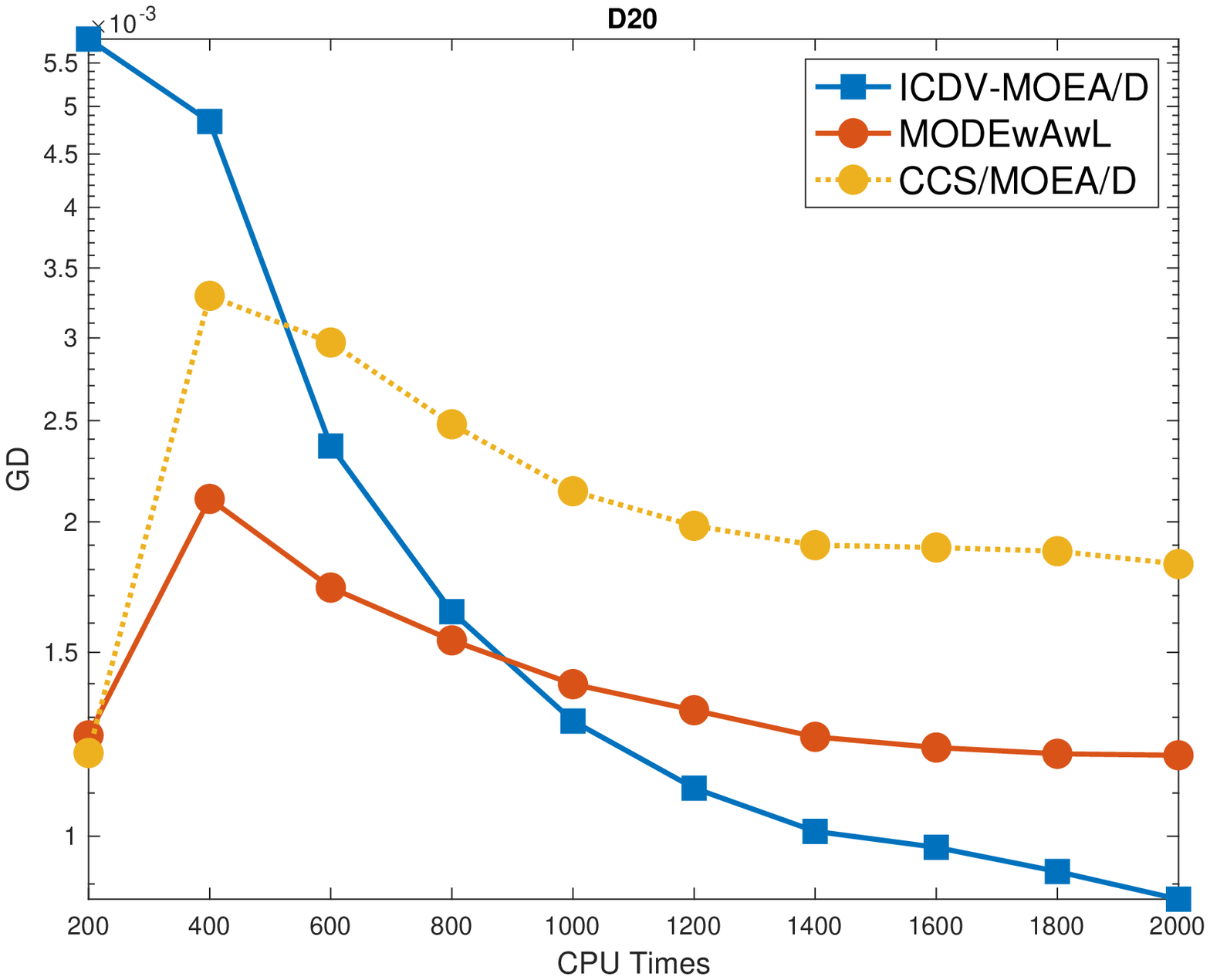}}
\centerline{(h)}
\end{minipage}
 \caption{The convergence curves of every method on different problems.} 
 \label{HV_convergence}
\end{figure*}

In terms of HV, Table~\ref{HV_table} illustrates AUGMECON2\&CPLEX is undoubtedly the best method for small instances, from $D_1$ to $D_{12}$. It gets the smallest HVs on 11 problems and has a very small gap with MODEwAwL on $D_7$, HVs of both of them are 7.65e-01. D\&O-MOEA/D outperforms two EAs slightly with regards to the total ranks on these 12 instances, which are 34, 38 and 35 respectively. Concerning the Wilcoxon rank-sum test, the dominance of D\&O-MOEA/D is also not obvious. Although the statistic of MODEwAwL is `2/9/1', it of CCS-MOEA/D is `5/4/3'. On problems including more than 225 assets, AUGMECON2\&CPLEX can not be completed in 1 day. Notwithstanding, concerning the Wilcoxon rank-sum test, the superiority of D\&O-MOEA/D becomes prominent when results on all 20 instances are included. These statistics of MODEwAwL and CCS-MOEA/D are `7/12/1' and `6/10/4'. They indicate D\&O-MOEA/D wins on more problems than the compared peer algorithms do. Fig~\ref{HV_convergence} illustrates the convergence curves of every method on $D_1$, $D_{12}$, $D_{19}$ and $D_{20}$. $D_1$ and $D_{12}$ are solvable for AUGMECON2\&CPLEX, so its HVs are also drawn. The HV lines start when AUGMECON2\&CPLEX finishes. In Figs.~\ref{HV_convergence} (a) and~\ref{HV_convergence} (b), the results imply AUGMECON2\&CPLEX is the best choice among these four methods for small-scale problems, it converges fast and well. Since the exact method is sufficient, so we do not need heuristics like D\&O-MOEA/D and the EAs for small-scale problems. In Fig.~\ref{HV_convergence} (c) and~\ref{HV_convergence} (d), D\&O-MOEA/D converges faster and the final HV of it is much better than the others on $D_{20}$. It implies that D\&O-MOEA/D is probably a good method for large-scale problems.
\begin{table}\scriptsize
\caption{Results on $D_1-D_{20}$ concerning HV with executions for 2000 seconds.}
\label{HV_table}
\begin{tabular}{|cccccc|}\hline
\multicolumn{2}{|c}{Algo.}&\multicolumn{1}{c}{A}&\multicolumn{1}{c}{B}&\multicolumn{1}{c}{C}&\multicolumn{1}{c|}{D}\\\hline
\multirow{2}{*}{$D_{1}$}
&Mean	&8.06e-01[2]	&8.03e-01[4]	&8.06e-01[3]	&\cellcolor{gray25}\\
&Std	&1.63e-04	&8.65e-04	&6.41e-04	&\multirow{-2}{*}{\cellcolor{gray25}8.08e-01[1]	}\\\hline
\multirow{2}{*}{$D_{2}$}
&Mean	&9.03e-01[2]	&9.00e-01[3]	&8.99e-01[4]	&\cellcolor{gray25}\\
&Std	&6.12e-04	&3.23e-03	&2.10e-03	&\multirow{-2}{*}{\cellcolor{gray25}9.05e-01[1]	}\\\hline
\multirow{2}{*}{$D_{3}$}
&Mean	&8.16e-01[3]	&8.14e-01[4]	&8.18e-01[2]	&\cellcolor{gray25}\\
&Std	&1.41e-03	&5.49e-04	&9.05e-04	&\multirow{-2}{*}{\cellcolor{gray25}8.21e-01[1]	}\\\hline
\multirow{2}{*}{$D_{4}$}
&Mean	&9.38e-01[2]	&9.36e-01[3]	&9.28e-01[4]	&\cellcolor{gray25}\\
&Std	&1.01e-03	&8.20e-04	&4.05e-03	&\multirow{-2}{*}{\cellcolor{gray25}9.39e-01[1]	}\\\hline
\multirow{2}{*}{$D_{5}$}
&Mean	&8.63e-01[3]	&8.60e-01[4]	&8.65e-01[2]	&\cellcolor{gray25}\\
&Std	&1.49e-03	&1.82e-03	&1.45e-03	&\multirow{-2}{*}{\cellcolor{gray25}8.68e-01[1]	}\\\hline
\multirow{2}{*}{$D_{6}$}
&Mean	&8.71e-01[3]	&8.66e-01[4]	&8.73e-01[2]	&\cellcolor{gray25}\\
&Std	&1.51e-03	&3.30e-03	&1.52e-03	&\multirow{-2}{*}{\cellcolor{gray25}8.75e-01[1]	}\\\hline
\multirow{2}{*}{$D_{7}$}
&Mean	&7.62e-01[4]	&\cellcolor{gray25}7.65e-01[1]	&7.63e-01[3]	&	\\
&Std	&4.95e-06	&3.10e-04	&1.31e-03	&\multirow{-2}{*}{7.65e-01[2]	}\\\hline
\multirow{2}{*}{$D_{8}$}
&Mean	&8.38e-01[3]	&8.37e-01[4]	&8.42e-01[2]	&\cellcolor{gray25}\\
&Std	&1.48e-03	&8.14e-04	&8.82e-04	&\multirow{-2}{*}{\cellcolor{gray25}8.44e-01[1]	}\\\hline
\multirow{2}{*}{$D_{9}$}
&Mean	&8.33e-01[3]	&8.33e-01[2]	&8.32e-01[4]	&\cellcolor{gray25}\\
&Std	&1.33e-04	&2.84e-04	&1.72e-03	&\multirow{-2}{*}{\cellcolor{gray25}8.34e-01[1]	}\\\hline
\multirow{2}{*}{$D_{10}$}
&Mean	&9.19e-01[2]	&9.17e-01[3]	&9.13e-01[4]	&\cellcolor{gray25}\\
&Std	&3.47e-03	&1.76e-03	&7.65e-03	&\multirow{-2}{*}{\cellcolor{gray25}9.25e-01[1]	}\\\hline
\multirow{2}{*}{$D_{11}$}
&Mean	&8.17e-01[3]	&8.16e-01[4]	&8.19e-01[2]	&\cellcolor{gray25}\\
&Std	&9.44e-04	&1.72e-03	&3.64e-03	&\multirow{-2}{*}{\cellcolor{gray25}8.23e-01[1]	}\\\hline
\multirow{2}{*}{$D_{12}$}
&Mean	&8.80e-01[4]	&8.82e-01[2]	&8.81e-01[3]	&\cellcolor{gray25}\\
&Std	&4.22e-03	&5.18e-04	&3.74e-03	&\multirow{-2}{*}{\cellcolor{gray25}8.87e-01[1]	}\\\hline
\multicolumn{2}{|c}{Total$^*$}
&34	&38	&35	&13	\\\hline
\multicolumn{2}{|c}{Final Rank$^*$}
&2	&4	&3	&1	\\\hline
\multicolumn{2}{|c}{$+,-,\thickapprox^*$}
&-	&2/9/1	&5/4/3	&12/0/0	\\\hline
\multirow{2}{*}{$D_{13}$}
&Mean	&8.79e-01[2]	&\cellcolor{gray25}8.80e-01[1]	&8.74e-01[3]	&	\\
&Std&6.40e-05	&1.73e-04	&2.63e-03	&\multirow{-2}{*}{-}	\\\hline
\multirow{2}{*}{$D_{14}$}
&Mean	&8.35e-01[2]	&\cellcolor{gray25}8.38e-01[1]	&8.32e-01[3]	&	\\
&Std&7.06e-05	&2.41e-04	&4.21e-03	&\multirow{-2}{*}{-}	\\\hline
\multirow{2}{*}{$D_{15}$}
&Mean	&8.23e-01[2]	&8.22e-01[3]	&\cellcolor{gray25}8.24e-01[1]	&	\\
&Std&7.81e-04	&9.82e-04	&3.85e-03	&\multirow{-2}{*}{-}	\\\hline
\multirow{2}{*}{$D_{16}$}
&Mean	&9.26e-01[3]	&\cellcolor{gray25}9.29e-01[1]	&9.27e-01[2]	&	\\
&Std&1.45e-03	&1.32e-03	&3.55e-03	&\multirow{-2}{*}{-}	\\\hline
\multirow{2}{*}{$D_{17}$}
&Mean	&7.70e-01[2]	&\cellcolor{gray25}7.70e-01[1]	&7.64e-01[3]	&	\\
&Std&4.64e-06	&8.77e-04	&3.54e-03	&\multirow{-2}{*}{-}	\\\hline
\multirow{2}{*}{$D_{18}$}
&Mean	&\cellcolor{gray25}8.77e-01[1]	&8.73e-01[2]	&8.73e-01[3]	&	\\
&Std&4.98e-04	&5.36e-03	&1.38e-03	&\multirow{-2}{*}{-}	\\\hline
\multirow{2}{*}{$D_{19}$}
&Mean	&7.30e-01[2]	&\cellcolor{gray25}7.34e-01[1]	&7.26e-01[3]	&	\\
&Std&4.11e-05	&4.86e-04	&2.95e-03	&\multirow{-2}{*}{-}	\\\hline
\multirow{2}{*}{$D_{20}$}
&Mean	&\cellcolor{gray25}8.69e-01[1]	&8.08e-01[3]	&8.23e-01[2]	&	\\
&Std&4.91e-04	&3.44e-02	&5.55e-02	&\multirow{-2}{*}{-}	\\\hline
\multicolumn{2}{|c}{Total}
&37	&40	&43	&-	\\\hline
\multicolumn{2}{|c}{Final Rank}
&1	&2	&3	&-	\\\hline
\multicolumn{2}{|c}{$+,-,\thickapprox$}
&-	&7/12/1	&6/10/4	&-	\\\hline
\end{tabular}
\end{table}

\begin{figure*}[htbp]\footnotesize
\graphicspath{{figs/}}
\centering
\begin{minipage}{0.24\linewidth}
\centerline{\includegraphics[width=1\columnwidth]{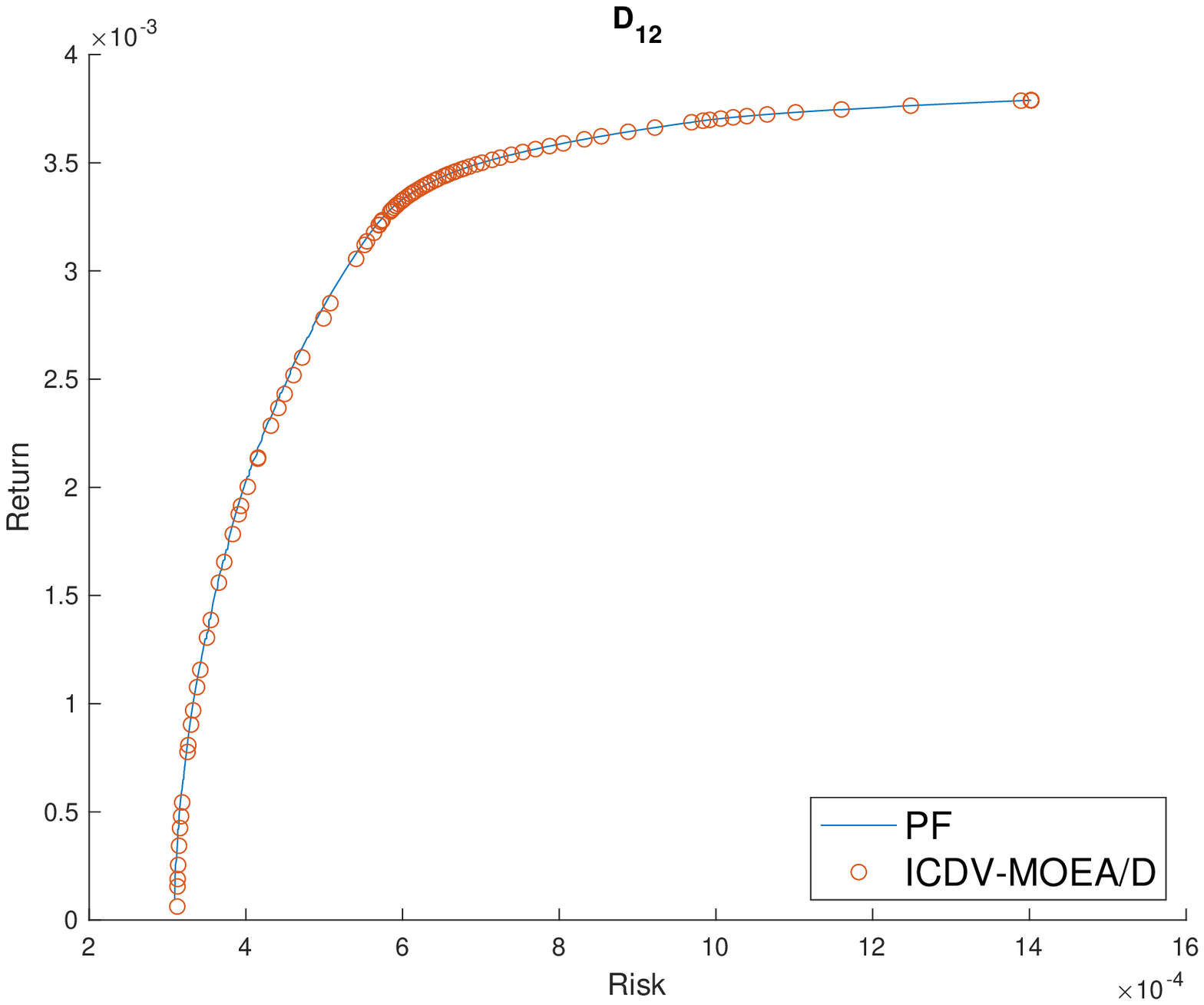}}
\centerline{(a)}
\end{minipage}
\begin{minipage}{0.24\linewidth}
\centerline{\includegraphics[width=1\columnwidth]{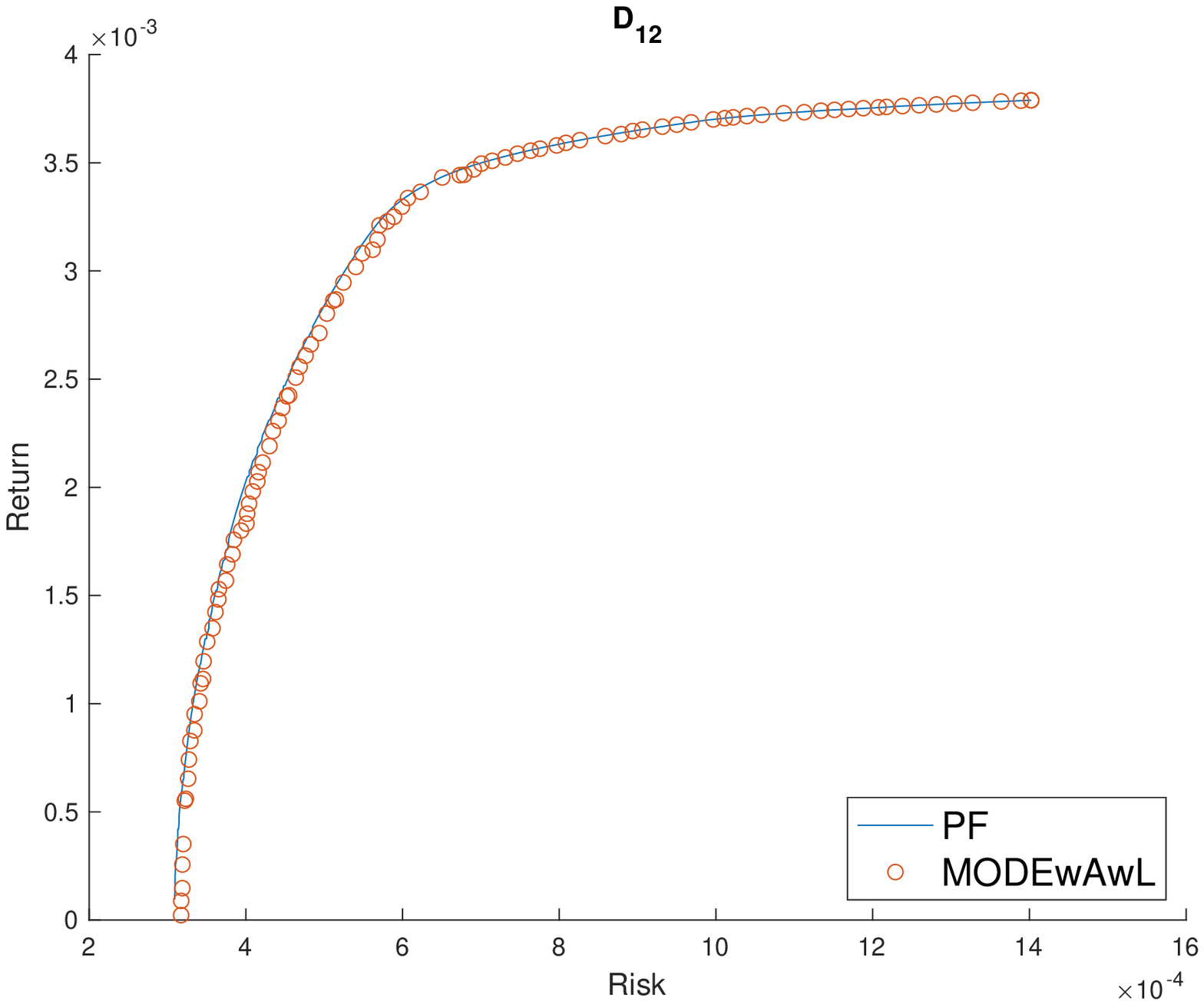}}
\centerline{(b)}
\end{minipage}
\begin{minipage}{0.24\linewidth}
\centerline{\includegraphics[width=1\columnwidth]{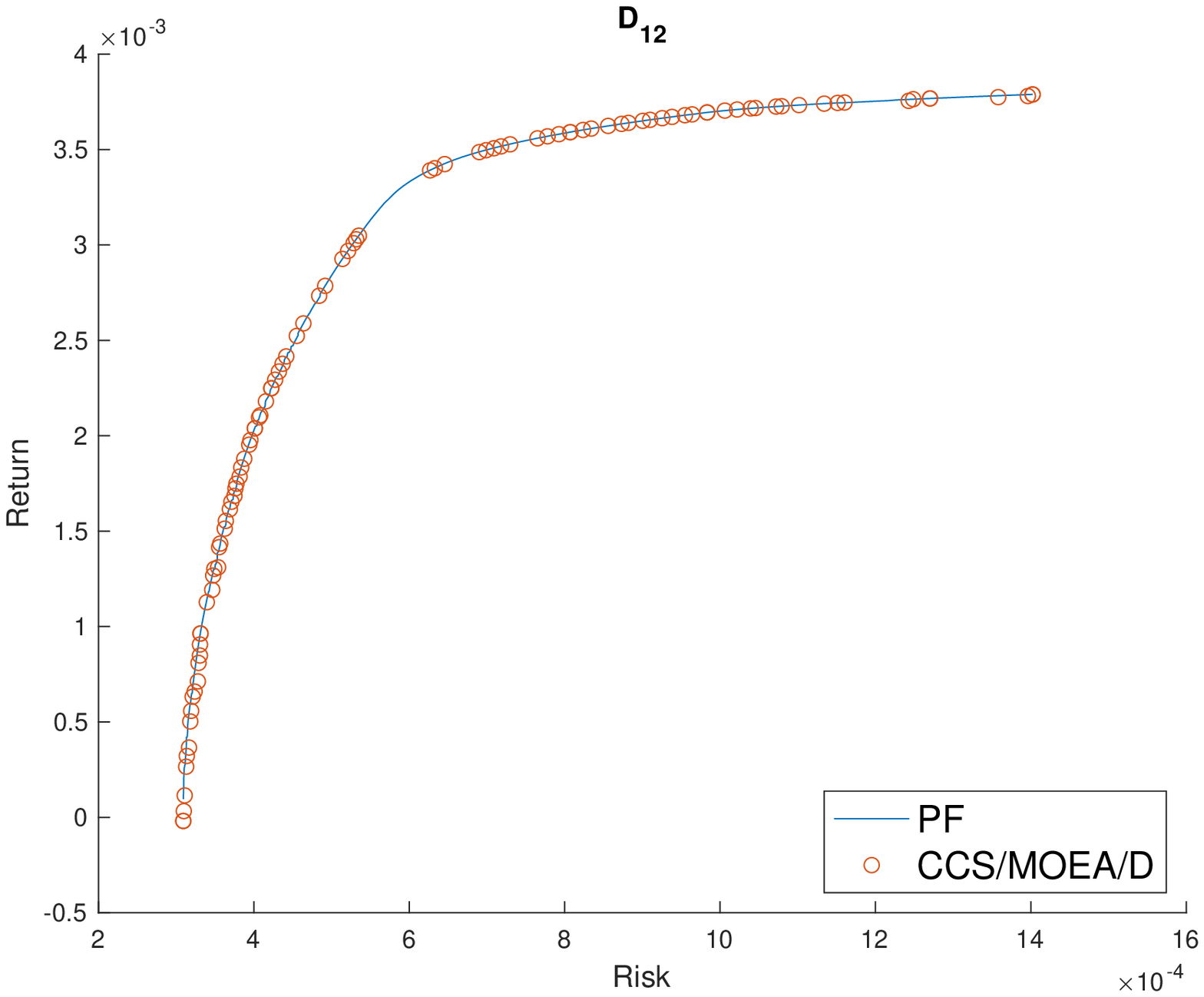}}
\centerline{(c)}
\end{minipage}
\begin{minipage}{0.24\linewidth}
\centerline{\includegraphics[width=1\columnwidth]{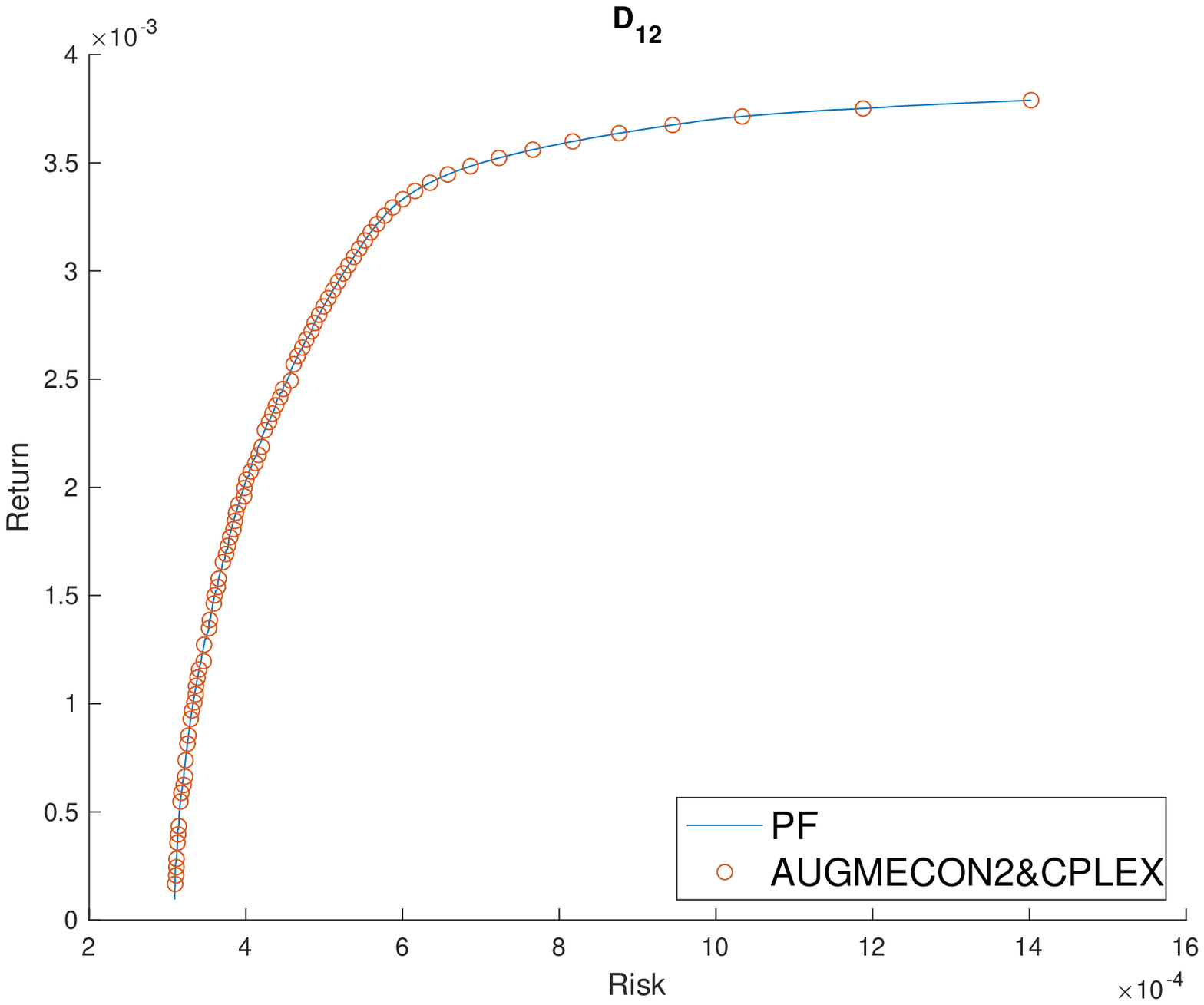}}
\centerline{(d)}
\end{minipage}\\
\begin{minipage}{0.24\linewidth}
\centerline{\includegraphics[width=1\columnwidth]{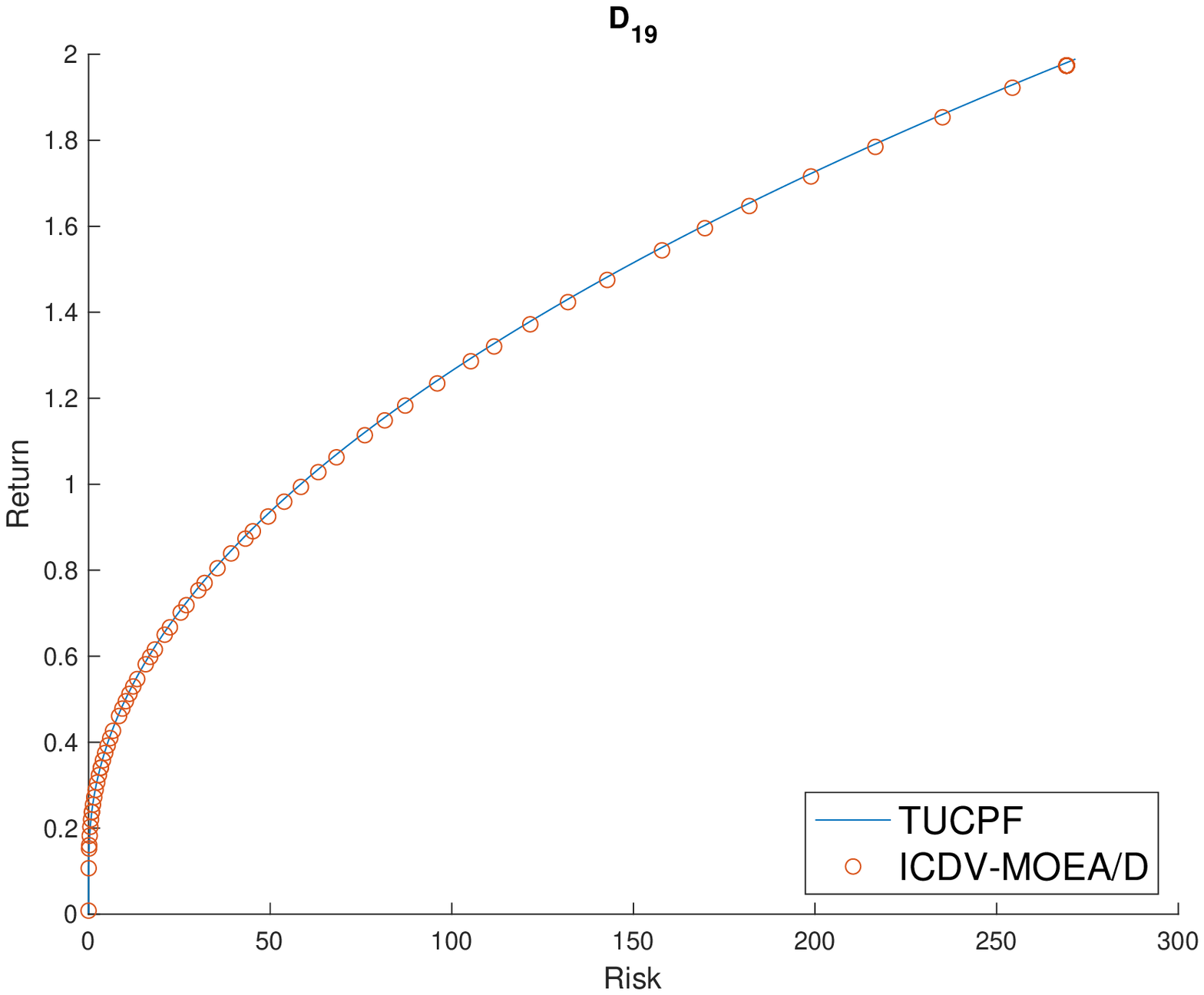}}
\centerline{(e)}
\end{minipage}
\begin{minipage}{0.24\linewidth}
\centerline{\includegraphics[width=1\columnwidth]{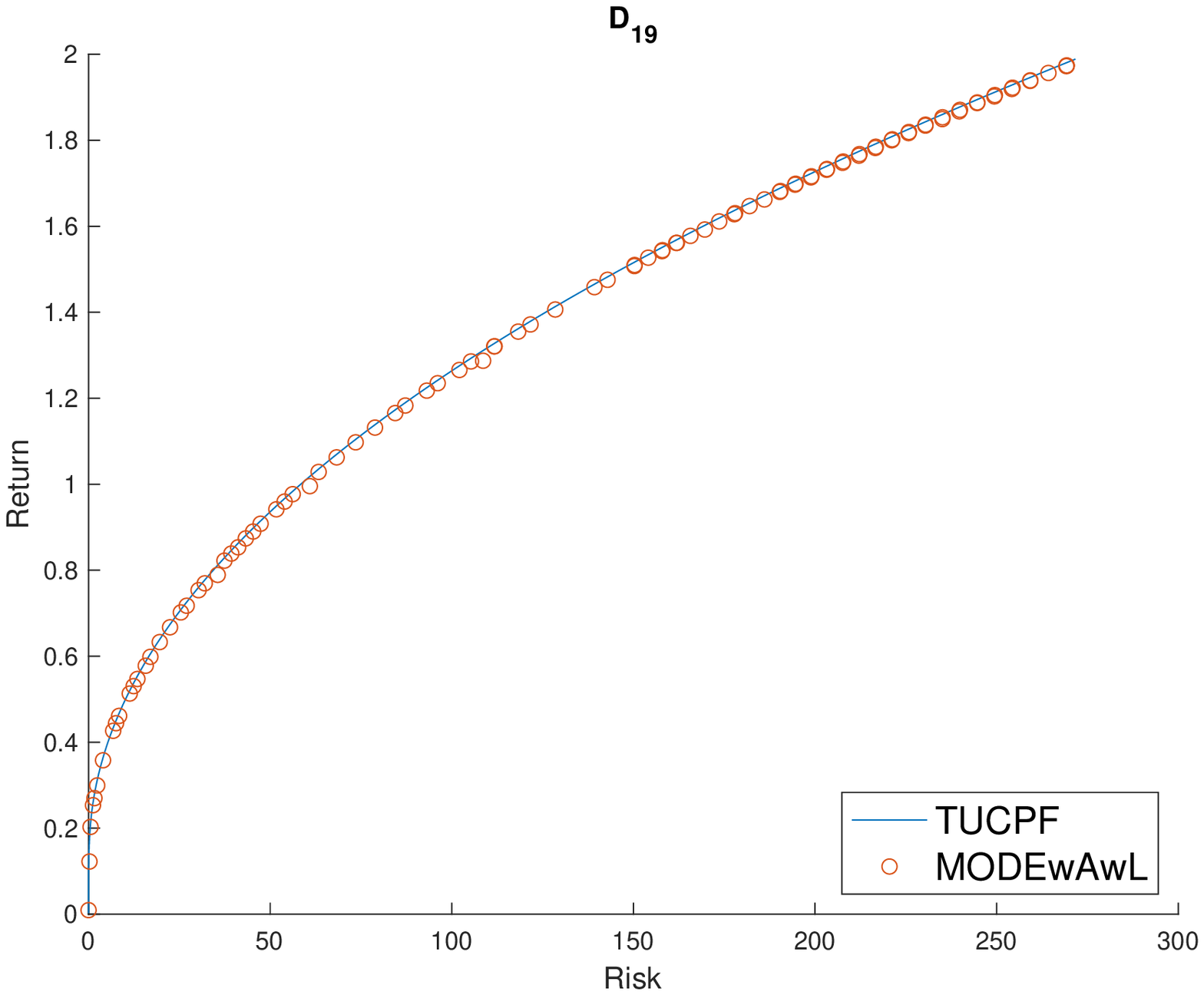}}
\centerline{(f)}
\end{minipage}
\begin{minipage}{0.24\linewidth}
\centerline{\includegraphics[width=1\columnwidth]{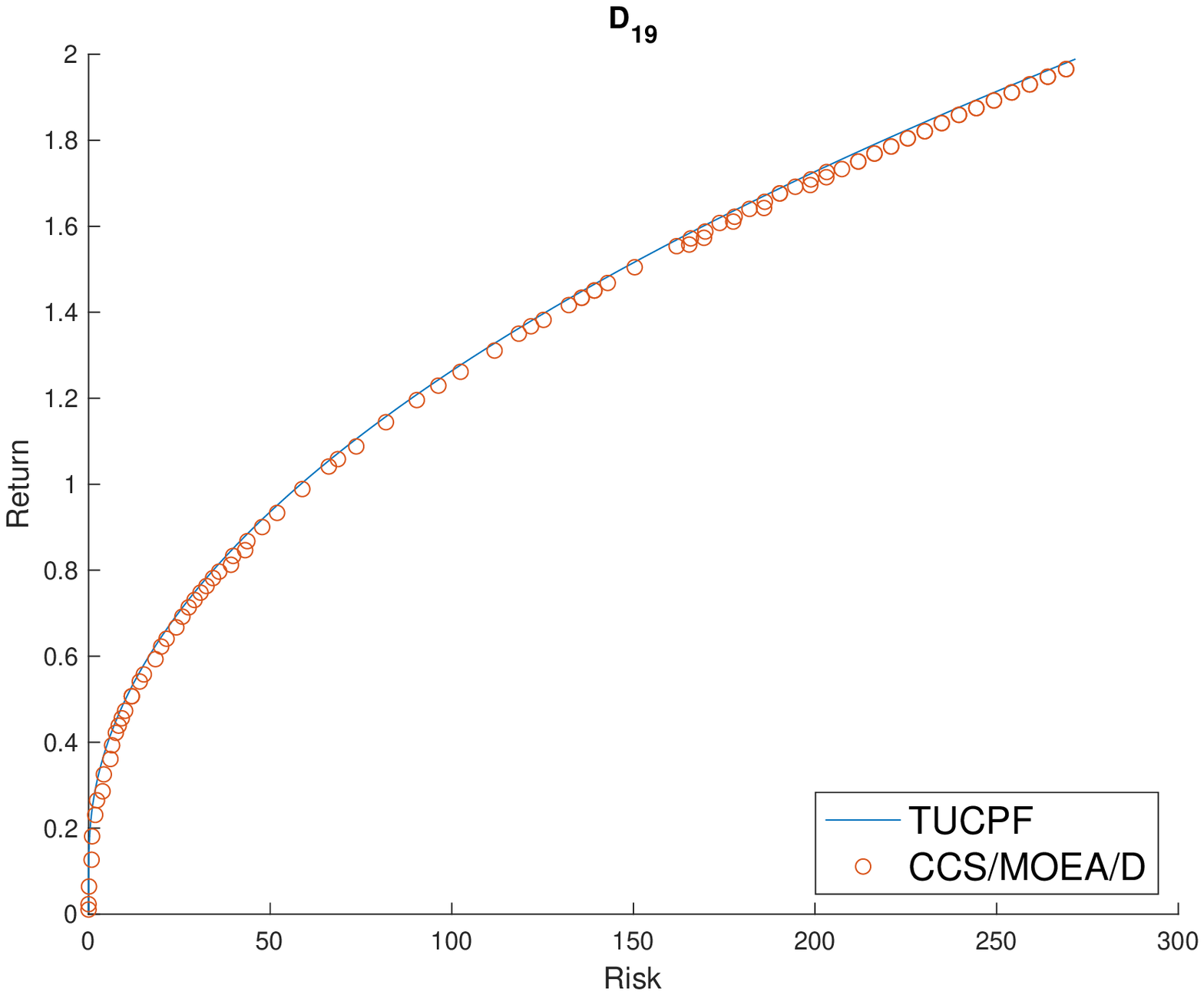}}
\centerline{(g)}
\end{minipage}\\
\begin{minipage}{0.24\linewidth}
\centerline{\includegraphics[width=1\columnwidth]{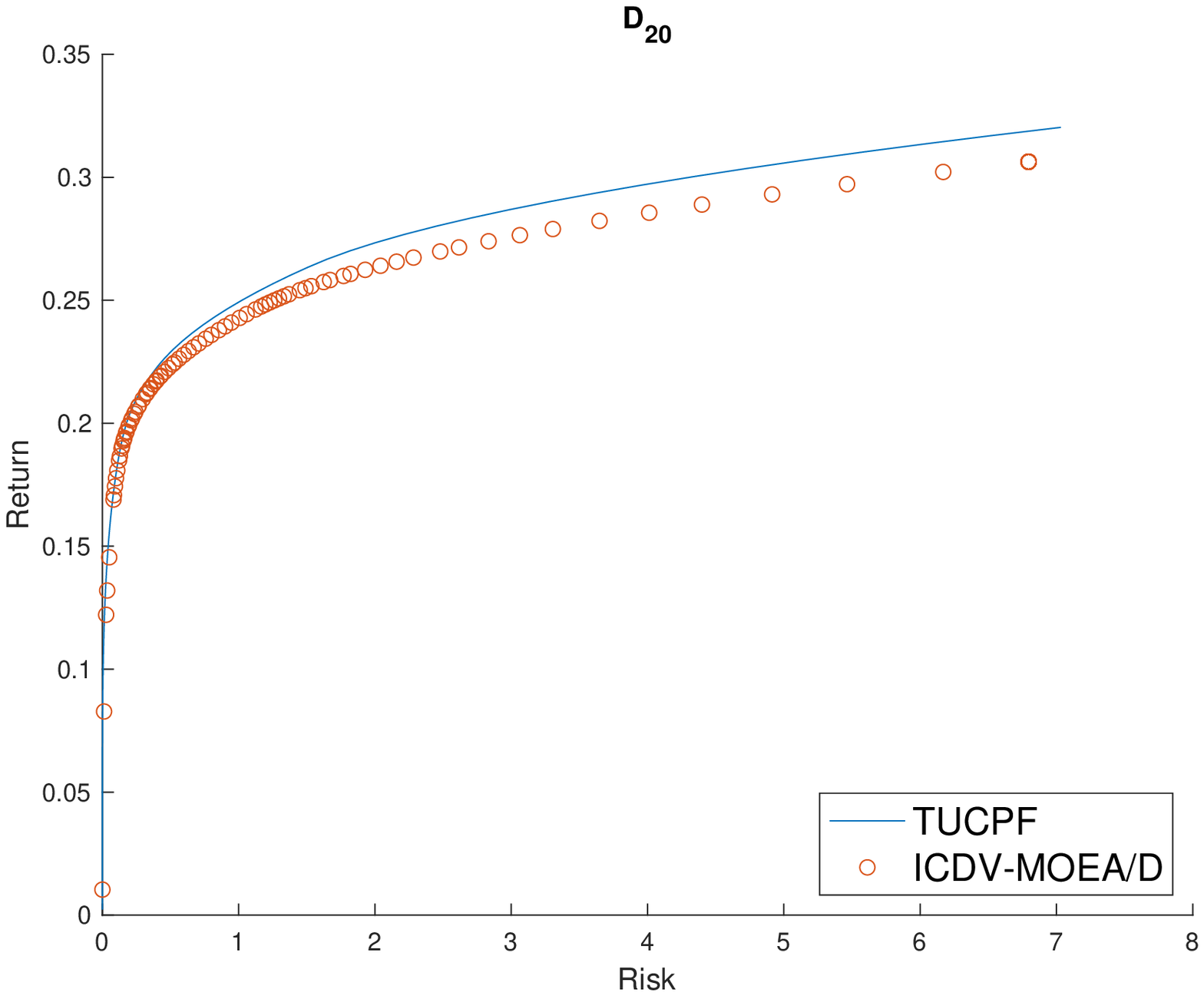}}
\centerline{(h)}
\end{minipage}
\begin{minipage}{0.24\linewidth}
\centerline{\includegraphics[width=1\columnwidth]{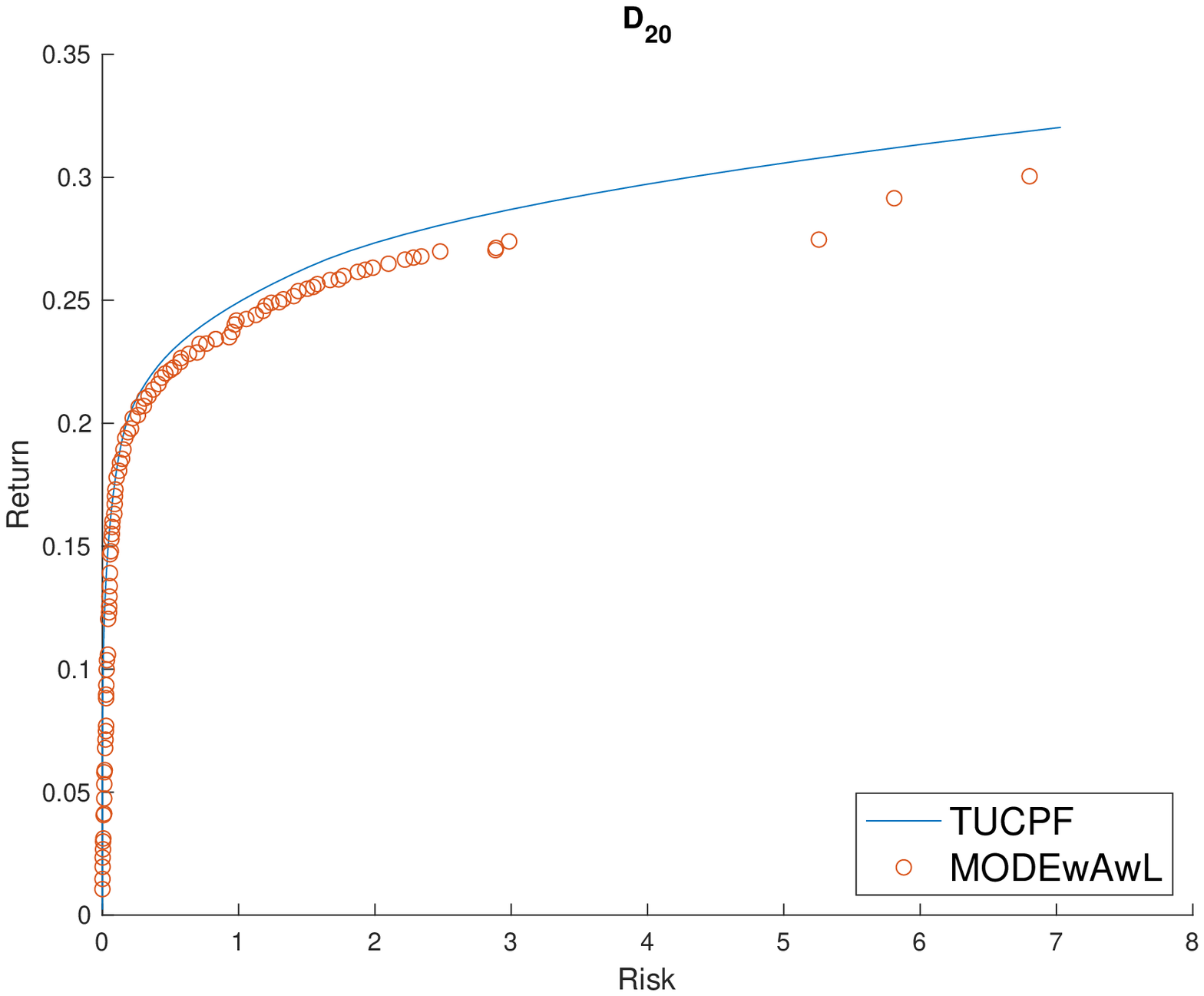}}
\centerline{(i)}
\end{minipage}
\begin{minipage}{0.24\linewidth}
\centerline{\includegraphics[width=1\columnwidth]{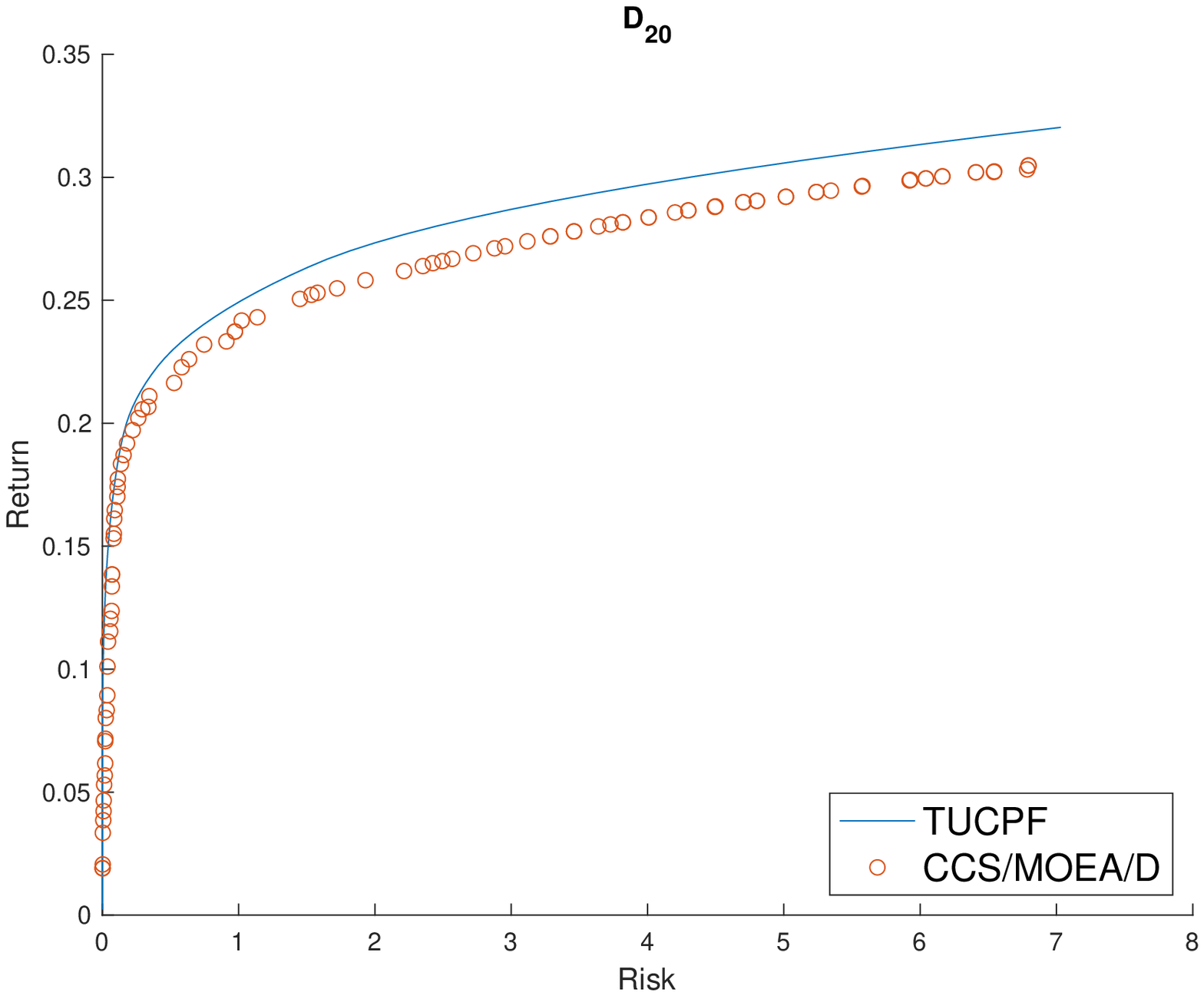}}
\centerline{(j)}
\end{minipage}
\captionsetup{justification=centering}
 \caption{The efficient fronts with the best HVs obtained by D\&O-MOEA/D, MODEwAwL, CCS/MOEA/D and AUGMECON2\&CPLEX with executions for 2000 seconds.} 
 \label{front_each}
\end{figure*}

To be honest, we can not demonstrate that D\&O-MOEA/D is much better than two EAs just based on HVs. Because D\&O-MOEA/D can not always obtain perfectly distributed solutions since it involves conventional optimization methods for MOPs, which pay much less attention to the diversity of solutions than EAs~\cite{miettinen2012nonlinear,mavrotas2009effective}. Furthermore, HV should not be the absolutely correct indicator of portfolio optimization. Because the diversity of solutions on portfolio problems are not always significant. For example, in Figs.~\ref{front_each} (a)-(d), EAs get better diversities when risk is more than 8e-4. Notwithstanding, are those solutions important? They are perhaps not. Because the return increases from 3.5e-3 to 3.6e-3 while the risk doubles, from 8e-4 to 1.6e-3. It is irrational to take such a large risk with a little improvement of return. In our opinion, the employed performance metric should not pay much attention to diversity on portfolio problems since they consist of objectives with preference. GD is hence introduced when HV is not definitely suitable sometimes. 

Concerning GD, in Table~\ref{GD_table}, the results still reveal the obvious advantage of AUGMECON2\&CPLEX on small-scale instances. Although the GDs are supposed to be 0 since all the obtained solutions are optima, some numerical errors occur due to different numbers of grids and the calculation accuracy of employed tools. Further, the performance of D\&O-MOEA/D closely follows AUGMECON2\&CPLEX on small-scale instances and it outperforms two EAs on most problems. Regarding the Wilcoxon rank-sum test, the statistics are `1/18/1' and `3/15/2' for MODEwAwL and CCS-MOEA/D. It indicates the superiority of D\&O-MOEA/D is outstanding. Moreover, Figs.~\ref{HV_convergence} (e)-(h) show that two EAs fall into local optima although they converge early. D\&O-MOEA/D converges better to the contrary since it gets lower GDs. Further, Figs.~\ref{front_each} (e)-(j) illustrate solutions from D\&O-MOEA/D are closer to the PF, although they do not distribute uniformly.

\begin{table}\scriptsize
\caption{Results on $D_1-D_{20}$ concerning GD with executions for 2000 seconds.}
\label{GD_table}
\begin{tabular}{|cccccc|}\hline
\multicolumn{2}{|c}{Algo.}&\multicolumn{1}{c}{A}&\multicolumn{1}{c}{B}&\multicolumn{1}{c}{C}&\multicolumn{1}{c|}{D}\\\hline
\multirow{2}{*}{$D_{1}$}
&Mean	&2.05e-07[2]	&1.85e-06[4]	&8.13e-07[3]	&\cellcolor{gray25}\\
&Std	&5.08e-08	&3.23e-07	&2.65e-07	&\multirow{-2}{*}{\cellcolor{gray25}2.02e-07[1]	}\\\hline
\multirow{2}{*}{$D_{2}$}
&Mean	&2.90e-07[2]	&1.35e-06[3]	&3.22e-06[4]	&\cellcolor{gray25}\\
&Std	&7.04e-08	&1.73e-06	&3.36e-06	&\multirow{-2}{*}{\cellcolor{gray25}2.47e-07[1]	}\\\hline
\multirow{2}{*}{$D_{3}$}
&Mean	&1.76e-06[3]	&1.55e-06[2]	&2.54e-06[4]	&\cellcolor{gray25}\\
&Std	&2.49e-07	&1.92e-07	&9.70e-07	&\multirow{-2}{*}{\cellcolor{gray25}1.24e-06[1]	}\\\hline
\multirow{2}{*}{$D_{4}$}
&Mean	&1.32e-05[2]	&1.95e-05[3]	&3.23e-05[4]	&\cellcolor{gray25}\\
&Std	&9.91e-06	&6.17e-06	&2.27e-05	&\multirow{-2}{*}{\cellcolor{gray25}0.00e+00[1]	}\\\hline
\multirow{2}{*}{$D_{5}$}
&Mean	&5.40e-06[2]	&8.60e-06[4]	&5.46e-06[3]	&\cellcolor{gray25}\\
&Std	&2.18e-06	&2.46e-06	&5.86e-06	&\multirow{-2}{*}{\cellcolor{gray25}1.53e-12[1]	}\\\hline
\multirow{2}{*}{$D_{6}$}
&Mean	&\cellcolor{gray25}1.55e-06[1]	&2.19e-06[3]	&2.61e-06[4]	&	\\
&Std	&4.36e-07	&3.11e-07	&1.21e-06	&\multirow{-2}{*}{1.62e-06[2]	}\\\hline
\multirow{2}{*}{$D_{7}$}
&Mean	&3.27e-05[2]	&8.28e-05[3]	&1.33e-04[4]	&\cellcolor{gray25}\\
&Std	&9.67e-06	&1.53e-05	&4.25e-05	&\multirow{-2}{*}{\cellcolor{gray25}0.00e+00[1]	}\\\hline
\multirow{2}{*}{$D_{8}$}
&Mean	&2.97e-05[3]	&4.00e-05[4]	&1.33e-05[2]	&\cellcolor{gray25}\\
&Std	&1.06e-05	&5.52e-06	&7.91e-06	&\multirow{-2}{*}{\cellcolor{gray25}0.00e+00[1]	}\\\hline
\multirow{2}{*}{$D_{9}$}
&Mean	&1.01e-04[2]	&1.97e-04[4]	&1.70e-04[3]	&\cellcolor{gray25}\\
&Std	&1.98e-05	&2.36e-05	&4.00e-05	&\multirow{-2}{*}{\cellcolor{gray25}0.00e+00[1]	}\\\hline
\multirow{2}{*}{$D_{10}$}
&Mean	&1.05e-05[2]	&1.72e-05[3]	&1.82e-05[4]	&\cellcolor{gray25}\\
&Std	&7.63e-06	&6.18e-06	&1.11e-05	&\multirow{-2}{*}{\cellcolor{gray25}0.00e+00[1]	}\\\hline
\multirow{2}{*}{$D_{11}$}
&Mean	&1.08e-05[3]	&2.24e-05[4]	&9.94e-06[2]	&\cellcolor{gray25}\\
&Std	&3.42e-06	&5.25e-06	&9.40e-06	&\multirow{-2}{*}{\cellcolor{gray25}0.00e+00[1]	}\\\hline
\multirow{2}{*}{$D_{12}$}
&Mean	&1.16e-06[2]	&1.25e-06[3]	&1.51e-06[4]	&\cellcolor{gray25}\\
&Std	&4.98e-07	&4.78e-07	&8.06e-07	&\multirow{-2}{*}{\cellcolor{gray25}4.40e-07[1]	}\\\hline
\multicolumn{2}{|c}{Total$^*$}
&26	&40	&41	&13	\\\hline
\multicolumn{2}{|c}{Final Rank$^*$}
&2	&3	&4	&1	\\\hline
\multicolumn{2}{|c}{$+,-,\thickapprox^*$}
&-	&1/10/1	&2/8/2	&9/0/3	\\\hline
\multirow{2}{*}{$D_{13}$}
&Mean	&\cellcolor{gray25}4.28e-04[1]	&4.79e-04[2]	&5.29e-04[3]	&	\\
&Std&7.47e-07	&6.19e-06	&2.28e-05	&\multirow{-2}{*}{-}	\\\hline
\multirow{2}{*}{$D_{14}$}
&Mean	&\cellcolor{gray25}9.50e-05[1]	&1.11e-04[2]	&1.40e-04[3]	&	\\
&Std&4.70e-06	&4.67e-06	&3.32e-05	&\multirow{-2}{*}{-}	\\\hline
\multirow{2}{*}{$D_{15}$}
&Mean	&2.09e-04[2]	&2.78e-04[3]	&\cellcolor{gray25}1.92e-04[1]	&	\\
&Std&2.35e-05	&2.75e-05	&5.03e-05	&\multirow{-2}{*}{-}	\\\hline
\multirow{2}{*}{$D_{16}$}
&Mean	&\cellcolor{gray25}6.11e-05[1]	&6.77e-05[3]	&6.71e-05[2]	&	\\
&Std&8.76e-06	&5.01e-06	&1.01e-05	&\multirow{-2}{*}{-}	\\\hline
\multirow{2}{*}{$D_{17}$}
&Mean	&\cellcolor{gray25}1.79e-02[1]	&2.48e-02[2]	&2.81e-02[3]	&	\\
&Std&1.22e-05	&1.43e-03	&1.47e-03	&\multirow{-2}{*}{-}	\\\hline
\multirow{2}{*}{$D_{18}$}
&Mean	&\cellcolor{gray25}6.01e-04[1]	&8.64e-04[3]	&8.20e-04[2]	&	\\
&Std&1.49e-05	&4.40e-05	&2.79e-05	&\multirow{-2}{*}{-}	\\\hline
\multirow{2}{*}{$D_{19}$}
&Mean	&\cellcolor{gray25}5.46e-03[1]	&6.80e-03[2]	&7.55e-03[3]	&	\\
&Std&4.96e-04	&3.11e-04	&4.20e-04	&\multirow{-2}{*}{-}	\\\hline
\multirow{2}{*}{$D_{20}$}
&Mean	&\cellcolor{gray25}8.70e-04[1]	&1.20e-03[2]	&1.82e-03[3]	&	\\
&Std&2.09e-05	&9.67e-05	&2.05e-04	&\multirow{-2}{*}{-}	\\\hline
\multicolumn{2}{|c}{Total}
&24	&47	&49	&-	\\\hline
\multicolumn{2}{|c}{Final Rank}
&1	&2	&3	&-	\\\hline
\multicolumn{2}{|c}{$+,-,\thickapprox$}
&-	&1/18/1	&3/15/2	&-	\\\hline
\end{tabular}
\end{table}

\subsubsection{Application for Chinese Stock Markets}
\begin{figure}[htbp]\footnotesize
\graphicspath{{figs/}}
\centering
\centerline{\includegraphics[width=0.65\columnwidth]{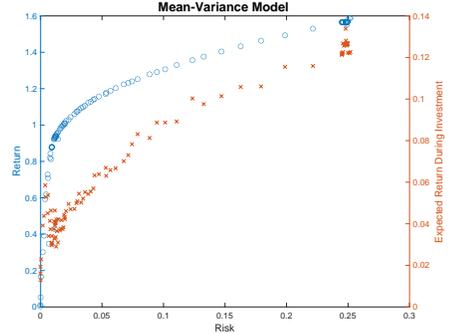}}
\captionsetup{justification=centering}
\caption{Portfolio obtained by D\&O-MOEA/D for Shanghai and Shenzhen A-shares.} 
\label{PSSA}
\end{figure}

\begin{figure}[htbp]\footnotesize
\graphicspath{{figs/}}
\centering
\includegraphics[width=0.65\columnwidth]{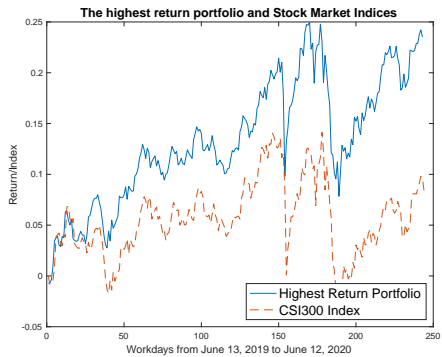}
\captionsetup{justification=centering}
 \caption{Profit and Securities Index} 
 \label{SCI}
\end{figure}

In order to test the proposed method in practice, data from Shanghai and Shenzhen stock markets are collected. The data consists of the close prices of 4510 funds and the trade date ranges from June 13, 2017 to June 12, 2020. Arbitrarily, we determine the data from June 13, 2017 to June 13, 2019 (2 years) as the training data and the data from June 13, 2019 to June 12, 2020 (1 year) as the test data. We normalize the close prices while setting the close prices at June 13, 2017 as the baselines. It is given as follows:
\begin{equation}
\label{normalization_china}
profit_{i,t} = \frac{price_{i,t} - price_{i,0}}{price_{i,0}},
\end{equation}where $price_{i,t}$ is the close price of asset $i$ at time $t$, $profit_{i,t}$ is the normalized return and $t=0$ is the start date. Then the expected return $\mu_i$ and the risk $\sigma_{ij}$ can be obtained from these normalized returns~\cite{kolm201460}. This data is from a financial data platform Tushare\footnote{\textcolor{blue}{hhttps://tushare.pro/register?reg=405855}} and is also uploaded with the source code of D\&O-MOEA/D. D\&O-MOEA/D with $N=100$ and 1000 fitness evaluations is implemented on this data while all the constraints, except the pre-assignment, are involved. In Fig.~\ref{PSSA}, with the Mean-Variance model, points represented with `o' are the performance of every portfolio and the left vertical coordinate is the return. This PF is regular and the return ranges from 0 to 1.6. Meanwhile, the performance of every corresponding portfolio is illustrated as `x' and the right vertical coordinate is the expected return during the investment period, from June 13, 2019 to June 12, 2020. These points show good consistency and
prominent expected return rates of these investments achieve about 13\%. Further, a portfolio with the highest return is picked out. We compare its profit rates with stock market trends, i.e. China Securities Index (CSI300). CSI300 reflects the general picture of the stock price changes in both Shanghai and Shenzhen stock markets. Fig~\ref{SCI} shows the profit rates of this portfolio are higher than the CSI300 most of the time. Moreover, the ups and downs of this portfolio and CSI300 are very consistent. The profit rate of this portfolio has reached about 23\% at the end. Whatever, this is just a very simple application and it is just utilized to confirm the feasibility and the values in application of our method. The real financial market should be much more complicated.
\subsection{Without A Perfect Optimizer}
The existence of a perfect optimizer, which can settle the subtask quickly and accurately, is a strong assumption indeed. Hence, some simulation experiments (black problems in~\cite{li2009multiobjective}) are conducted and a simple yet effective EA, JADE~\cite{zhang2009jade}, is an alternative to the perfect $\textbf{Opt}$ for D\&O-MOEA/D. Since MOEA/D-DE is simultaneously proposed with these problems and it performs well, we make a comparison between D\&O-MOEA/D and MOEA/D-DE on these simple experiments. The parameters are the same as in~\cite{li2009multiobjective} for D\&O-MOEA/D and MOEA/D-DE while the maximum fitness evaluation is 1000. Further, the parameters of JADE follow~\cite{zhang2009jade} and the population size is 20 while the maximum fitness evaluation is 1000. Table~\ref{TEC_HV_GD} shows the summary statistics. The symbol ``$+,-,\thickapprox$'' denotes D\&O-MOEA/D is significantly better than, worse than and similar to MOEA/D-DE. More details are presented in the supplement. The results of MOEA/D-DE are worse than D\&O-MOEA/D, however, the running times raise to about an hour from several seconds. Apparently, exchanging massive computational resources with limited performance improvement is not rational. Therefore, a perfect optimizer $\textbf{Opt}$ should be found first when implementing D\&O-MOEA/D.
\begin{table}\scriptsize
\caption{Summary statistics on $F_1-F_9$.}
\label{TEC_HV_GD}
\begin{tabular}{|ccc|}\hline
\multicolumn{1}{|c}{Metrics}&\multicolumn{1}{c}{HV}&\multicolumn{1}{c|}{GD}\\\hline
\multicolumn{1}{|c}{$+,-,\thickapprox$}
&8/1/0	&6/2/1	\\\hline
\end{tabular}
\end{table}

\section{Conclusion and Future Work}
This work is the first one that discusses a concept of variable division and optimization (D\&O), to our best knowledge. It should be one potential research realm since this technique is frequently used in the practices of EAs. More detailed and in-depth discussions will not only help us clearly understand the nature of this technique but also assist us to develop new theories and methods for EAs. According to the discussions, D\&O-MOEA/D, in which the variable of the original problem is divided and the partial variables are optimized respectively, is proposed. Then, it is achieved via integrating with a quadratic optimizer of CPLEX for constrained portfolio problems, mixed-integer problems. D\&O-MOEA/D, representing the hybrid method, is compared to one exact method and two EAs. The empirical studies show that different kinds of methods have their own advantages. The exact method shows its superiority on small-scale instances. On the other hand, the hybrid one performs well on small-scale problems while it shows its superiority and versatility for large-scale instances. Further, we apply the proposed method to the real stock markets, the results show the feasibility and potential of D\&O-MOEA/D although this application is simple. Since the stock markets are dynamic, we will extend the adopted single period MV model to multi-period in the future and thereby the requirement of new methods will emerge. Further, while few works discuss D\&O, a huge gap is waiting to be filled. We will not stop exploring this realm.

\footnotesize
\bibliographystyle{ieeetr}
\bibliography{Overall_Reference}

\begin{thebibliography}{10}

\bibitem{boyd2004convex}
S.~Boyd and L.~Vandenberghe, {\em Convex optimization}.
\newblock Cambridge university press, 2004.

\bibitem{cormen2009introduction}
T.~H. Cormen, C.~E. Leiserson, R.~L. Rivest, and C.~Stein, {\em Introduction to
  algorithms}.
\newblock MIT press, 2009.

\bibitem{antonio2017coevolutionary}
L.~M. Antonio and C.~A.~C. Coello, ``Coevolutionary multi-objective
  evolutionary algorithms: A survey of the state-of-the-art,'' {\em IEEE
  Transactions on Evolutionary Computation}, 2017.

\bibitem{ma2018survey}
X.~Ma, X.~Li, Q.~Zhang, K.~Tang, Z.~Liang, W.~Xie, and Z.~Zhu, ``A survey on
  cooperative co-evolutionary algorithms,'' {\em IEEE Transactions on
  Evolutionary Computation}, 2018.

\bibitem{yao1999evolving}
X.~Yao, ``Evolving artificial neural networks,'' {\em Proceedings of the IEEE},
  vol.~87, no.~9, pp.~1423--1447, 1999.

\bibitem{moral2006selection}
R.~Moral-Escudero, R.~Ruiz-Torrubiano, and A.~Su{\'a}rez, ``Selection of
  optimal investment portfolios with cardinality constraints,'' in {\em 2006
  IEEE International Conference on Evolutionary Computation}, pp.~2382--2388,
  IEEE, 2006.

\bibitem{yang2017turning}
P.~Yang, K.~Tang, and X.~Yao, ``Turning high-dimensional optimization into
  computationally expensive optimization,'' {\em IEEE Transactions on
  Evolutionary Computation}, vol.~22, no.~1, pp.~143--156, 2017.

\bibitem{jin2005comprehensive}
Y.~Jin, ``A comprehensive survey of fitness approximation in evolutionary
  computation,'' {\em Soft computing}, vol.~9, no.~1, pp.~3--12, 2005.

\bibitem{jin2011surrogate}
Y.~Jin, ``Surrogate-assisted evolutionary computation: Recent advances and
  future challenges,'' {\em Swarm and Evolutionary Computation}, vol.~1, no.~2,
  pp.~61--70, 2011.

\bibitem{sinha2017review}
A.~Sinha, P.~Malo, and K.~Deb, ``A review on bilevel optimization: From
  classical to evolutionary approaches and applications,'' {\em IEEE
  Transactions on Evolutionary Computation}, vol.~22, no.~2, pp.~276--295,
  2017.

\bibitem{omidvar2013cooperative}
M.~N. Omidvar, X.~Li, Y.~Mei, and X.~Yao, ``Cooperative co-evolution with
  differential grouping for large scale optimization,'' {\em IEEE Transactions
  on evolutionary computation}, vol.~18, no.~3, pp.~378--393, 2013.

\bibitem{yang2008large}
Z.~Yang, K.~Tang, and X.~Yao, ``Large scale evolutionary optimization using
  cooperative coevolution,'' {\em Information Sciences}, vol.~178, no.~15,
  pp.~2985--2999, 2008.

\bibitem{sun2018evolving}
Y.~Sun, G.~G. Yen, and Z.~Yi, ``Evolving unsupervised deep neural networks for
  learning meaningful representations,'' {\em IEEE Transactions on Evolutionary
  Computation}, 2018.

\bibitem{lu2019nsga}
Z.~Lu, I.~Whalen, V.~Boddeti, Y.~Dhebar, K.~Deb, E.~Goodman, and W.~Banzhaf,
  ``Nsga-net: neural architecture search using multi-objective genetic
  algorithm,'' in {\em Proceedings of the Genetic and Evolutionary Computation
  Conference}, pp.~419--427, 2019.

\bibitem{yao1997new}
X.~Yao and Y.~Liu, ``A new evolutionary system for evolving artificial neural
  networks,'' {\em IEEE transactions on neural networks}, vol.~8, no.~3,
  pp.~694--713, 1997.

\bibitem{liu1996population}
Y.~Liu and X.~Yao, ``A population-based learning algorithm which learns both
  architectures and weights of neural networks,'' {\em Chinese Journal of
  Advanced Software Research}, vol.~3, pp.~54--65, 1996.

\bibitem{lu2020nsganetv2}
Z.~Lu, K.~Deb, E.~Goodman, W.~Banzhaf, and V.~N. Boddeti, ``Nsganetv2:
  Evolutionary multi-objective surrogate-assisted neural architecture search,''
  in {\em European Conference on Computer Vision}, pp.~35--51, Springer, 2020.

\bibitem{branke2009portfolio}
J.~Branke, B.~Scheckenbach, M.~Stein, K.~Deb, and H.~Schmeck, ``Portfolio
  optimization with an envelope-based multi-objective evolutionary algorithm,''
  {\em European Journal of Operational Research}, vol.~199, no.~3,
  pp.~684--693, 2009.

\bibitem{kolm201460}
P.~N. Kolm, R.~T{\"u}t{\"u}nc{\"u}, and F.~J. Fabozzi, ``60 years of portfolio
  optimization: Practical challenges and current trends,'' {\em European
  Journal of Operational Research}, vol.~234, no.~2, pp.~356--371, 2014.

\bibitem{ponsich2013survey}
A.~Ponsich, A.~L. Jaimes, and C.~A.~C. Coello, ``A survey on multiobjective
  evolutionary algorithms for the solution of the portfolio optimization
  problem and other finance and economics applications,'' {\em IEEE
  Transactions on Evolutionary Computation}, vol.~17, no.~3, pp.~321--344,
  2013.

\bibitem{ertenlice2018survey}
O.~Ertenlice and C.~B. Kalayci, ``A survey of swarm intelligence for portfolio
  optimization: Algorithms and applications,'' {\em Swarm and evolutionary
  computation}, vol.~39, pp.~36--52, 2018.

\bibitem{1952MarkowitzPS}
H.~Markowitz, ``Portfolio selection,'' {\em The Journal of Finance}, vol.~7,
  pp.~77--91, Mar 1952.

\bibitem{DBLP:journals/corr/abs-1909-08748}
Y.~Chen, A.~Zhou, and S.~Das, ``Utilizing dependence among variables in
  evolutionary algorithms for mixed-integer programming: A case study on
  multi-objective constrained portfolio optimization,'' {\em CoRR},
  vol.~abs/1909.08748, 2019.

\bibitem{steinmean}
D.-W.-I.~M. Stein, ``Mean-variance portfolio selection with complex
  constraints,''

\bibitem{2007QingfuMOEAD}
Q.~Zhang and H.~Li, ``{MOEA/D: A multiobjective evolutionary algorithm based on
  decomposition},'' {\em IEEE Transactions on Evolutionary Computation},
  vol.~11, no.~6, pp.~712--731, 2007.

\bibitem{zhou2011multiobjective}
A.~Zhou, B.-Y. Qu, H.~Li, S.-Z. Zhao, P.~N. Suganthan, and Q.~Zhang,
  ``Multiobjective evolutionary algorithms: A survey of the state of the art,''
  {\em Swarm and Evolutionary Computation}, vol.~1, no.~1, pp.~32--49, 2011.

\bibitem{li2009multiobjective}
H.~Li and Q.~Zhang, ``{Multiobjective optimization problems with complicated
  Pareto sets, MOEA/D and NSGA-II},'' {\em IEEE Transactions on Evolutionary
  Computation}, vol.~13, no.~2, pp.~284--302, 2009.

\bibitem{wang2015adaptive}
Z.~Wang, Q.~Zhang, A.~Zhou, M.~Gong, and L.~Jiao, ``Adaptive replacement
  strategies for moea/d,'' {\em IEEE transactions on cybernetics}, vol.~46,
  no.~2, pp.~474--486, 2015.

\bibitem{qi2014moea}
Y.~Qi, X.~Ma, F.~Liu, L.~Jiao, J.~Sun, and J.~Wu, ``{MOEA/D} with adaptive
  weight adjustment,'' {\em Evolutionary computation}, vol.~22, no.~2,
  pp.~231--264, 2014.

\bibitem{ma2017tchebycheff}
X.~Ma, Q.~Zhang, G.~Tian, J.~Yang, and Z.~Zhu, ``On tchebycheff decomposition
  approaches for multiobjective evolutionary optimization,'' {\em IEEE
  Transactions on Evolutionary Computation}, vol.~22, no.~2, pp.~226--244,
  2017.

\bibitem{miettinen2012nonlinear}
K.~Miettinen, {\em Nonlinear multiobjective optimization}, vol.~12.
\newblock Springer Science \& Business Media, 2012.

\bibitem{1994BeanGenetic}
J.~C. Bean, ``Genetic algorithms and random keys for sequencing and
  optimization,'' {\em ORSA Journal on Computing}, vol.~6, no.~2, pp.~154--160,
  1994.

\bibitem{1996RainerDE}
R.~Storn and K.~Price, ``Differential evolution--a simple and efficient
  heuristic for global optimization over continuous spaces,'' {\em Journal of
  Global Optimization}, vol.~11, no.~4, pp.~341--359, 1997.

\bibitem{das2010differential}
S.~Das and P.~N. Suganthan, ``Differential evolution: A survey of the
  state-of-the-art,'' {\em IEEE transactions on evolutionary computation},
  vol.~15, no.~1, pp.~4--31, 2010.

\bibitem{2002DebNSGA2}
K.~Deb, A.~Pratap, S.~Agarwal, and T.~Meyarivan, ``{A fast and elitist
  multiobjective genetic algorithm: NSGA-II},'' {\em IEEE Transactions on
  Evolutionary Computation}, vol.~6, no.~2, pp.~182--197, 2002.

\bibitem{2014KhinMODEwawl}
K.~Lwin, R.~Qu, and G.~Kendall, ``A learning-guided multi-objective
  evolutionary algorithm for constrained portfolio optimization,'' {\em Applied
  Soft Computing}, vol.~24, pp.~757--772, 2014.

\bibitem{chen2018evolutionary}
Y.~Chen, A.~Zhou, and L.~Dou, ``An evolutionary algorithm with a new operator
  and an adaptive strategy for large-scale portfolio problems,'' in {\em
  Proceedings of the Genetic and Evolutionary Computation Conference
  Companion}, pp.~247--248, ACM, 2018.

\bibitem{mavrotas2009effective}
G.~Mavrotas, ``Effective implementation of the $\varepsilon$-constraint method
  in multi-objective mathematical programming problems,'' {\em Applied
  mathematics and computation}, vol.~213, no.~2, pp.~455--465, 2009.

\bibitem{mavrotas2013improved}
G.~Mavrotas and K.~Florios, ``An improved version of the augmented
  $\varepsilon$-constraint method (augmecon2) for finding the exact pareto set
  in multi-objective integer programming problems,'' {\em Applied Mathematics
  and Computation}, vol.~219, no.~18, pp.~9652--9669, 2013.

\bibitem{liu2018termination}
Y.~Liu, A.~Zhou, and H.~Zhang, ``Termination detection strategies in
  evolutionary algorithms: a survey,'' in {\em Proceedings of the Genetic and
  Evolutionary Computation Conference}, pp.~1063--1070, ACM, 2018.

\bibitem{zitzler1999multiobjective}
E.~Zitzler and L.~Thiele, ``Multiobjective evolutionary algorithms: a
  comparative case study and the strength pareto approach,'' {\em IEEE
  Transactions on Evolutionary Computation}, vol.~3, no.~4, pp.~257--271, 1999.

\bibitem{van1999multiobjective}
D.~A. Van~Veldhuizen, ``Multiobjective evolutionary algorithms:
  classifications, analyses, and new innovations,'' tech. rep., AIR FORCE INST
  OF TECH WRIGHT-PATTERSONAFB OH SCHOOL OF ENGINEERING, 1999.

\bibitem{tian2017platemo}
Y.~Tian, R.~Cheng, X.~Zhang, and Y.~Jin, ``Platemo: A matlab platform for
  evolutionary multi-objective optimization [educational forum],'' {\em IEEE
  Computational Intelligence Magazine}, vol.~12, no.~4, pp.~73--87, 2017.

\bibitem{ishibuchi2018specify}
H.~Ishibuchi, R.~Imada, Y.~Setoguchi, and Y.~Nojima, ``How to specify a
  reference point in hypervolume calculation for fair performance comparison,''
  {\em Evolutionary computation}, vol.~26, no.~3, pp.~411--440, 2018.

\bibitem{elsken2018neural}
T.~Elsken, J.~H. Metzen, and F.~Hutter, ``Neural architecture search: A
  survey,'' {\em arXiv preprint arXiv:1808.05377}, 2018.

\bibitem{zhang2009jade}
J.~Zhang and A.~C. Sanderson, ``Jade: adaptive differential evolution with
  optional external archive,'' {\em IEEE Transactions on evolutionary
  computation}, vol.~13, no.~5, pp.~945--958, 2009.

\bibitem{deb2005scalable}
K.~Deb, L.~Thiele, M.~Laumanns, and E.~Zitzler, ``Scalable test problems for
  evolutionary multiobjective optimization,'' in {\em Evolutionary
  multiobjective optimization}, pp.~105--145, Springer, 2005.

\end{thebibliography}
\clearpage

This supplement discusses various local Pareto fronts (PFs) and presents the results of the supplementary experiment. There should be many different local PFs, however, only $two$ cases of them can utilize the existing method to determine which local PF is better. First, when the local PF of $x_{\uppercase\expandafter{\romannumeral1}}$ is just one point. Second, there is always at least one point in a local PF that dominates any point from the other one. A D\&O based multiobjective minimization problem is stated as follows
\begin{equation*}
\begin{gathered}
\min \ F(x)=(f_1(x_{\uppercase\expandafter{\romannumeral1}},x_{\uppercase\expandafter{\romannumeral2}}),\cdots,f_m(x_{\uppercase\expandafter{\romannumeral1}},x_{\uppercase\expandafter{\romannumeral2}})) \\
s.t \quad x \in \Omega, \  x_{\uppercase\expandafter{\romannumeral1}} \in \Omega_{\uppercase\expandafter{\romannumeral1}} , \  x_{\uppercase\expandafter{\romannumeral2}} \in \Omega_{\uppercase\expandafter{\romannumeral2}}  \ ,
\end{gathered}
\end{equation*}where $F$ is a multiobjective function consists of $m$ real-valued objective function, $x$ is the decision variable, $x_{\uppercase\expandafter{\romannumeral1}}$ and $x_{\uppercase\expandafter{\romannumeral2}}$ are two partial variables, and $x=(x_{\uppercase\expandafter{\romannumeral1}},x_{\uppercase\expandafter{\romannumeral2}})$. $\Omega_{(\cdot)}$ is the search space of each variable, and $\Omega_{\uppercase\expandafter{\romannumeral1}}\times \Omega_{\uppercase\expandafter{\romannumeral2}} = \Omega$. Formal discussions are presented below.
\subsubsection{Case One}
When the local PF of $x_{\uppercase\expandafter{\romannumeral1}}$ is just one point if the following proposition holds.\\
\textit{Proposition 1.}  $\forall x_{\uppercase\expandafter{\romannumeral1}} \in \Omega_{\uppercase\expandafter{\romannumeral1}}, \forall x_{\uppercase\expandafter{\romannumeral2}} \in \Omega_{\uppercase\expandafter{\romannumeral2}}\ there\ is\ a\ (or\ some) \ x_{\uppercase\expandafter{\romannumeral2}}^*  \in \Omega_{\uppercase\expandafter{\romannumeral2}}\ that\ F(x_{\uppercase\expandafter{\romannumeral1}},x_{\uppercase\expandafter{\romannumeral2}}^*)\preceq F(x_{\uppercase\expandafter{\romannumeral1}},x_{\uppercase\expandafter{\romannumeral2}})$.

\begin{figure}[htbp]\footnotesize
\graphicspath{{figs/}}
\centerline{\includegraphics[width=0.5\columnwidth]{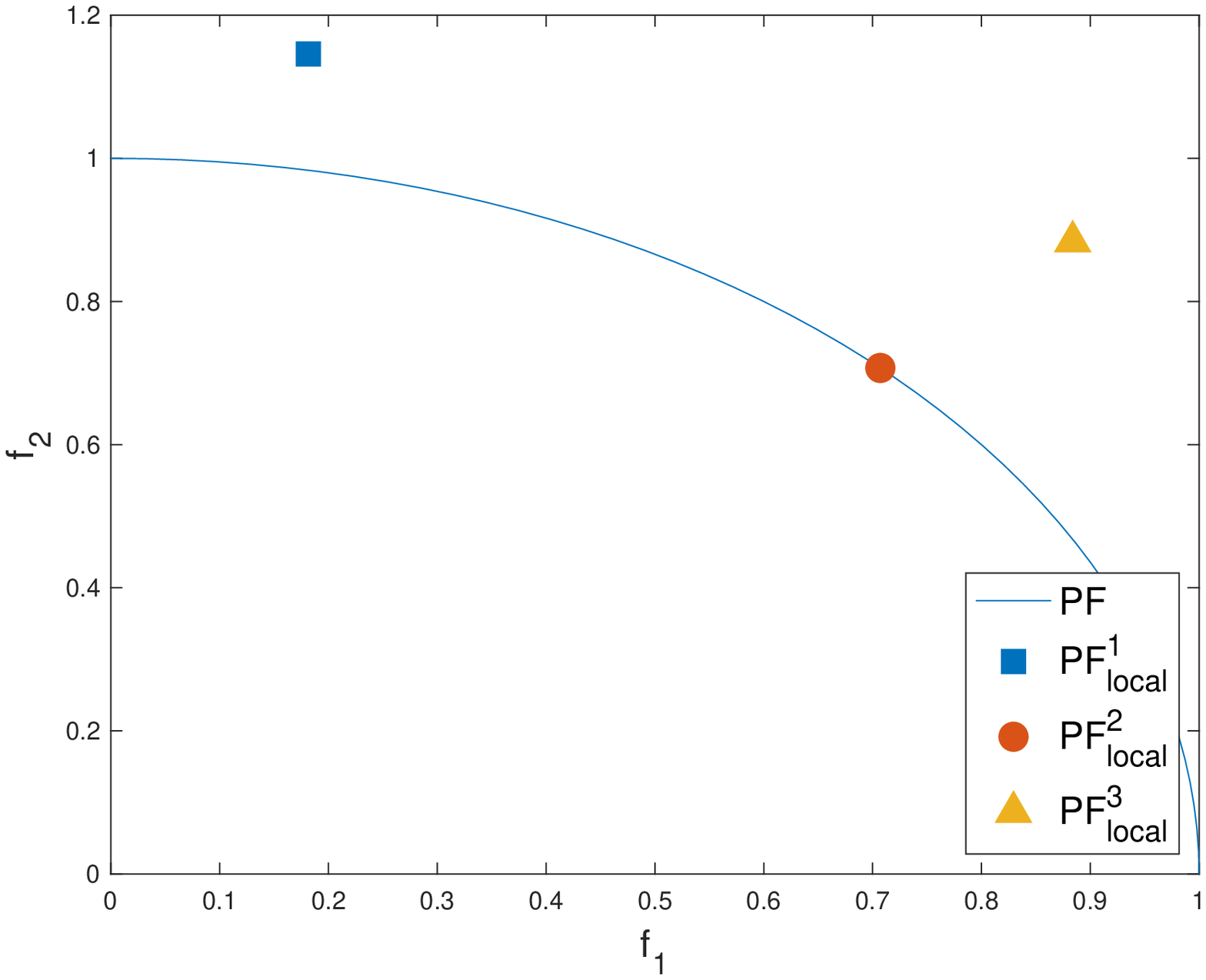}}
 \caption{Simplified local optimal solution search for a multiobjective problem.} 
 \label{LPF1}
\end{figure}

For example, an alteration of DTLZ2~\cite{deb2005scalable} is given as
\begin{equation}
\label{aDTLZ2}
\begin{gathered}
min\ f_1(x) = (1+\sum_{i=2,3}(x_i-0.5)^2)cos(\frac{\pi}{2}x_1)\\
min\ f_2(x) = (1+\sum_{i=2,3}(x_i-0.5)^2)sin(\frac{\pi}{2}x_1)\\
s.t \quad x \in [0,1]^3,
\end{gathered}
\end{equation}where the Pareto optimal solutions are $x=(\cdot,0.5,0.5)$.
Let $x_{\uppercase\expandafter{\romannumeral1}} = (x_1,x_2)$ and $x_{\uppercase\expandafter{\romannumeral2}}=x_3$. The unique Pareto optimal solution of $x_{\uppercase\expandafter{\romannumeral1}} = (x_1,x_2)$ is always $x=(x_1,x_2,0.5)$, i.e. $x_3=0.5$. It is shown in Fig.~\ref{LPF1}. Existing point-based methods can rank these local PFs since they are points.

\subsubsection{Case Two}
There is always at least one point in a local PF, $PF_{local}^1$, that dominates any point from the other one, $PF_{local}^2$, if the following proposition holds.\\
\textit{Proposition 2.}  $\forall F(x_{\uppercase\expandafter{\romannumeral1}}^2,x_{\uppercase\expandafter{\romannumeral2}}^2) \in PF_{local}^2, \exists F(x_{\uppercase\expandafter{\romannumeral1}}^1,x_{\uppercase\expandafter{\romannumeral2}}^1) \in PF_{local}^1\ that\ F(x_{\uppercase\expandafter{\romannumeral1}}^1,x_{\uppercase\expandafter{\romannumeral2}}^1)\preceq F(x_{\uppercase\expandafter{\romannumeral2}}^1,x_{\uppercase\expandafter{\romannumeral2}}^2)$.

\begin{figure}[htbp]\footnotesize
\graphicspath{{figs/}}
\centerline{\includegraphics[width=0.5\columnwidth]{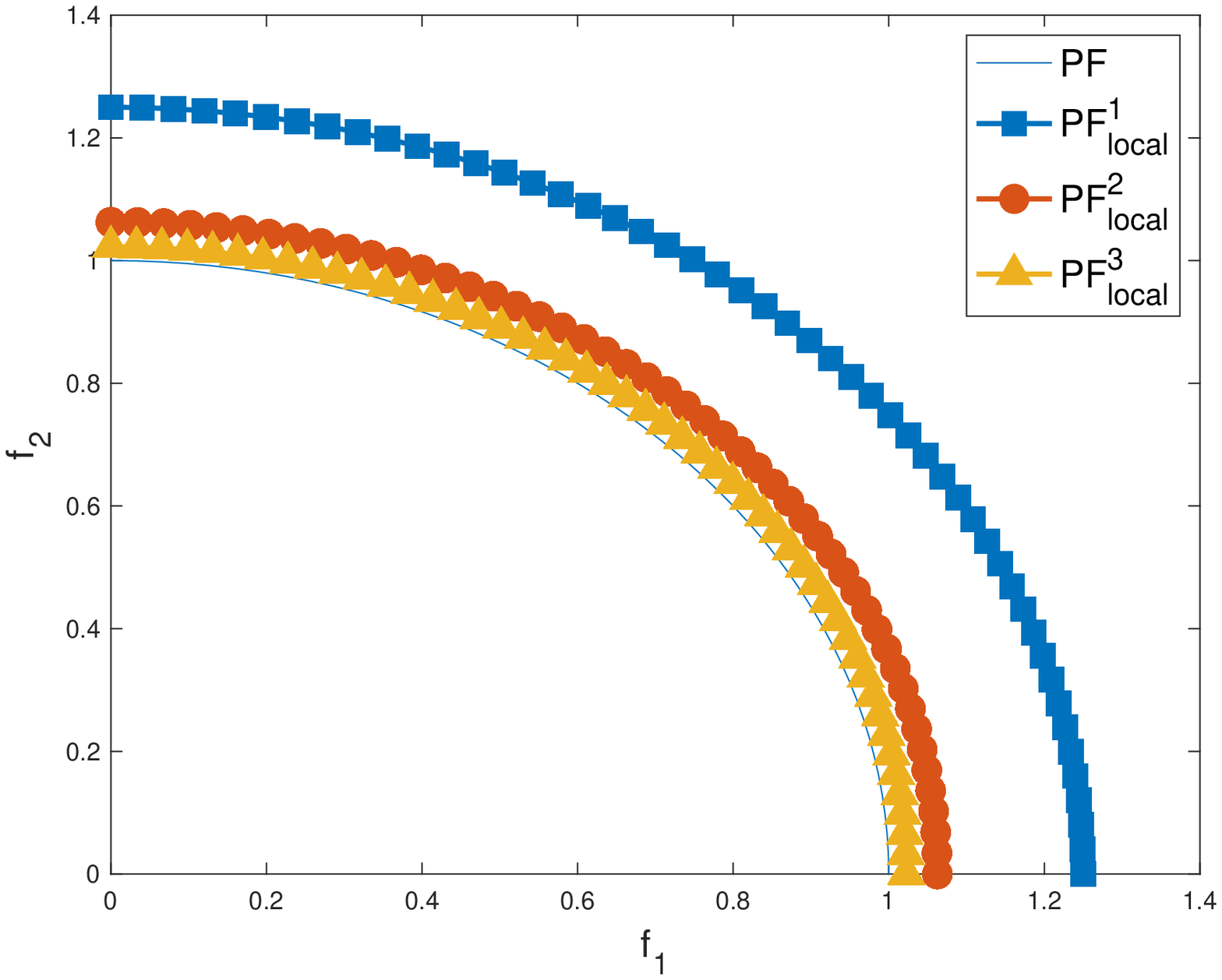}}
 \caption{Simplified local optimal solution search for a multiobjective problem.} 
 \label{LPF2}
\end{figure}

The problem above is taken as example again. Let $x_{\uppercase\expandafter{\romannumeral1}} = x_2$ and $x_{\uppercase\expandafter{\romannumeral2}}=(x_1,x_3)$. Various local PFs of $x_{\uppercase\expandafter{\romannumeral1}}$ will show a front dominance as the \textit{Proposition 2}. Fig.~\ref{LPF2} illustrates different local PFs that can be easily ranked.

Except for these two cases, the determination about which local PF is better can not be performed based on existing methods.

The following tables and figures show the result of the simulation experiment on the test problems from MOEA/D-DE.


\begin{table}\scriptsize
\caption{Results on $F_1-F_9$ concerning HV.}
\begin{tabular}{|cccc|}\hline
\multicolumn{2}{|c}{Algo.}&\multicolumn{1}{c}{MOEA/D-DE}&\multicolumn{1}{c|}{D\&O-MOEA/D}\\\hline
\multirow{2}{*}{$F_{1}$}
&Mean	&5.52e-01[2]	&\cellcolor{gray25}6.12e-01[1]	\\
&Std	&2.19e-02	&1.42e-02	\\\hline
\multirow{2}{*}{$F_{2}$}
&Mean	&4.96e-02[2]	&\cellcolor{gray25}3.26e-01[1]	\\
&Std	&3.55e-02	&2.02e-02	\\\hline
\multirow{2}{*}{$F_{3}$}
&Mean	&4.89e-01[2]	&\cellcolor{gray25}5.32e-01[1]	\\
&Std	&1.66e-02	&1.50e-02	\\\hline
\multirow{2}{*}{$F_{4}$}
&Mean	&4.90e-01[2]	&\cellcolor{gray25}5.35e-01[1]	\\
&Std	&1.53e-02	&1.64e-02	\\\hline
\multirow{2}{*}{$F_{5}$}
&Mean	&5.18e-01[2]	&\cellcolor{gray25}5.59e-01[1]	\\
&Std	&1.06e-02	&9.65e-03	\\\hline
\multirow{2}{*}{$F_{6}$}
&Mean	&\cellcolor{gray25}1.89e-01[1]	&3.30e-02[2]	\\
&Std	&3.87e-02	&1.02e-02	\\\hline
\multirow{2}{*}{$F_{7}$}
&Mean	&3.95e-02[2]	&\cellcolor{gray25}1.37e-01[1]	\\
&Std	&4.11e-02	&4.30e-02	\\\hline
\multirow{2}{*}{$F_{8}$}
&Mean	&2.91e-02[2]	&\cellcolor{gray25}1.22e-01[1]	\\
&Std	&4.13e-02	&2.73e-02	\\\hline
\multirow{2}{*}{$F_{9}$}
&Mean	&1.82e-03[2]	&\cellcolor{gray25}4.75e-02[1]	\\
&Std	&4.28e-03	&1.59e-02	\\\hline
\multicolumn{2}{|c}{Total$$}
&17	&10	\\\hline
\multicolumn{2}{|c}{Final Rank$$}
&2	&1	\\\hline
\multicolumn{2}{|c}{$+,-,\thickapprox$}
&-	&8/1/0	\\\hline
\end{tabular}
\end{table}
\begin{table}\scriptsize
\caption{Results on $F_1-F_9$ concerning GD.}
\begin{tabular}{|cccc|}\hline
\multicolumn{2}{|c}{Algo.}&\multicolumn{1}{c}{MOEA/D-DE}&\multicolumn{1}{c|}{D\&O-MOEA/D}\\\hline
\multirow{2}{*}{$F_{1}$}
&Mean	&8.89e-03[2]	&\cellcolor{gray25}4.35e-03[1]	\\
&Std	&1.30e-03	&1.26e-03	\\\hline
\multirow{2}{*}{$F_{2}$}
&Mean	&5.84e-02[2]	&\cellcolor{gray25}4.21e-02[1]	\\
&Std	&5.66e-03	&5.95e-03	\\\hline
\multirow{2}{*}{$F_{3}$}
&Mean	&\cellcolor{gray25}1.63e-02[1]	&2.49e-02[2]	\\
&Std	&4.32e-03	&6.41e-03	\\\hline
\multirow{2}{*}{$F_{4}$}
&Mean	&\cellcolor{gray25}1.74e-02[1]	&1.95e-02[2]	\\
&Std	&5.30e-03	&6.27e-03	\\\hline
\multirow{2}{*}{$F_{5}$}
&Mean	&\cellcolor{gray25}1.37e-02[1]	&1.95e-02[2]	\\
&Std	&2.84e-03	&4.11e-03	\\\hline
\multirow{2}{*}{$F_{6}$}
&Mean	&1.45e-01[2]	&\cellcolor{gray25}7.04e-02[1]	\\
&Std	&4.63e-02	&6.25e-03	\\\hline
\multirow{2}{*}{$F_{7}$}
&Mean	&1.38e-01[2]	&\cellcolor{gray25}4.21e-02[1]	\\
&Std	&2.83e-02	&1.92e-02	\\\hline
\multirow{2}{*}{$F_{8}$}
&Mean	&1.16e-01[2]	&\cellcolor{gray25}5.06e-02[1]	\\
&Std	&1.40e-02	&4.17e-03	\\\hline
\multirow{2}{*}{$F_{9}$}
&Mean	&5.91e-02[2]	&\cellcolor{gray25}3.67e-02[1]	\\
&Std	&5.75e-03	&4.85e-03	\\\hline
\multicolumn{2}{|c}{Total$$}
&15	&12	\\\hline
\multicolumn{2}{|c}{Final Rank$$}
&2	&1	\\\hline
\multicolumn{2}{|c}{$+,-,\thickapprox$}
&-	&6/2/1	\\\hline
\end{tabular}
\end{table}
\begin{figure*}[htbp]\footnotesize
\graphicspath{{figs/}}
\centering
\begin{minipage}{0.24\linewidth}
\centerline{\includegraphics[width=1\columnwidth]{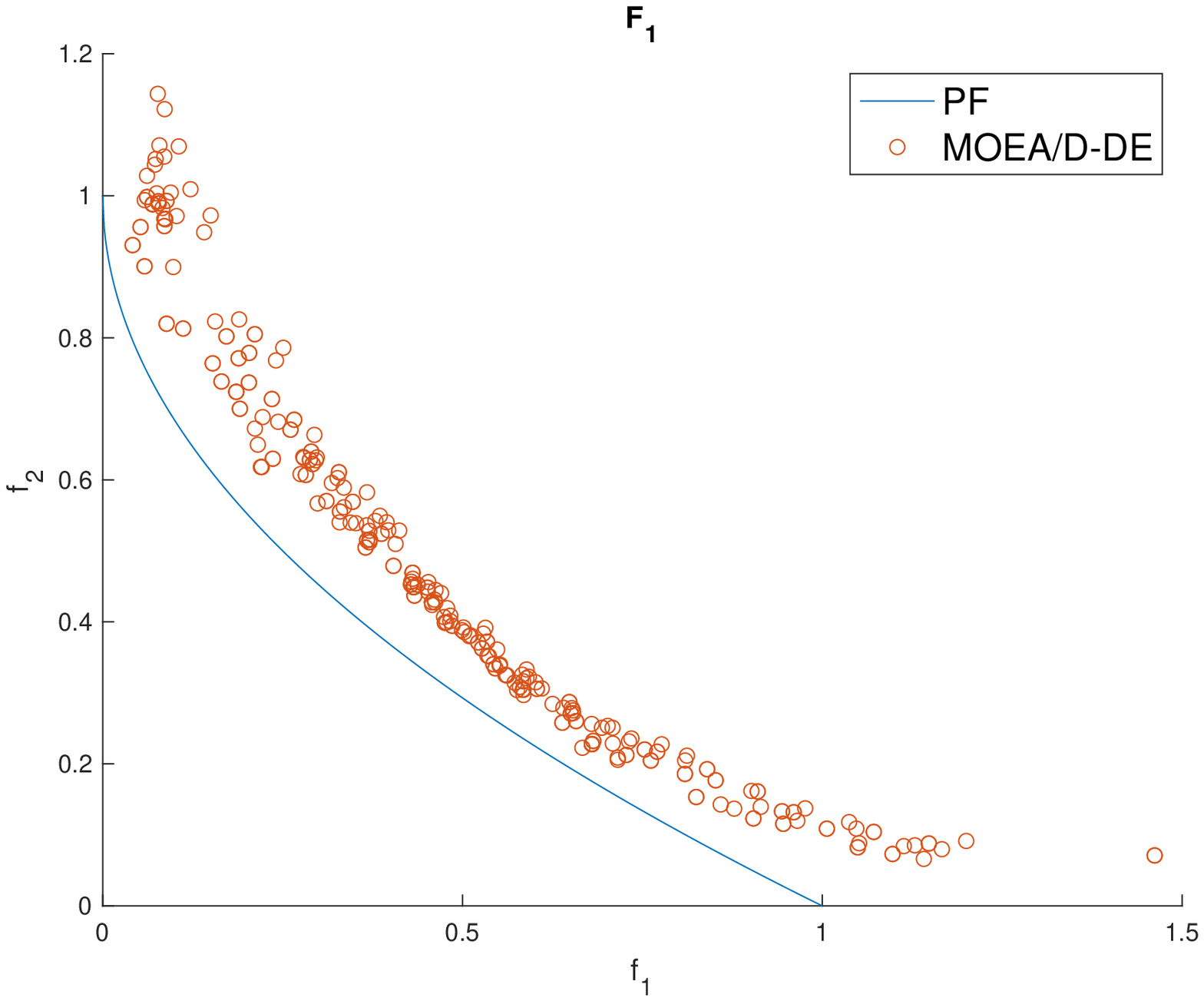}}
\end{minipage}
\begin{minipage}{0.24\linewidth}
\centerline{\includegraphics[width=1\columnwidth]{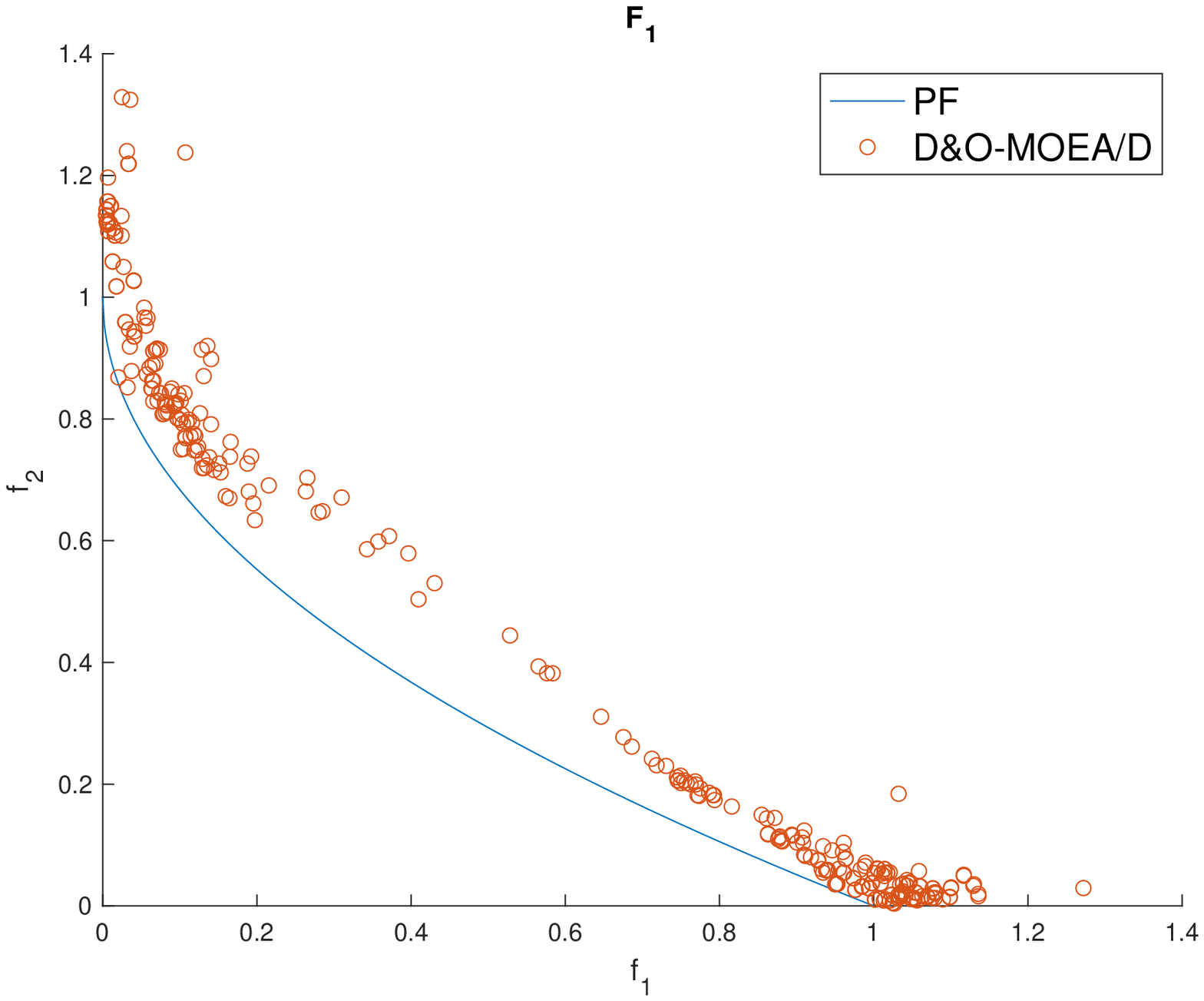}}
\end{minipage}
\begin{minipage}{0.24\linewidth}
\centerline{\includegraphics[width=1\columnwidth]{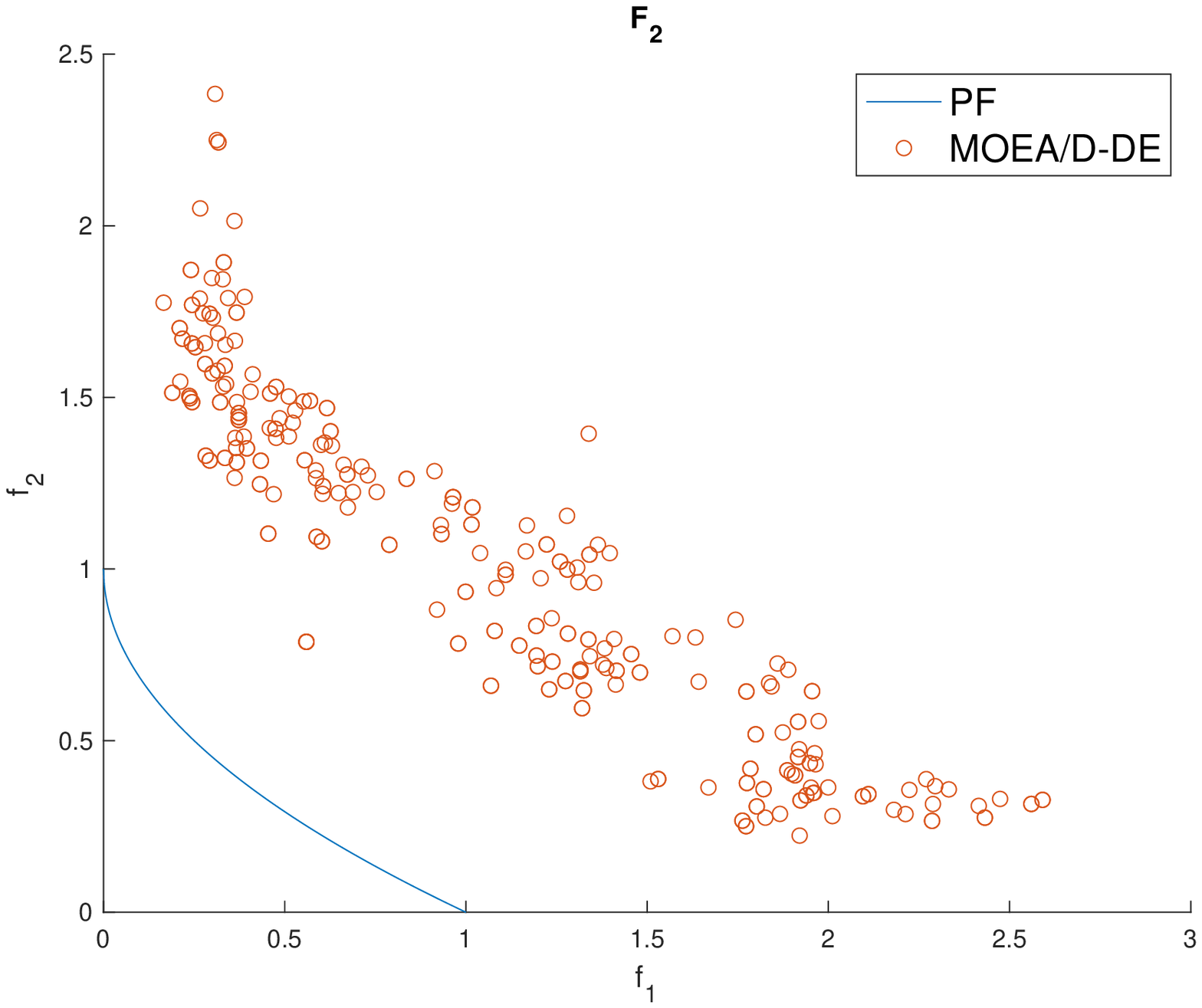}}
\end{minipage}
\begin{minipage}{0.24\linewidth}
\centerline{\includegraphics[width=1\columnwidth]{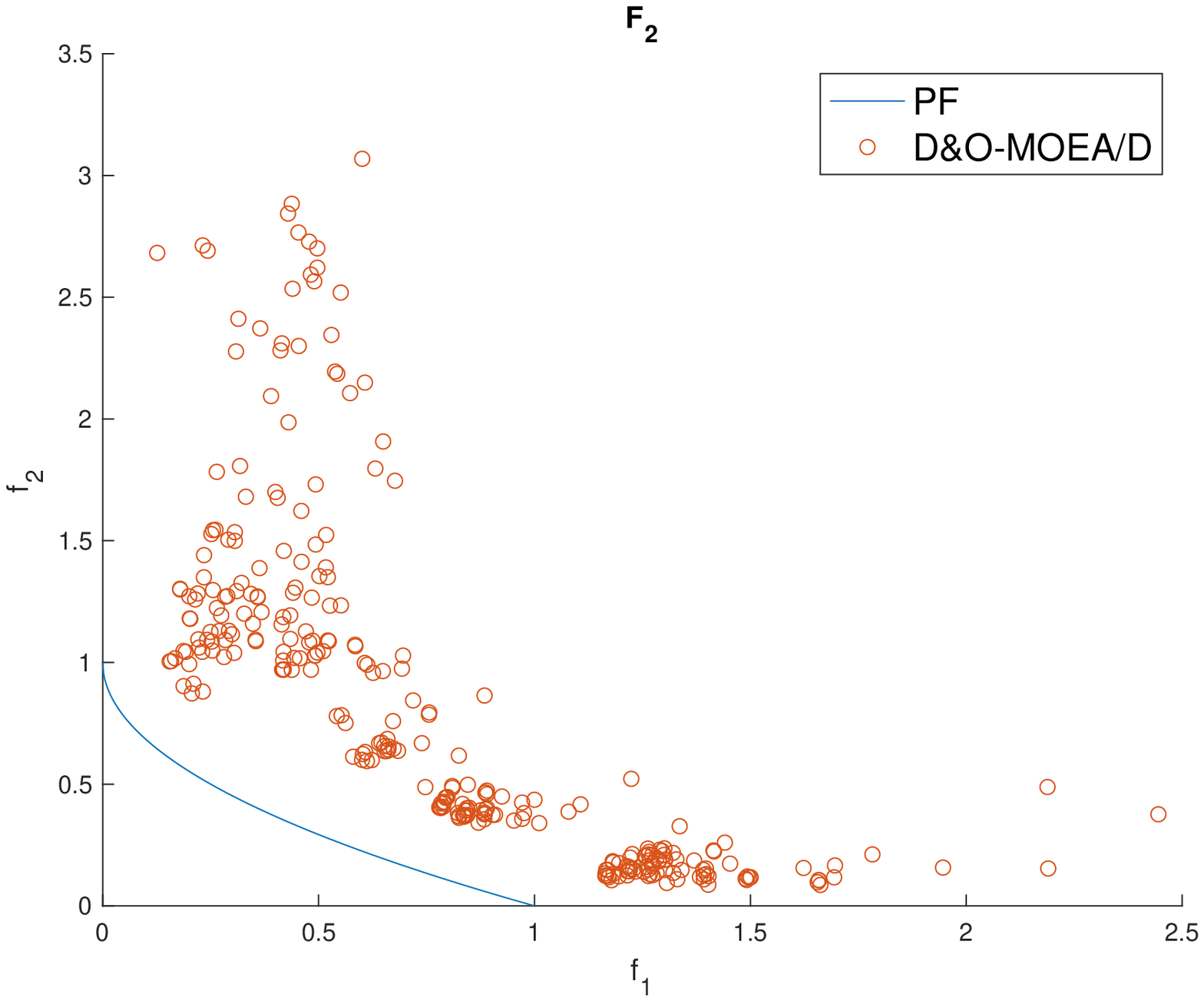}}
\end{minipage}\\
\begin{minipage}{0.24\linewidth}
\centerline{\includegraphics[width=1\columnwidth]{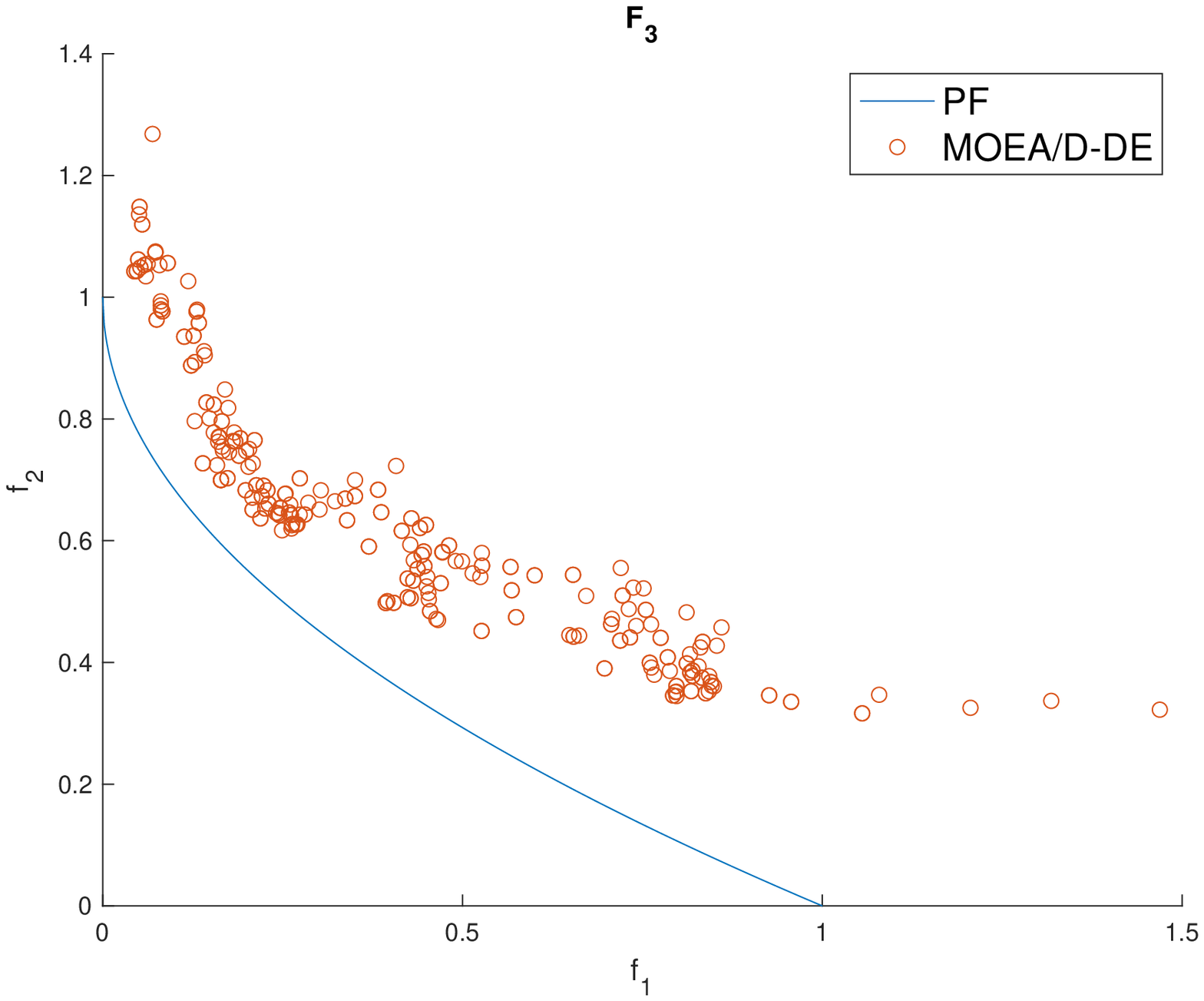}}
\end{minipage}
\begin{minipage}{0.24\linewidth}
\centerline{\includegraphics[width=1\columnwidth]{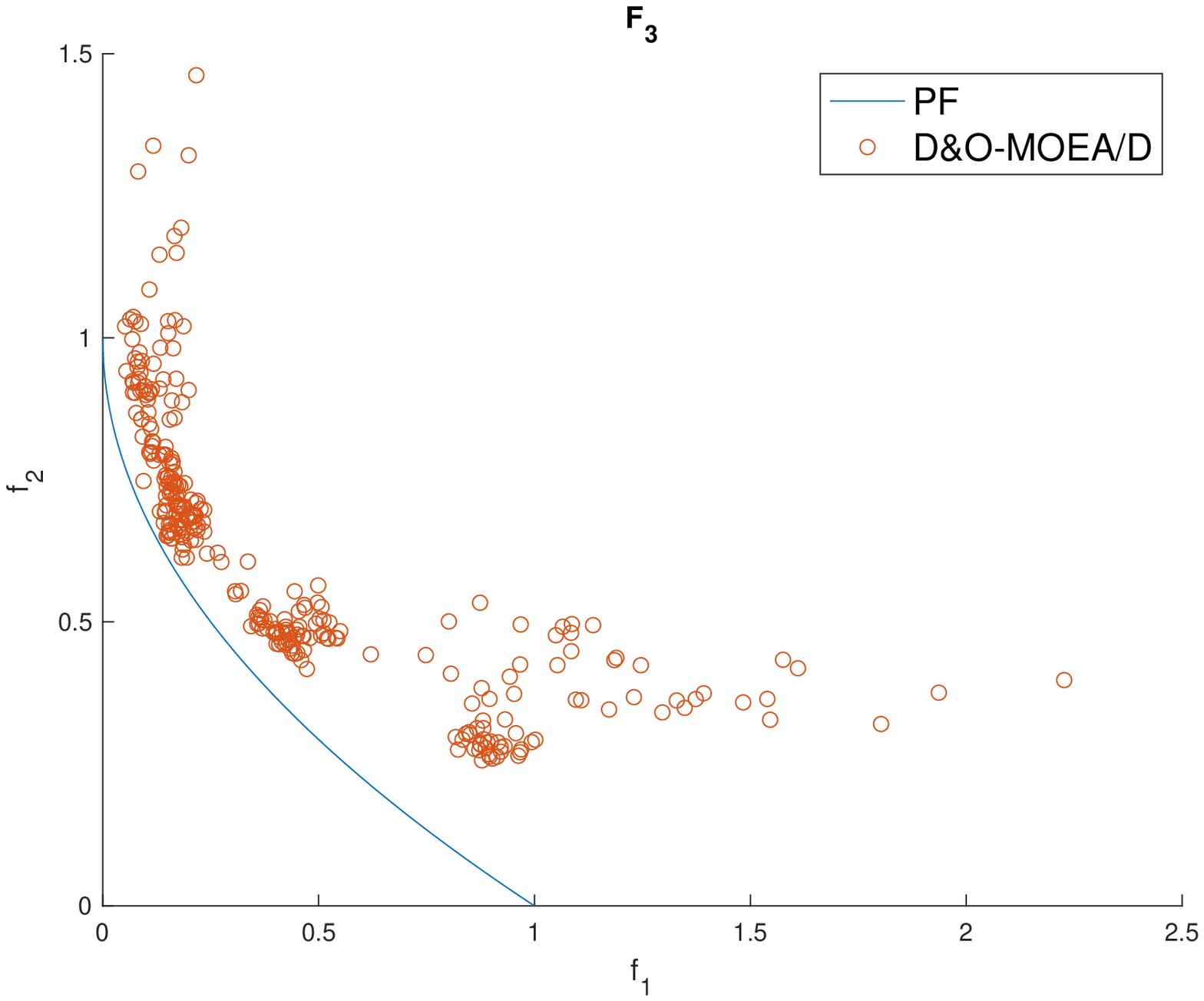}}
\end{minipage}
\begin{minipage}{0.24\linewidth}
\centerline{\includegraphics[width=1\columnwidth]{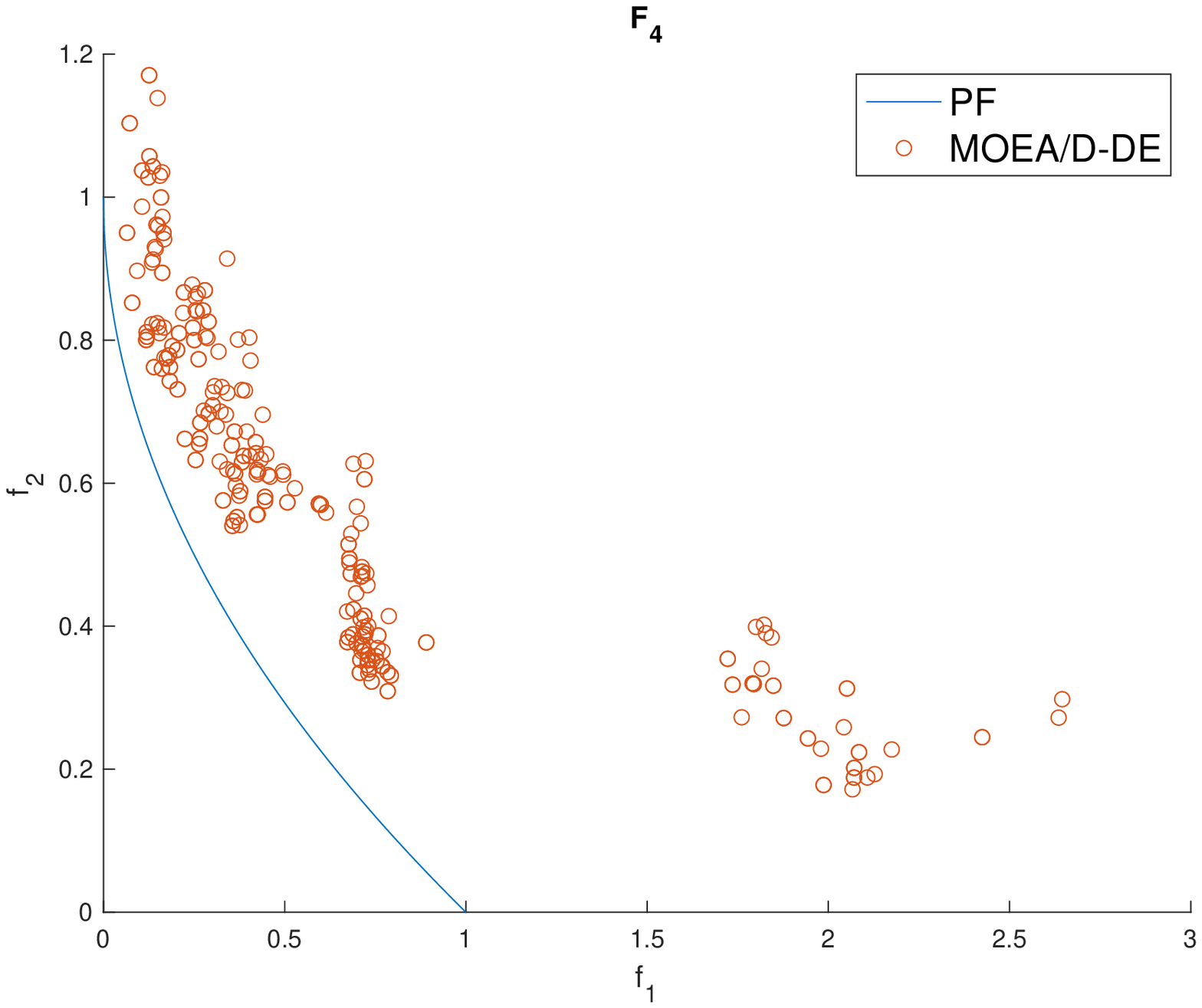}}
\end{minipage}
\begin{minipage}{0.24\linewidth}
\centerline{\includegraphics[width=1\columnwidth]{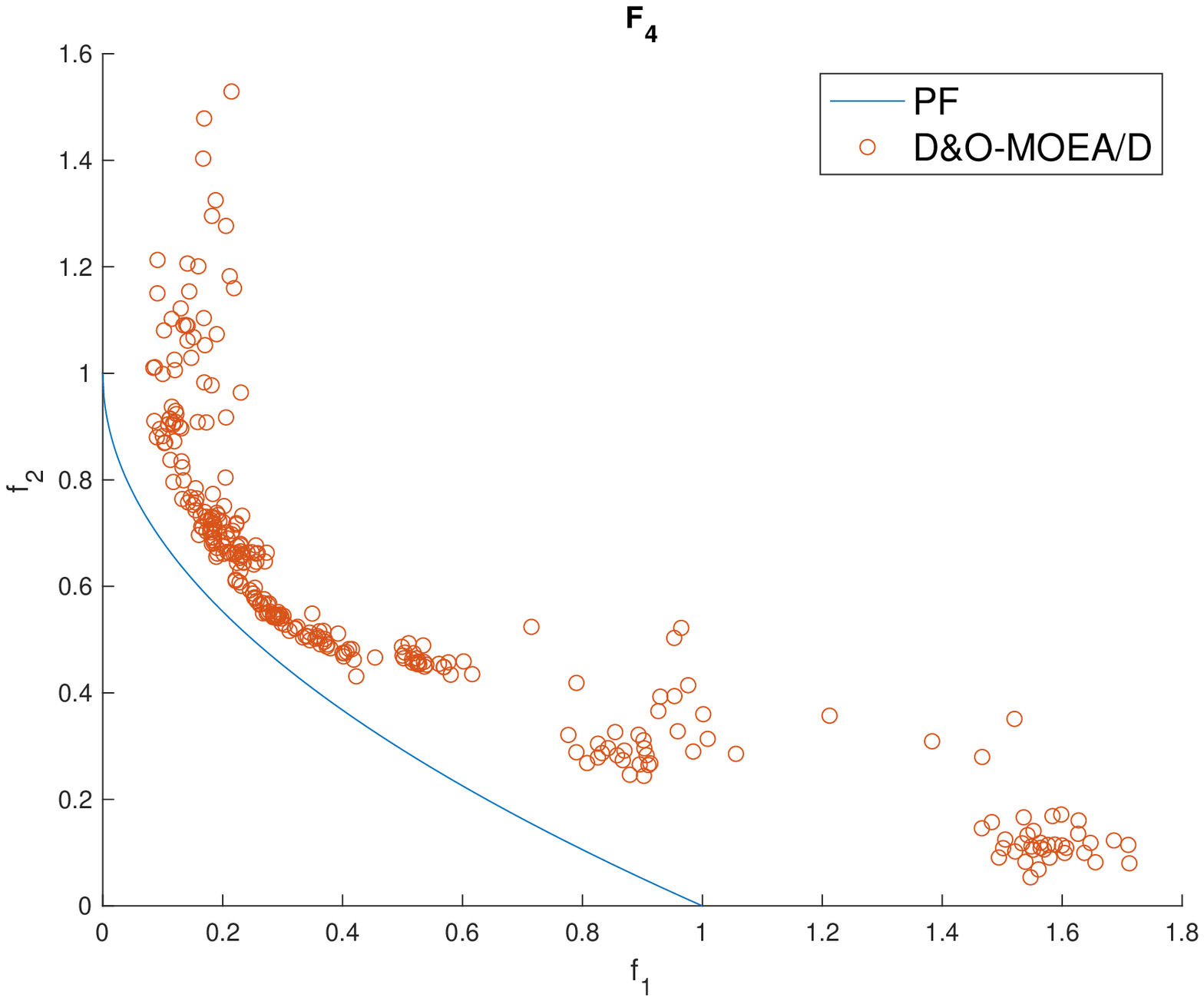}}
\end{minipage}\\
\begin{minipage}{0.24\linewidth}
\centerline{\includegraphics[width=1\columnwidth]{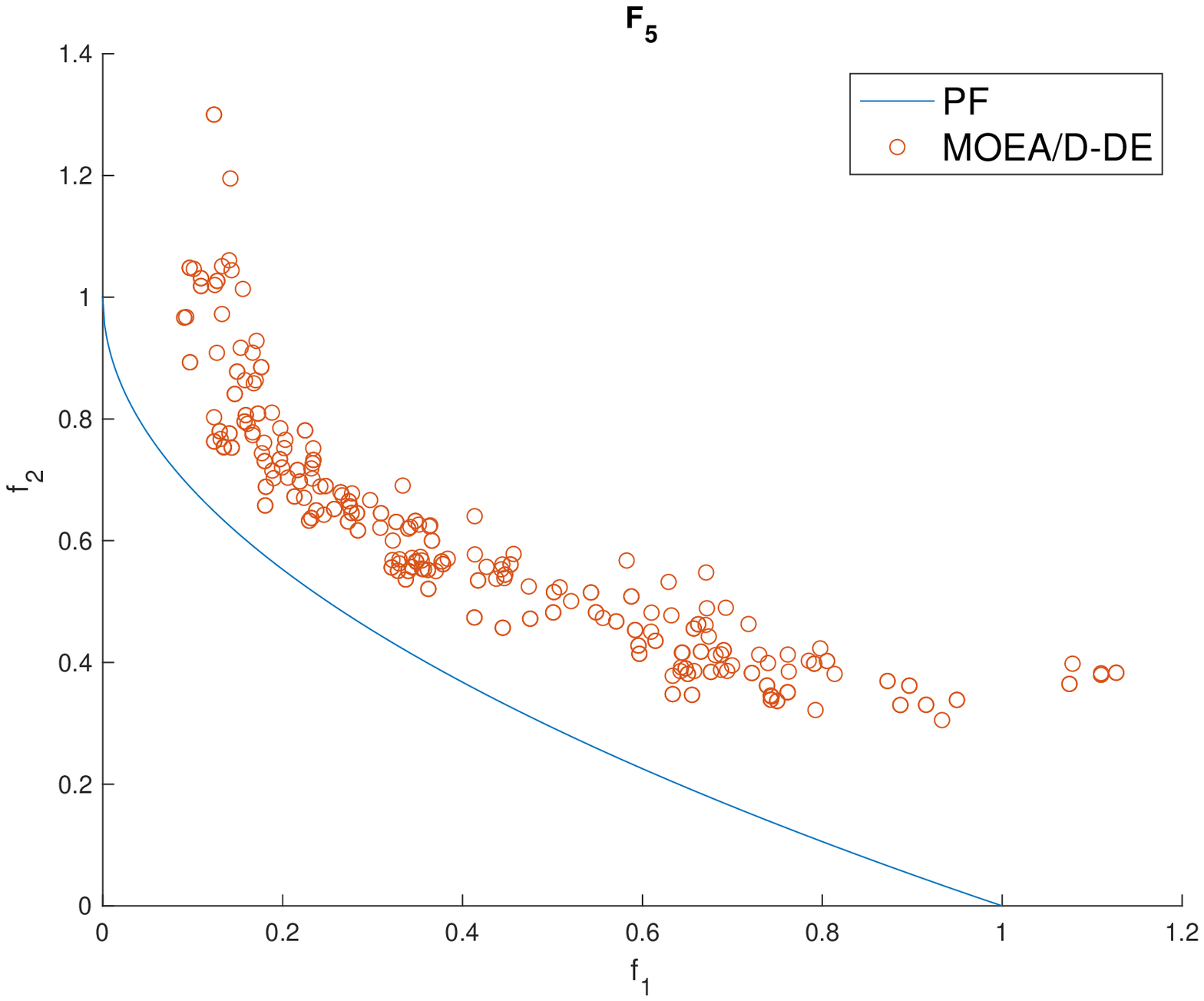}}
\end{minipage}
\begin{minipage}{0.24\linewidth}
\centerline{\includegraphics[width=1\columnwidth]{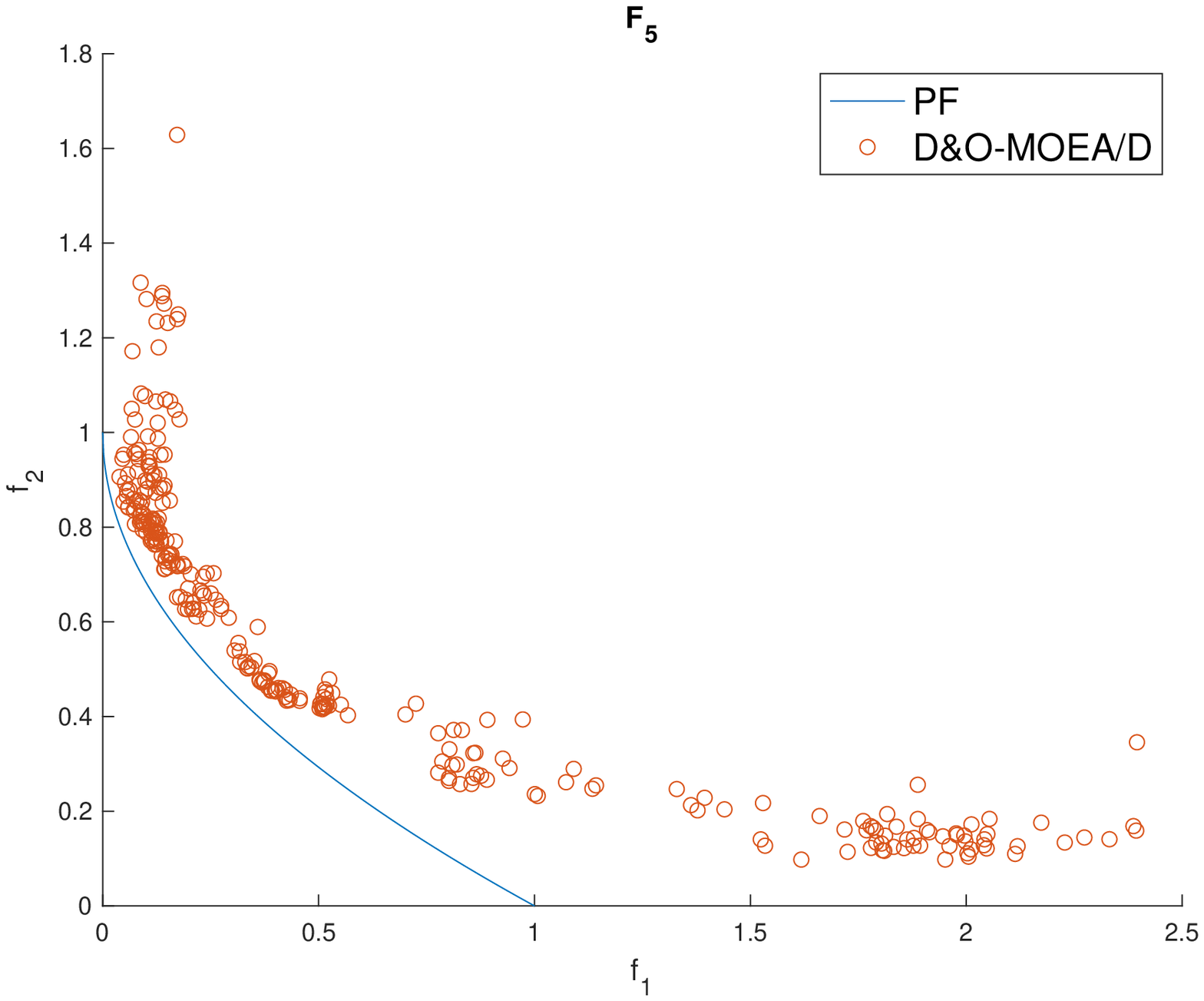}}
\end{minipage}
\begin{minipage}{0.24\linewidth}
\centerline{\includegraphics[width=1\columnwidth]{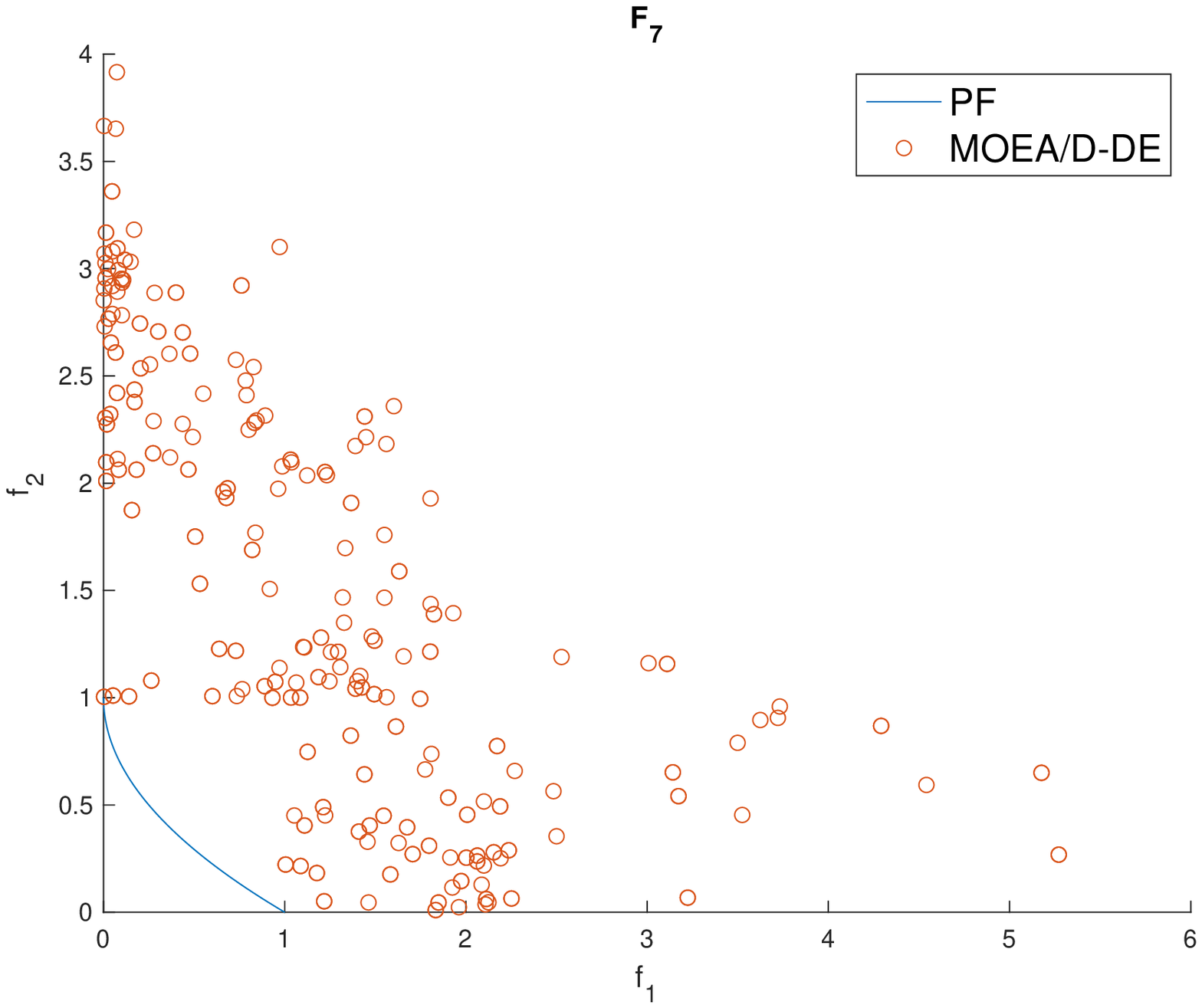}}
\end{minipage}
\begin{minipage}{0.24\linewidth}
\centerline{\includegraphics[width=1\columnwidth]{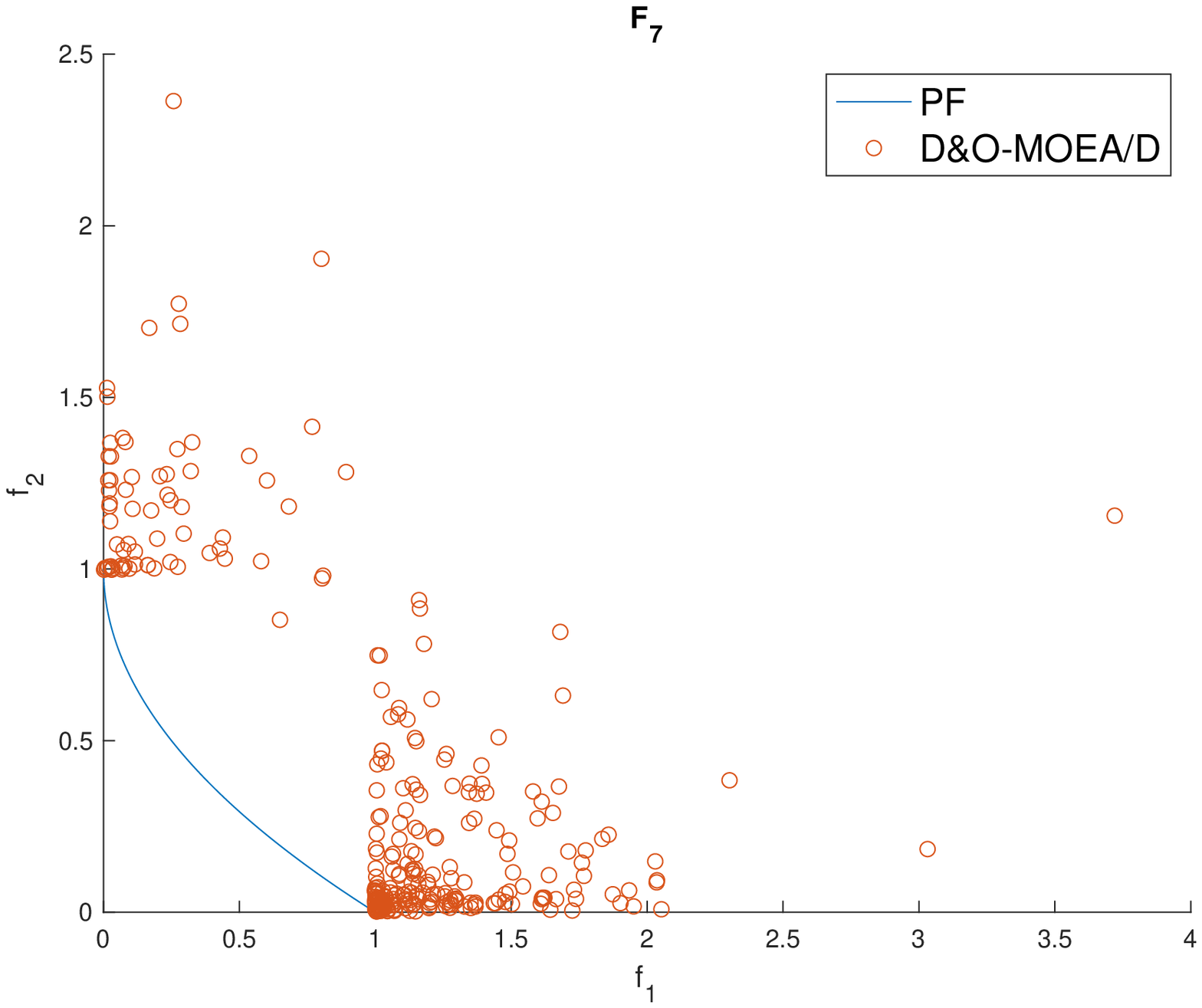}}
\end{minipage}\\
\begin{minipage}{0.24\linewidth}
\centerline{\includegraphics[width=1\columnwidth]{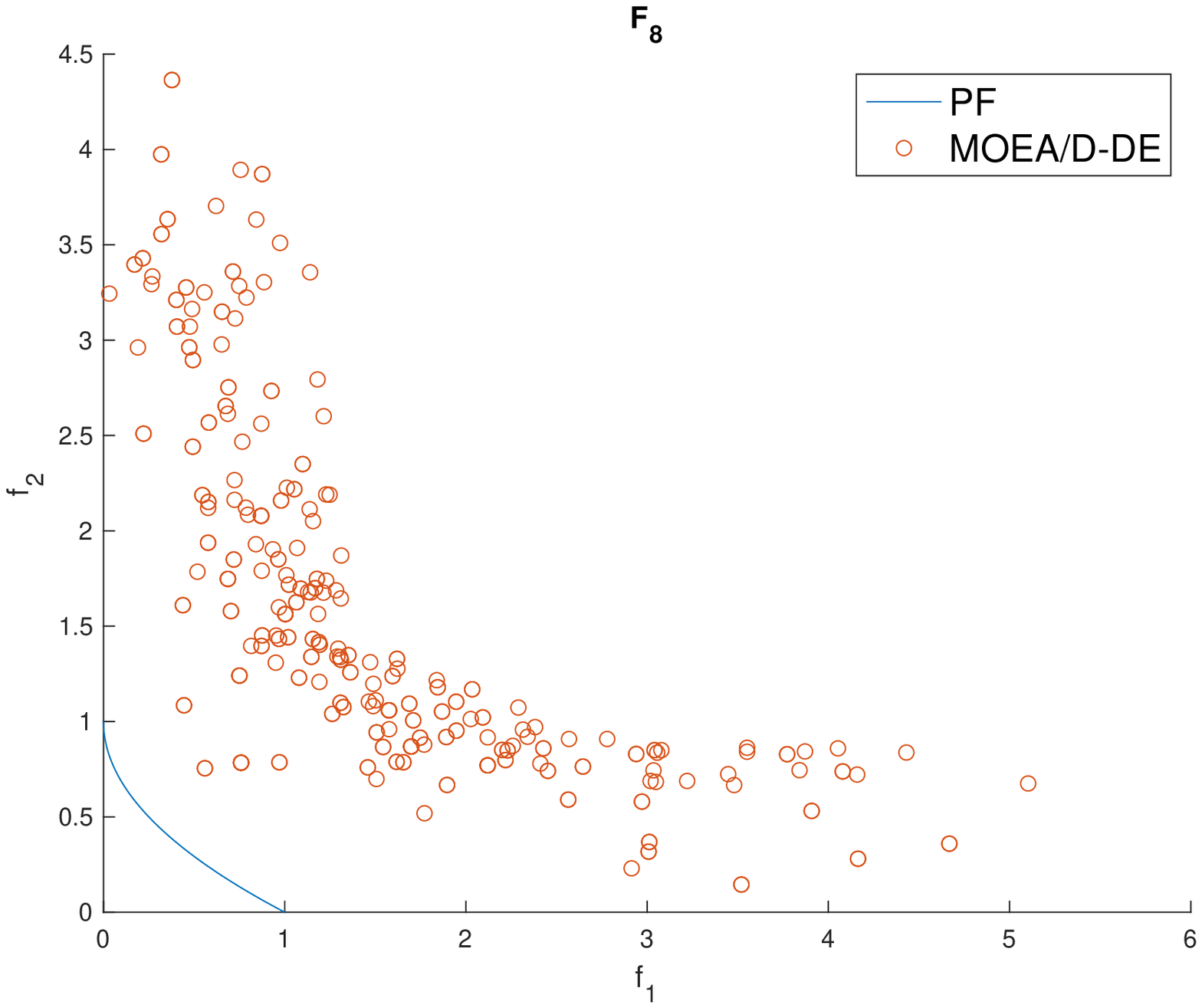}}
\end{minipage}
\begin{minipage}{0.24\linewidth}
\centerline{\includegraphics[width=1\columnwidth]{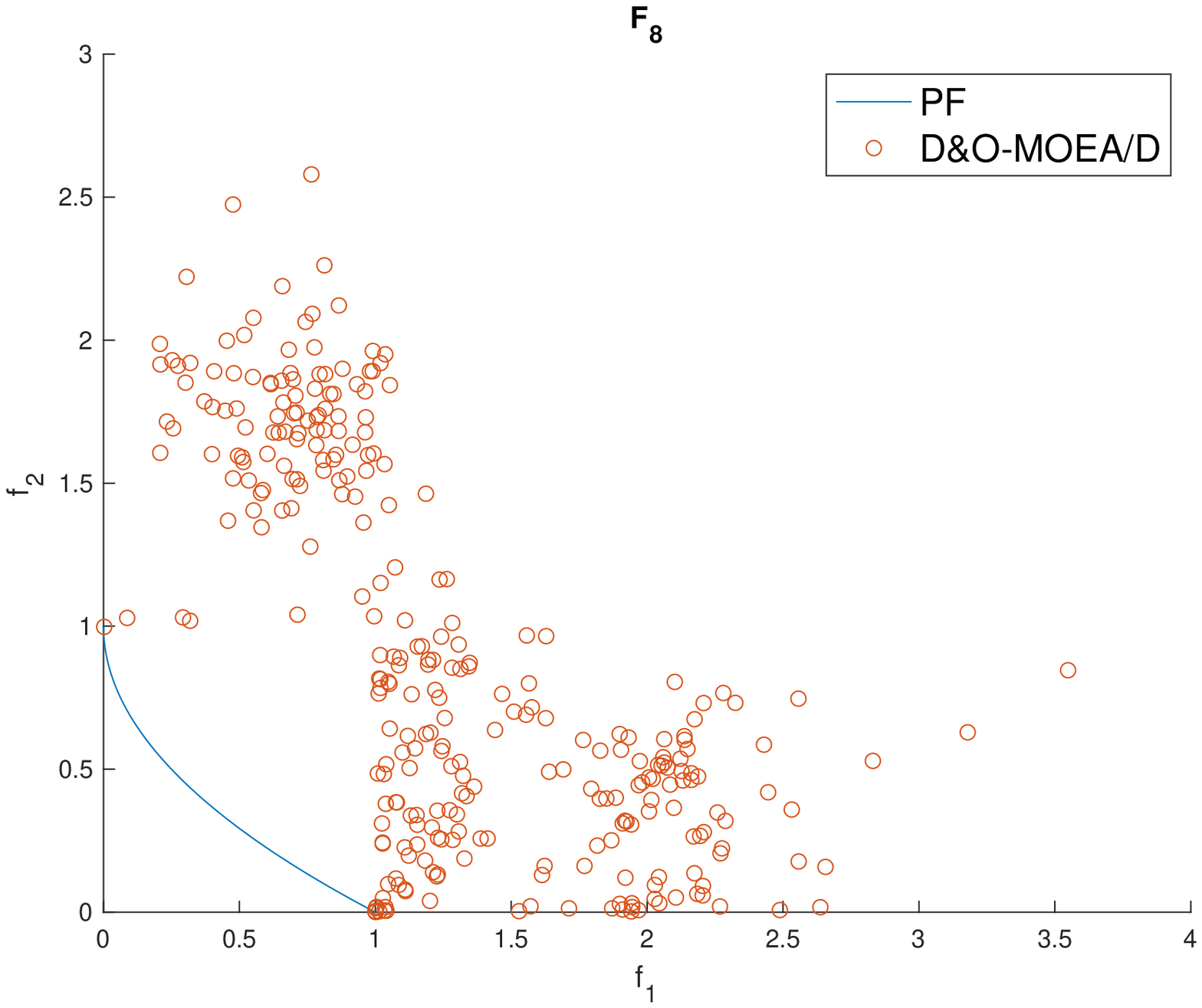}}
\end{minipage}
\begin{minipage}{0.24\linewidth}
\centerline{\includegraphics[width=1\columnwidth]{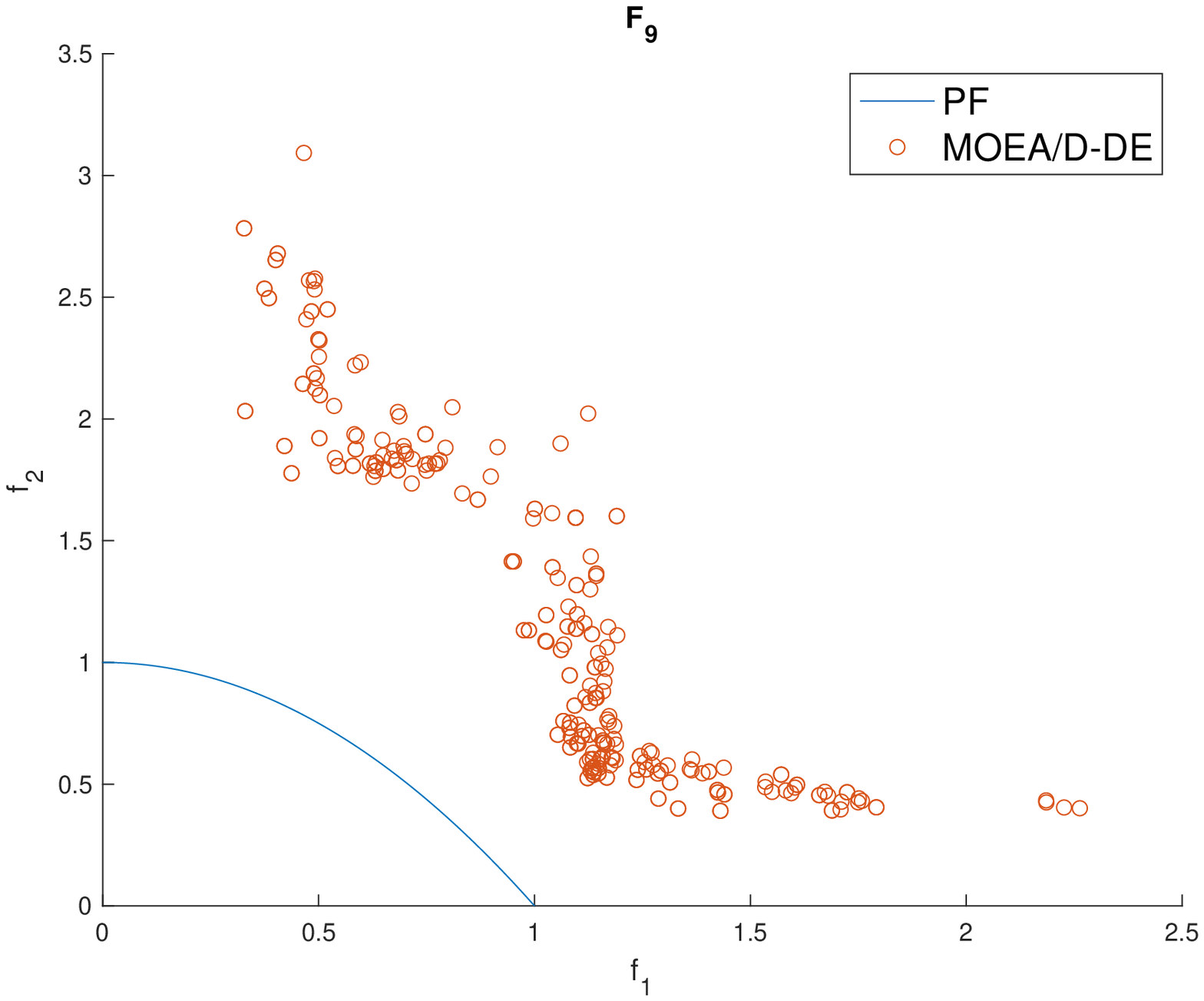}}
\end{minipage}
\begin{minipage}{0.24\linewidth}
\centerline{\includegraphics[width=1\columnwidth]{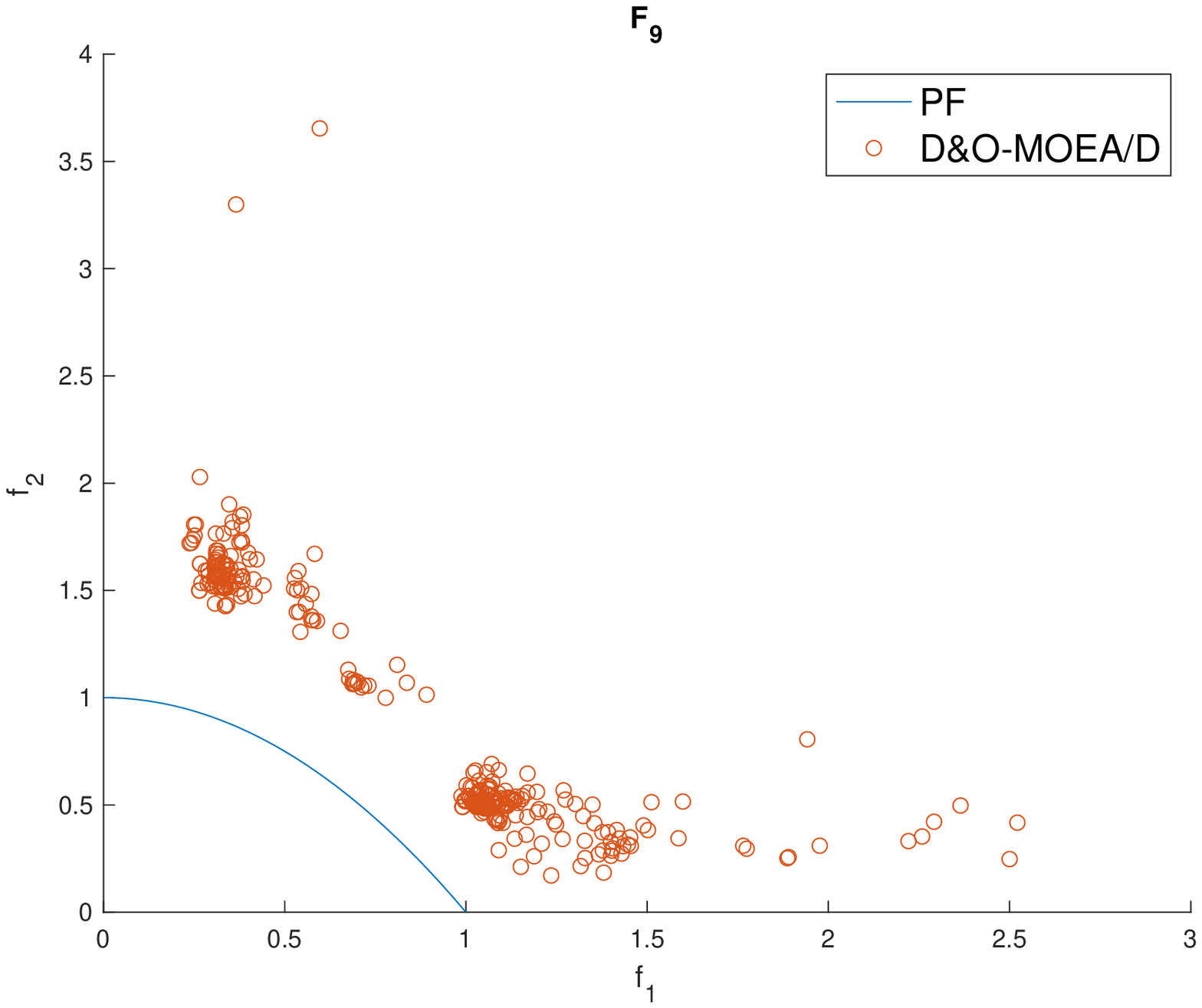}}
\end{minipage}\\
 \caption{The efficient fronts with the best HVs obtained by MOEA/D-DE and D\&O-MOEA/D.} 
 \label{nopt_each_fronts}
\end{figure*}

\end{document}